 \documentclass[final,authoryear,5p,times,twocolumn]{elsarticle}

\usepackage{graphicx}
\usepackage{hyperref}
\hypersetup{backref,colorlinks=true}
\usepackage{amssymb}
\usepackage[cmex10]{amsmath}
\usepackage{makecell}
\usepackage[english]{babel}
\usepackage{bbm}
\usepackage{wrapfig}
\usepackage[]{subfig}
\usepackage{caption}
\usepackage{epstopdf}
\usepackage{epsfig}
\usepackage{rotating}
\usepackage{array}
\usepackage{soul}
\usepackage{multirow}
\usepackage{pifont}
\newcolumntype{M}
{>{\centering\arraybackslash}m{\dimexpr.25\linewidth-2\tabcolsep}}
\usepackage{color}
\definecolor{MyGreen}{rgb}{0,0.5,0}
\definecolor{MyRed}{rgb}{0.8,0,0}
\definecolor{MyPurple}{rgb}{1,0,1}
\definecolor{MyLightGray}{rgb}{.8,0.8,0.8}

\newcommand{\note}[1]{{#1}}

\newcommand{\bs}[1]{{\boldsymbol{#1}}}
\newcommand{\tickyes}[0]{{\color{MyGreen}\checkmark}}
\newcommand{\tickno}[0]{{\color{MyRed}\ding{55}}}
\usepackage{rotating}

\journal{arXiv.org}

\begin{document}
\begin{frontmatter}


\title{Incorporating prior knowledge in medical image segmentation: a survey}


\author{Masoud S. Nosrati and   Ghassan Hamarneh}
\address{Medical Image Analysis Lab, School of Computing Science, Simon Fraser University, 8888 University Dr., BC, Canada\\
\{smn6, hamarneh\}@sfu.ca}

\begin{abstract}
Medical image segmentation, the task of partitioning an image into meaningful parts, is an important step toward automating medical image analysis and is at the crux of a variety of medical imaging applications, such as computer aided diagnosis, therapy planning and delivery, and computer aided interventions. 
However, the existence of noise, low contrast and objects' complexity in medical images are critical obstacles that stand in the way of achieving an ideal segmentation system.
Incorporating prior knowledge into image segmentation algorithms has proven useful for obtaining more accurate and plausible results.
This paper surveys the different types of prior knowledge that have been utilized in different segmentation frameworks. We focus our survey on  optimization-based methods that incorporate prior information into their frameworks. We review and compare these methods in terms of the types of prior employed, the domain of formulation (continuous vs. discrete),  and the optimization techniques (global vs. local).  We also created an interactive online database of  existing works and categorized them based on the type of prior knowledge they use. Our website is  interactive so that researchers can contribute to keep the database up to date. We conclude the survey by discussing different aspects of designing an energy functional for image segmentation, open problems, and future perspectives.

\end{abstract}

\end{frontmatter}
% \linenumbers

\section{Introduction}
\label{sec:intro}
Image segmentation is the process of partitioning an image into smaller meaningful regions based in part on
some homogeneity characteristics. The goal of segmentation is to delineate (extract or contour) targeted objects for further analysis.

For example, in medical image analysis (MIA), image segmentation of organs or tissue types is a necessary first step for numerous applications, e.g. measuring tumour burden (or volume)  from positron emission   tomography  (PET) or computed tomography (CT) scans \citep{hatt2009fuzzy,bagci2013joint}, analyzing vasculature from magnetic resonance angiography (MRA) (e.g. measuring tortuosity) \citep{bullitt2003measuring,yan2006segmentation}, grading cancer from histopathology images \citep{tabesh2007multifeature}, performing fetal measurements from prenatal ultrasound \citep{carneiro2008detection}, performing augmented reality in robotic image guided surgery \citep{su2009augmented,pratt2012effective}, building statistical atlas for population studies and voxel-based morphometry \citep{ashburner2000voxel}.

Given an input image, $I$, the goal of a typical image segmentation system is to assign every pixel in $I$ a specific label where each label represents a structure of interest.
%More formally, given an $m$-channel $n$-dimensional image $I:\Omega\subset\mathbb{R}^n\rightarrow\mathbb{R}^m$ and a set of labels $\mathcal{L}=\{l_1,\cdots,l_k\}$, that represents $k$ objects of interest in $I$, a typical image segmentation system  assigns every pixel $\boldsymbol{x}\in\Omega$ a label $l\in \mathcal{L}$. Here, we denote the segmented image by $S:\Omega\rightarrow \mathcal{L}$. 
 Several traditional segmentation algorithms have been proposed for assigning labels to pixels; these include thresholding \citep{otsu1975threshold,sahoo1988survey}, region-growing \citep{adams1994seeded,pohle2001segmentation,pan2007bayes}, watershed \citep{vincent1991watersheds,grau2004improved,hamarneh2009watershed} and optimization-based methods \citep{grady2012targeted,mcintoshmedical,ulen2013efficient}. The existence of noise, low contrast and objects complexity in medical images, typically cause the aforementioned methods to fail. 
In addition, all these traditional methods assume that objects' entire appearance have some notion of homogeneity; however, this is not necessarily the case for complex  objects (e.g. multi-region cells with membrane, nucleus and nucleolus; or brain regions affected by magnetic field  of a magnetic resonance imaging (MRI) device  non-uniformity). Many real-world objects are better described by a combination of regions with distinct appearance models. This is where more elaborate \textit{prior information} about the targeted objects becomes helpful.

The majority of state-of-the-art image segmentation methods are formulated as optimization problems, i.e. energy minimization or maximum-a-posteriori estimation, mainly because of their: 1) formal and rigorous mathematical formulation, 2) availability of  mathematical tools for optimization, 3) capability to incorporate multiple (competing) criteria as terms in the objective function, 4) ability to quantitatively measure the extent by which a method satisfies the different criteria/terms, and 5) ability to examine the relative performance of different solutions.

In this paper, we review the various types of prior information that are utilized in different optimization-based frameworks for segmentation of targeted objects. Prior information can take many forms: user interaction; appearance models; boundaries and edge polarity; shape models; topology specification; moments (e.g. area/volume and centroid constraints); geometrical interaction and distance prior between different regions/labels; and atlas or  pre-known models.
We compare the different methods utilizing prior information in image segmentation in terms of the type of  prior information utilized, domain of formulation (continuous vs. discrete) and optimization techniques (global vs. local) used.

The rest of the paper is organized as follows. In Section \ref{sec:surveys}, we briefly review the previous surveys that covered the medical image segmentation (MIS)  problems and justify the need for our survey. In Section \ref{sec:fundamental}, we review the fundamentals of optimization-based image segmentation techniques. In Section \ref{sec:prior},
we give a concrete overview of the different types of prior knowledge devised to improve image segmentation. Finally, in Section \ref{sec:conclusion}, we summarize our notes and elaborate on future perspectives.

%applications of image segmentation
%instead focus on identifying those pixels belonging to a specific object or objects (which we will
%call targeted segmentation) for which there are some known characteristics.
%
% The segmentation
%of a specific object from the background is not just a special case of the traditional image segmentation
%problem which is restricted to two labels. Instead, a targeted image segmentation algorithm must input the
%additional information that determines which object is being segmented. 
%
%
%This additional information, which
%we will call targeted specification, can take many forms: user interaction, appearance models, pairwise pixel
%affinity models, contrast polarity, shape models, topology specification, relational information and/or feature
%

\section{Why yet another survey paper on MIS?}
\label{sec:surveys}
Many survey papers on the topic of medical image segmentation have appeared before and in this section, we explore the related surveys.

\cite{mcinerney1996deformable} reviewed the development and application of deformable models to problems in medical image analysis, including segmentation, shape representation, matching and motion tracking. \cite{pham2000current}  surveyed some of the segmentation approaches such as thresholding, region growing and deformable models  with an emphasis on the advantages and disadvantages of these methods for medical imaging applications.  \cite{olabarriaga2001interaction} presented a review of the user interaction in image segmentation. In \cite{olabarriaga2001interaction}, the goal was to  identify patterns in the use of user-interaction and to propose a criteria to evaluate interactive segmentation techniques.  All of these surveys are limited to specific segmentation techniques that adopted a few basic priors such as intensity and texture. In addition, the aforementioned surveys \citep{mcinerney1996deformable, pham2000current,olabarriaga2001interaction} are more than 10 years old and many important new developments have appeared since then.

\cite{elnakib2011medical} and \cite{hu2009survey} focus on appearance and shape features and overview most popular medical image segmentation techniques such as deformable models and atlas-based segmentation techniques.   \cite{heimann2009statistical} reviewed several statistical shape modelling approaches. Shape representation, shape correspondence, model construction, local appearance model, and search algorithms structured their survey. \cite{Peng20131020} present five categories of graph-based discrete optimization strategies to MIS within a graph-theoretical perspective and hence do not discuss specific forms of priors the way we do in this survey.  
 \cite{crc2013a} surveyed the field of energy  minimization in MIS	and provided an overview on the energy function, the segmentation and image representation, the training data, and the minimizers. Many advanced prior information proposed in recent years have not been included in the aforementioned surveys.

Some surveys are focused on specific modalities only. As an example,  \cite{noble2006ultrasound} focused on reviewing methods on  ultrasound segmentation in different medical applications, including cardiology, breast, and prostate. In \citep{sharma2010automated}, the details of automated segmentation methods, specifically in the context of CT and MR images, were	 discussed. All the method discussed in this survey adopted limited forms of priors such as appearance, edge, and shape.

Other surveys focused on specific organs. For example \cite{lesage2009review} reviewed literature on  3D vessel lumen segmentation while \cite{petitjean2011review} reviewed fully and semi-automated methods performing segmentation in short axis images of cardiac cine MRI sequences. They propose a categorization for cardiac segmentation methods based on what level of external information (prior knowledge) is required and how it is used to constrain segmentation. However, the discussed priors in \cite{petitjean2011review} are limited to shape and appearance information in deformable contour and atlas-based methods. 

The recent paper, \cite{grady2012targeted} is  most similar to this survey and includes a solid introduction to targeted object segmentation. However, this survey focused on graph-based methods only and studied only a subset of priors we cover in this report. 

In this survey, we categorize several prior information based on their types and discuss how these priors have been encoded into segmentation frameworks both in continuous and discrete settings.  The prior information discussed in this survey has been proposed in both computer vision and medical image analysis communities.  We hope this survey would be useful for the MIA community and the users of medical image segmentation techniques by: summarizing what types of priors exist so far; which ones can be useful for one's own targeted segmentation problems; which methods (papers) have incorporated such priors in their formulation already; the approach adopted for incorporating certain priors (the associated complexities and trade-offs). We also hope that by surveying what has been done, researchers can more easily identify existing ``gaps'' and ``weaknesses'', e.g.  identifying other important priors that have been missed so far and require future research; or proposing better techniques for incorporating known priors.

\section{Fundamentals of image segmentation}
\label{sec:fundamental}

\subsection{Traditional image segmentation methods}
\label{sec:traditional}
As mentioned in Section \ref{sec:intro},  several traditional segmentation algorithms have been proposed in the literature including thresholding, region-growing, and watershed.

\emph{Thresholding} is the simplest segmentation technique where, for a simple case of binary segmentation (foreground vs. background), the input image is divided into regions: regions with values either less or more than a threshold. Determining more than one threshold value is called multithresholding \citep{sahoo1988survey}. In thresholding, segmentation results depend on the image properties and on how the threshold is chosen (e.g. using Otsu's method \citep{otsu1975threshold}). Thresholding techniques do not take into account the spatial relationships between features in an image and thus are very sensitive to noise. 

\emph{Region growing} methods start from a set of seed pixels defined by the user and examine neighbouring pixels of the seeds to  determine whether the neighbouring pixels should be added to the region preserving some uniformity and connectivity criteria \citep{adams1994seeded}. 
Different variations of this technique have been applied on different medical image modalities \citep{pohle2001segmentation, pan2007bayes}.  
Region growing methods consider the neighbourhood information of pixels, and hence,  they are more robust to noise compared to thresholding methods. However, these methods are sensitive to the chosen  ``uniformity predicate'' (a logical statement for evaluating the membership of a pixel) and corresponding threshold \citep{adams1994seeded},  the location of seeds, and type of pixel connectivity.

In conventional watershed algorithms \citep{vincent1991watersheds}, an image may be seen as a topographic relief, where the intensity value of a pixel is interpreted as its altitude in the relief. Suppose that the entire topography is flooded with water through virtual ``holes'' at the bottom of basins, then as the water level rises and water from different basins are about to merge, a dam is built to prevent merging. These dam boundaries correspond to the watershed lines or objects boundaries.  \note{An improved version of the watershed technique has been used to segment brain MR images} \citep{grau2004improved,hamarneh2009watershed}. In practice, watershed  produces  over-segmentation due to noise or local irregularities in the  image. Marker-based watershed \citep{grau2004improved,vincent1993morphological,beucher1994watershed} prevents over-segmentation by limiting  the number of regional minima. However, this method is also sensitive to noise.

Having prior knowledge about the objects of interest and incorporating this knowledge  into the segmentation framework helps us overcome the shortcomings associated with the traditional methods and obtain more plausible results. As mentioned earlier, formulating image segmentation as an optimization problem allows for the use  of multiple criteria and prior information as energy terms in an objective functional.  In the following section, we briefly review the fundamentals of optimization-based techniques for medical image segmentation.

\subsection{Optimization-based image segmentation}
\label{sec:optimizationbased}
Given an image $I:\Omega\subset\mathbb{R}^n\rightarrow\mathbb{R}^m$, image segmentation partitions $\Omega$ into $k$ disjoint regions $\bs{S}=\{S_1,\cdots,S_k\}\subset \mathcal{S}$ such that $\Omega=\cup_{i=1}^kS_i$ and $S_i\cap S_j=\emptyset,\;\forall i\neq j$. 
$\mathcal{S}$ is the solution space. The aforementioned partitioning is referred to as a crisp binary (when $k=2$) or multi-region ($k>2$) segmentation. In a  fuzzy or probabilistic segmentation, each element in $\Omega$ (e.g. a pixel) is assigned a vector $\bs{p}$ of length $k$ quantifying the memberships or probabilities of belonging to each of the $k$ classes, $\bs{p}=[p_1, p_2,\cdots,p_k]$ where $p_i\geq 0,~i=1,\cdots,k$ and $\sum_i^k p_i=1$.
This task of image partitioning can be formulated as an energy minimization problem. An energy function, $E:\mathcal{S}\rightarrow \mathbb{R}$, usually consists of several objectives that are divided into two main categories: \emph{regularization terms},  $R_i:\mathcal{S}\rightarrow\mathbb{R}$, and \emph{data terms},  $D_i:\mathcal{S}\rightarrow\mathbb{R}$. The regularization terms correspond to priors on the space of feasible solutions and penalize any deviation from the enforced prior such as shape, length, etc. The data terms measure how strongly a pixel should be associated with a specific label/segment. These objectives (regularization and data terms) can then be scalarized as:
\begin{align}
E(\bs{S}) = \lambda\sum_{i} R_i( \bs{S})+ \sum_j  D_j( \bs{S};I)
\;.
\end{align}
 The optimization problem is then formulated as:
\begin{align}
\bs{S}^* = \arg\min_{\bs{S}} E(\bs{S}) = \arg\min_{\bs{S}}  \lambda\mathcal{R}(\bs{S})+\mathcal{D}(\bs{S};I)
\;,
\label{eq0}
\end{align}
where $\bs{S}^*=\{S^*_1,\cdots,S_k^*\}$ are the optimal solutions and, for simplicity, $\mathcal{R}$ and $\mathcal{D}$ represent all the regularization and data terms, respectively.  $\lambda$ is a constant weight that balances the contribution/importance of the data term and the regularization term in the minimization problem.

%S_1^*,\cdots,S_k^*

 One example of such energy is written as:
\begin{align}
S_1^*,\cdots,S_k^* = \arg\min_{S_1,\cdots,S_k}\bigg\{\lambda\sum_{i=1}^k\int_{\partial S_i}d\bs{x}+\sum_{i=1}^k\int_{S_i}D_i(\bs{x};I)d\bs{x}\bigg\}\;,
\label{eq1}
\end{align}
where the first term (regularization term) measures the perimeter of the segmented regions $S_i$ and penalizes large perimeters, thus favouring smooth boundaries.  %This term is called \emph{regularization term} ($\mathcal{R}$) as it regularizes the segmented regions' boundaries in \eqref{eq1}. 
$D_i(\bs{x}):\Omega\rightarrow \mathbb{R}$, associated with region $S_i$, 
measures how strongly pixel $\bs{x}\in \Omega$ should be associated with region $S_i$. %The second term in \eqref{eq1} is called \emph{data term} ($\mathcal{D}$). 
 In Section \ref{sec:regularization}, we  discuss different types of regularization terms used in image segmentation problems.

An optimization-based image segmentation problem can also be formulated as a maximization problem:
\begin{align}
\bs{S}^*=\arg\max_{\bs{S}} P(\bs{S}|I)
\;,
\label{eq2}
\end{align}
where $\bs{S}^*$ is the optimal segmentation. Using Bayes' theorem, \eqref{eq2} can be written as:
\begin{align}
\bs{S}^*=\arg\max_{\bs{S}} \frac{P(I|\bs{S})P(\bs{S})}{P(I)}\equiv \arg\max_{\bs{S}}P(I|\bs{S})P(\bs{S}).
\label{eq3}
\end{align}
%The right hand side of \eqref{eq3} is written based on the assumption that the prior probability on the image evidence $P(I)$ has a uniform distribution. 
In \eqref{eq2} and \eqref{eq3}, $P(\bs{S}|I)$ is the \emph{posterior probability} that defines the degree of belief in $\bs{S}$ given the evidence $I$ (or some features of $I$), $P(I|\bs{S})$ is the image \emph{likelihood} measuring the probability of the evidence in $I$ given the segmentations $\bs{S}$, and $p(\bs{S})$ is the \emph{prior probability} that indicates the initial (prior to observing $I$) degree of belief in $\bs{S}$. Maximizing the posterior probability \eqref{eq3} is equivalent to minimizing its negative logarithm:
\begin{align}
\bs{S}^*=\arg\min_{\bs{S}}-\log P(I|\bs{S})-\log P(\bs{S}).
\label{eq4}
\end{align}
The probability \eqref{eq4} and energy \eqref{eq0} notations are related via the Gibbs or Boltzmann distribution. Ignoring the Boltzmann's constant and thermodynamic temperature (as they do not affect the optimization) and substituting $P(I|\bs{S}) \propto e^{-\mathcal{D}(\bs{S};I)}$ and $P(\bs{S})\propto e^{-\lambda \mathcal{R}(\bs{S})}$ into \eqref{eq4}, we obtain \eqref{eq0}.

To avoid terminological confusion, we emphasize that to improve a segmentation, prior knowledge can be incorporated into one or both of the regularization and data terms. Hence, the term prior knowledge itself should not be confused with the prior probability in  \eqref{eq3}.

\subsection{Domain of formulation: continuous vs. discrete}\label{compare}
\begin{figure}[!t]
\begin{center}
{\includegraphics[trim=0mm 0mm 0mm 0mm,width=1\linewidth,clip=true]{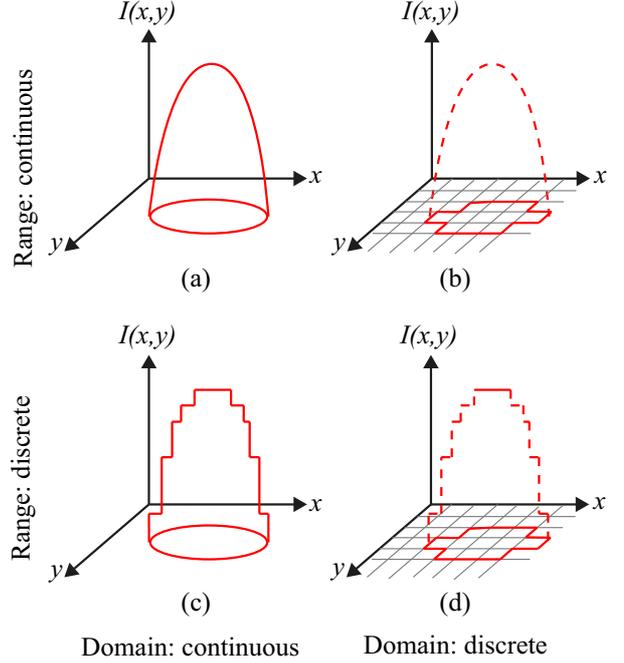}}
\end{center}
\caption{Image as a mapping. Continuous vs. discrete domain and image values.}
\label{fig:contdisc}
\end{figure}

In general, a segmentation problem can be formulated in a spatially discrete or continuous domain.  In the community that advocates continuous methods, it is assumed that the world we live in is a continuous world (continuous $\Omega$). However, images captured by digital cameras are discrete both in space and color/intensity. The discretization in space is called \emph{sampling} (discrete $\Omega$) and the discretization in color/intensity or value space is called \emph{quantization}. Given this categorization,  we have four different cases for image representation (Figure \ref{fig:contdisc}).

\begin{figure}[!t]
\begin{center}
{\includegraphics[trim=0mm 0mm 0mm 0mm,width=1\linewidth,clip=true]{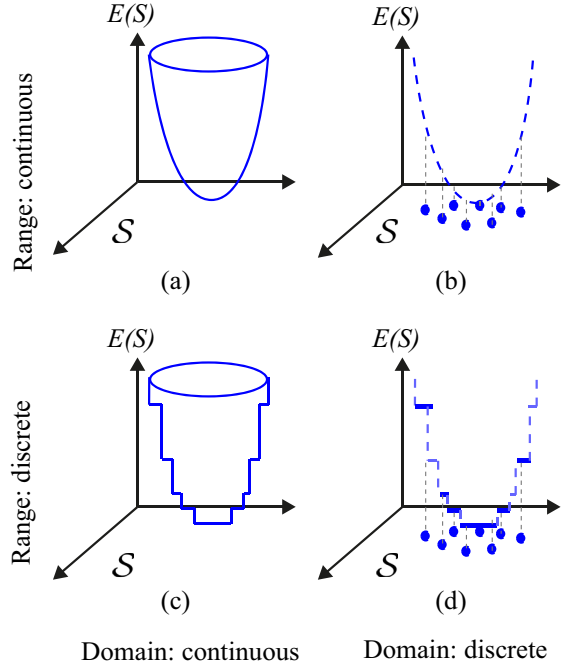}}
\end{center}
\caption{Energy function: continuous vs. discrete. $\mathcal{S}$ is the space of possible segmentations.}
\label{fig:contdiscEnergy}
\end{figure}

The energy function describing a segmentation problem can also be formulated in a discrete or continuous domain. Depending on the solution space (discrete vs. continuous) and the energy values,  four possible cases can be considered for an energy functional (Figure \ref{fig:contdiscEnergy}).
In the spatially discrete setting, the energy function is defined over a set of finite variables (nodes $\mathcal{P}\subset\Omega$ and edges), leading to the adoption of graphical models \citep{Wang20131610}. One of the most commonly used graphical models is the Markov random field (MRF) \citep{Wang20131610}. In MRF formulations, solutions are often calculated using graph cut methods, e.g. max-flow/min-cut algorithms or graph partitioning methods. Conversely, in the spatially continuous setting, energy functionals are continuous and so are the optimality conditions, which are written in terms of a set of partial differential equations (PDE). 
The minimization problem in \eqref{eq1} is a continuous version of a multi-region segmentation functional, often called \emph{minimal partition problem} in the PDE community \citep{nieuwenhuis2013survey}. Note that in Figure \ref{fig:contdiscEnergy}, the objective function is a cost or an energy function that has to be minimized. Nevertheless, an objective function can also be a fitness or utility function that has to be maximized.

In the discrete setting, the segmentation task usually begins with  an undirected graph, $\mathcal{G}(\mathcal{P},\mathcal{E})$, that is composed of vertices $\mathcal{P}$ and undirected edges $\mathcal{E}$. Each node of the graph ($p\in\mathcal{P}$)  represents a random variable ($f^i_p$) taking on different labels ($i\in\mathcal{L}=\{l_1,\cdots,l_k\}$) and each edge encodes the dependency between neighbouring variables. The corresponding optimization problem of \eqref{eq1} in the discrete domain is:
\begin{align}
&\min_{\boldsymbol{f}}\bigg\{\sum_{pq\in\mathcal{N}^i} V(f_p^i,f_q^i)+\sum_{p\in\mathcal{P}}
D_p(f_p)\bigg\} \label{discrete}\\
&\text{s.t.}~~\sum_{i\in\mathcal{L}} f^i_p =1,~~\forall p\in\mathcal{P}\notag
\;,
\end{align}
where $V$ is the regularization term (pairwise term) that encourages spatial coherence by penalizing discontinuities between neighbouring pixels, $D$ is the data penalty term (unary term),  $\boldsymbol{f}\in\mathbb{B}^{\mathcal{L}\times\mathcal{P}}$  are the binary variables ($f^i_p=1$ if pixel $p\in\mathcal{P}$ belongs to region $i\in\mathcal{L}$ and $f^i_p=0$ otherwise) and  $\mathcal{N}^i$ is the neighbourhood which is typically defined as nearest neighbour grid connectivity. 

There are several advantages and drawbacks associated with discrete and continuous methods:
%\begin{figure}[!t]
%\begin{center}
%\subfloat[Continuous]{\includegraphics[trim=0mm 0mm 0mm 0mm,width=.3\linewidth,clip=true]{cont.png}}~~~~~~~
%\subfloat[Discrete]{\includegraphics[trim=0mm 0mm 0mm 0mm,width=.3\linewidth,clip=true]{disc.png}}
%\end{center}
%\caption{Continuous vs. discrete (sampling and quantization) domain.}
%\label{fig:contdisc}
%\end{figure}
\begin{itemize}
\item \textbf{Parameter tuning:} in the continuous domain, PDE-based approaches typically require setting a step size during the optimization procedure. More formally, in the PDE community, it is stated that the Euler-Lagrange equation provides a sufficient condition for the existence of a stationary point of the energy functional. Let $u$ be a differentiable labeling function in a continuous domain and   $E(u)$ be an energy functional. Then, the Euler-Lagrange equation applied to $E$ is:
\begin{align}
\frac{\partial E}{\partial u}-\frac{d}{dx}\left(\frac{\partial E}{\partial u_x}\right)-\frac{d}{dy}\left(\frac{\partial E}{\partial u_y}\right)=0
\;,
\end{align}
where $u_x$ and $u_y$ are the derivatives of $u$ 
in $x$ and $y$ directions, respectively. The minimizer of $E$ may be computed via the steady state solution of the following update equation:
\begin{align}
\frac{\partial u}{\partial t}=-\frac{\partial E}{\partial u}\;,
\end{align}
where $\partial t$ is an artificial time step size. A step size too large  leads to a non-optimal solution and numerical instability, while a step size too small increases the convergence time. 
One way to ensure numerical stability during the optimization is to place an upper bound on the time-step using the Courant-Friedrichs-Lewy (CFL) condition \citep{courant1967partial}.  Under some conditions, the optimal step sizes may be computed automatically as proposed  by \cite{pock2011diagonal}.  On the other hand, in discrete domain, graph cuts-based methods do not require such parameter tuning and have proven to be numerically stable.

Note that other parameters in the segmentation energy function, including weighting parameters to balance the energy terms (e.g. $\lambda$ in \eqref{eq0}) and hyper parameters within each energy term or objective (e.g. number of histogram bins in calculating the regional/data term) are common between continuous and discrete approaches. Setting parameters can be done based on training data (learning-based) \citep{gennert1988determining,mcintosh2007single} or based on the image content \cite{rao2010adaptive}.

\item \textbf{Termination criterion:} While graph-based methods have an exact termination criterion, finding a general-purpose
termination criteria for PDE-based methods is difficult. Strategies for stopping the optimization procedure include performing a fixed number of iterations and/or iterating until the change in the solution or energy is smaller than a predefined threshold.
\item \textbf{Metrication error:} 
Metrication error, also known as grid bias, is defined as the artifacts which appear in graph-based segmentation methods due to penalizing region boundaries only across axis aligned edges. \note{Figure \ref{fig:metrication} compares the discrete and continuous version of a max-flow algorithm. As seen in Figure \ref{fig:metrication}, the contours obtained by graph cuts are noticeably blocky in the areas with weak regional cues (weak data term), while the contours obtained by the continuous method are smooth.}
 The discrete nature of graph-based methods makes it difficult to efficiently implement a convex regularizer like total variation in the discrete domain. Metrication error can be reduced in graph-based methods by increasing the graph connectivity, e.g. \citep{boykov2003computing}, but that also increases memory usage and computation time. In contrast, within the continuous domain, there is no such limitation and  regularizers can be implemented efficiently that makes the PDE approaches free from metrication error. Note that although approaches with continuous energy formulations  do not induce metrication errors, due to the discrete nature of digital images, all continuous operations are estimated by their discrete versions in the implementation stage.
\begin{figure}[!t]
\begin{center}
\subfloat[GF]{\includegraphics[trim=0mm 0mm 0mm 0mm,width=.3\linewidth,clip=true]{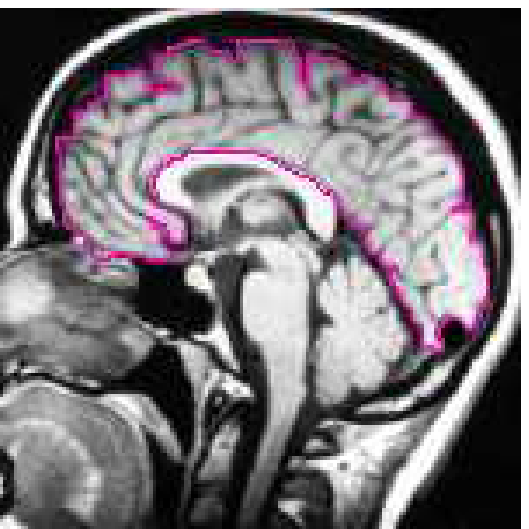}}~
\subfloat[CCMF]{\includegraphics[trim=0mm 0mm 0mm 0mm,width=.3\linewidth,clip=true]{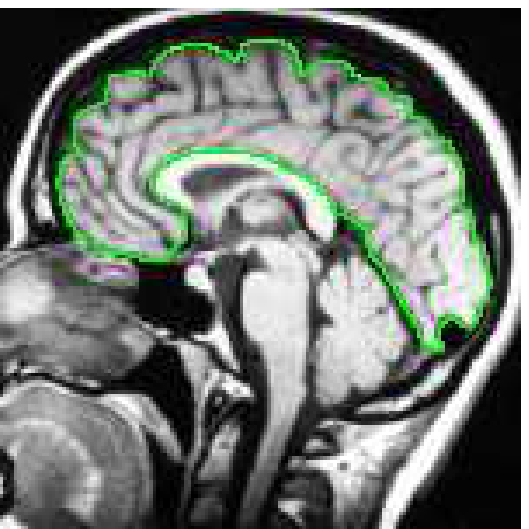}}\\
\subfloat[]{\includegraphics[trim=0mm 0mm 0mm 0mm,width=.2\linewidth,clip=true]{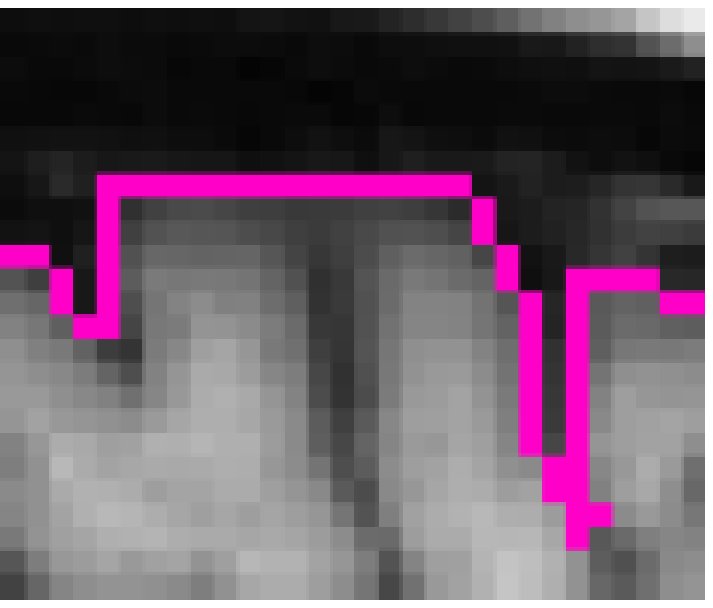}}~
\subfloat[]{\includegraphics[trim=0mm 0mm 0mm 0mm,width=.2\linewidth,clip=true]{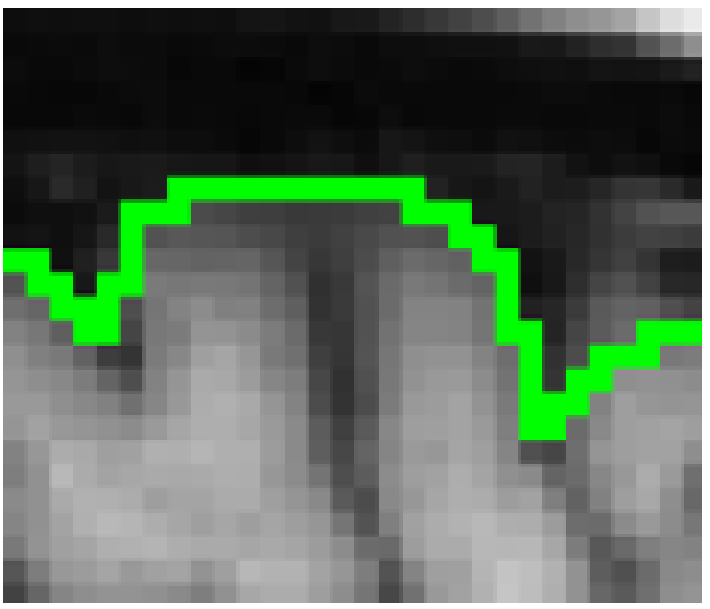}}~~~
\subfloat[]{\includegraphics[trim=0mm 0mm 0mm 0mm,width=.2\linewidth,clip=true]{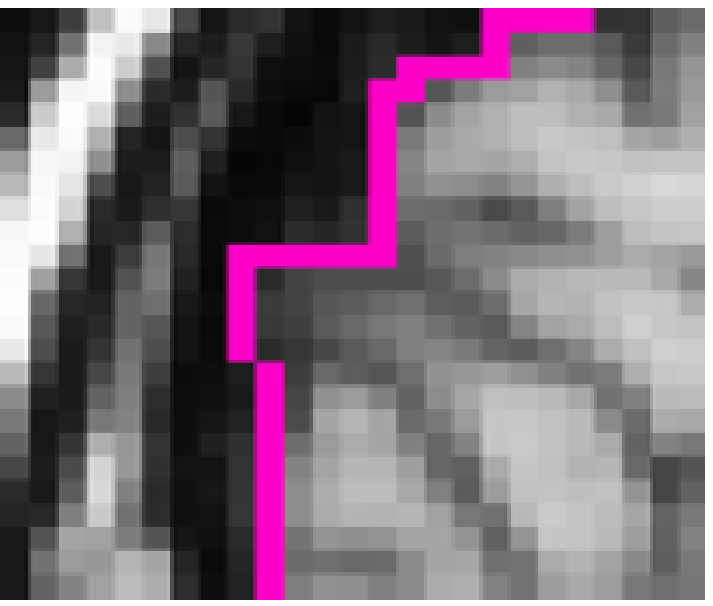}}~
\subfloat[]{\includegraphics[trim=0mm 0mm 0mm 0mm,width=.2\linewidth,clip=true]{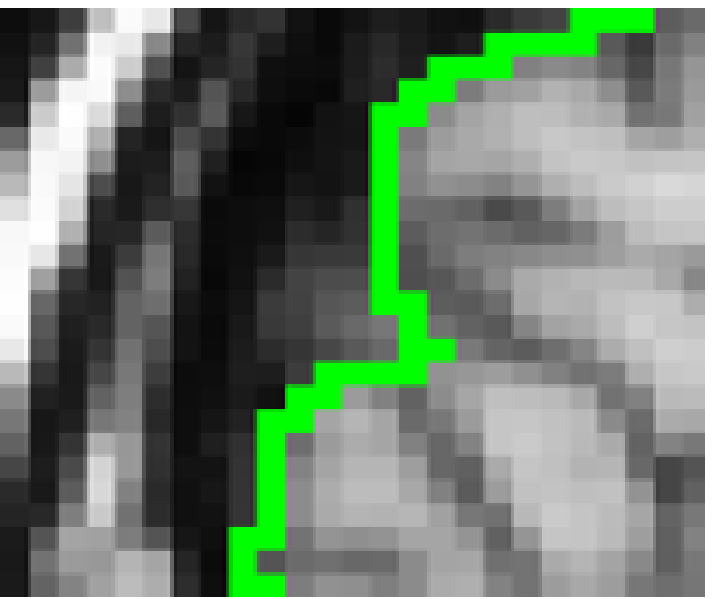}}
\caption{\note{Metrication artifacts. Brain segmentation using (a)  classical max-flow algorithm or graph cuts (GC) and (b) combinatorial continuous max-flow (CCMF) \citep{couprie2011combinatorial}.  (c,e) Zoomed regions of (a). (d,f) Zoomed regions of (b).\\
\small (Images adopted from \citep{couprie2011combinatorial})}}
\label{fig:metrication}
\end{center}
\end{figure}

\item \textbf{Parallelization:} 
Unlike PDE approaches that are easily parallelizable on GPUs, graph-based techniques are not straightforward to parallelize. As an example, the max-flow/min-cut, a core algorithm of many state-of-the-art graph-based segmentation methods, is a P-complete problem, which is probably not efficiently parallelizable \citep{goldschlager1982maximum,nieuwenhuis2013survey} due to two reasons: (1) augmenting path operations in min-cut/max-flow algorithms are interdependent as different augmentation paths can share edges; (2) the updates of the edge residuals have to be performed simultaneously in each augmentation operation as they all depend on the minimum capacity within the augmentation path \citep{nieuwenhuis2013survey}. Several attempts have focused on parallelizing the max-flow/min-cut computation. Push-relabel algorithms \citep{boykov1998markov,delong2008scalable} relaxed the first issue mentioned above but the update operations are still interdependent. Other techniques split the graph into multiple parts and obtained the global optimum by iteratively solving sub-problems in parallel \citep{strandmark2010parallel,liu2010parallel} while \cite{shekhovtsov2013distributed} combined the path augmentation and push-relabel techniques.

\item \textbf{Memory usage:} 
With respect to memory consumption, the continuous optimization methods are often the winner. While continuous methods require few floating point values for each pixel in the image, the graphical models require an explicit storage of edges as well as one
floating value for each edge. This difference becomes important when we deal with very large images and when the large number of graph edges required to be implemented, e.g. hundreds of millions pixels of microscopy images, and 3D volumes \citep{appleton2006globally}.

%Due to the graphical representation of an image, graph-based methods often consume more memory compared with methods in continuous domain due to their large connectivity between nodes. In practice, memory usage becomes important 
\item \textbf{Runtime:} The runtime variance in graph-based methods is higher than PDE-based methods. For example, considering the $\alpha$-expansion \citep{boykov2001fast} as a popular multi-label optimization technique, the number of max-flow problems that need to be solved highly depends on the input image and the chosen label order. In addition, the number of augmentation steps needed to solve a max-flow problem depends on the graph structure and edge residuals \citep{nieuwenhuis2013survey}. On the other hand, PDE-based methods have less runtime variance as they perform the same computation steps on each pixel. 
\end{itemize}

For more qualitative and quantitative comparisons between continuous and discrete domain, refer to \citep{nieuwenhuis2013survey,couprie2011combinatorial,nosrati2014local}.

\subsection{Optimization: convex (submodular) vs. non-convex (non-submodular)}
\begin{figure}
\begin{center}
\subfloat[Non-convex]{\includegraphics[trim=0mm 0mm 0mm 0mm,width=.25\linewidth,clip=true]{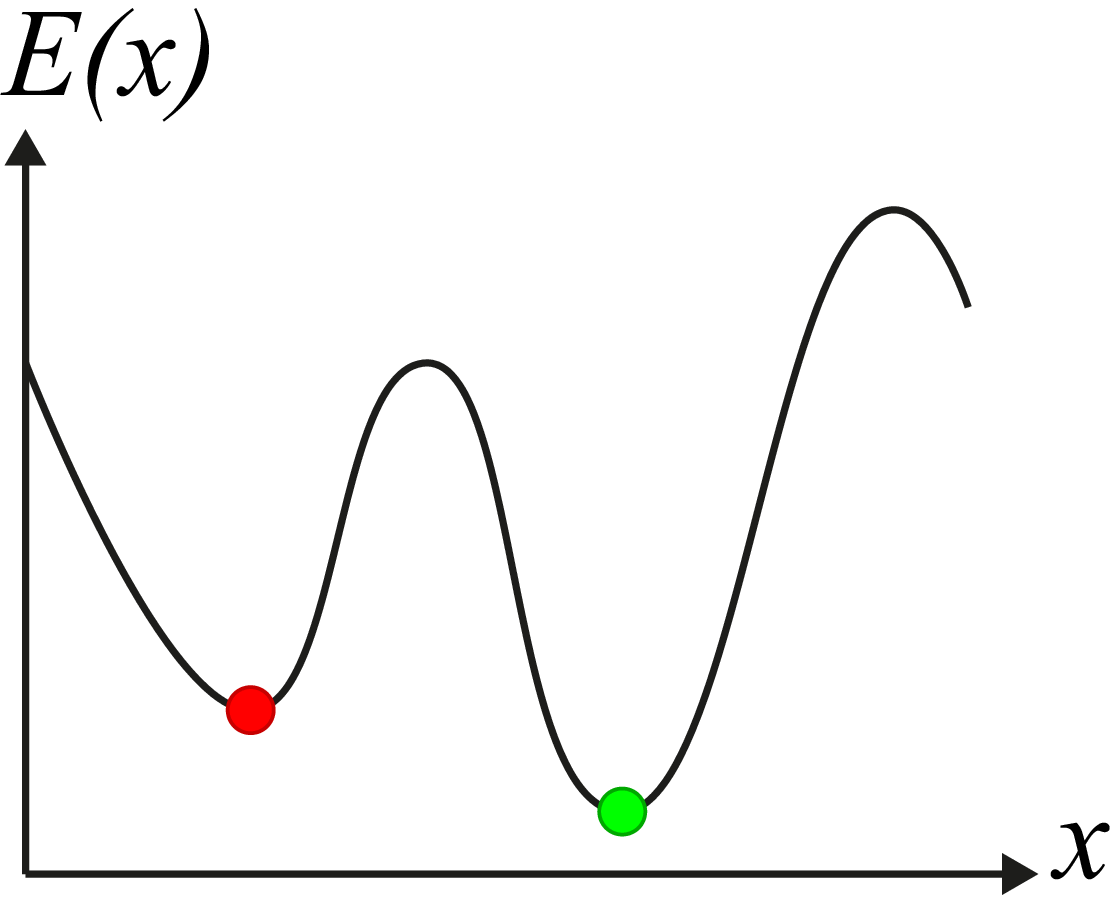}}~
\subfloat[Convex]{\includegraphics[trim=0mm 0mm 0mm 0mm,width=.25\linewidth,clip=true]{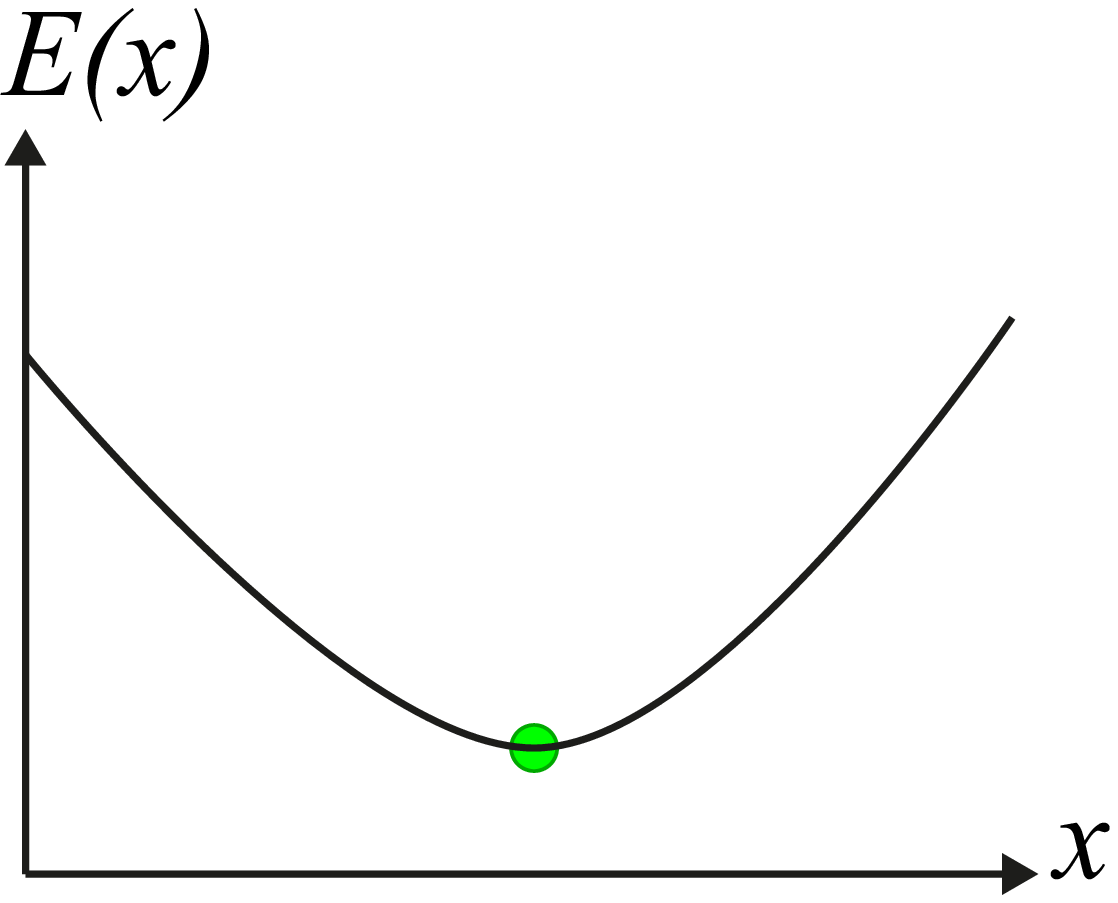}}~
\subfloat[Pseudoconvex]{\includegraphics[trim=0mm 0mm 0mm 0mm,width=.25\linewidth,clip=true]{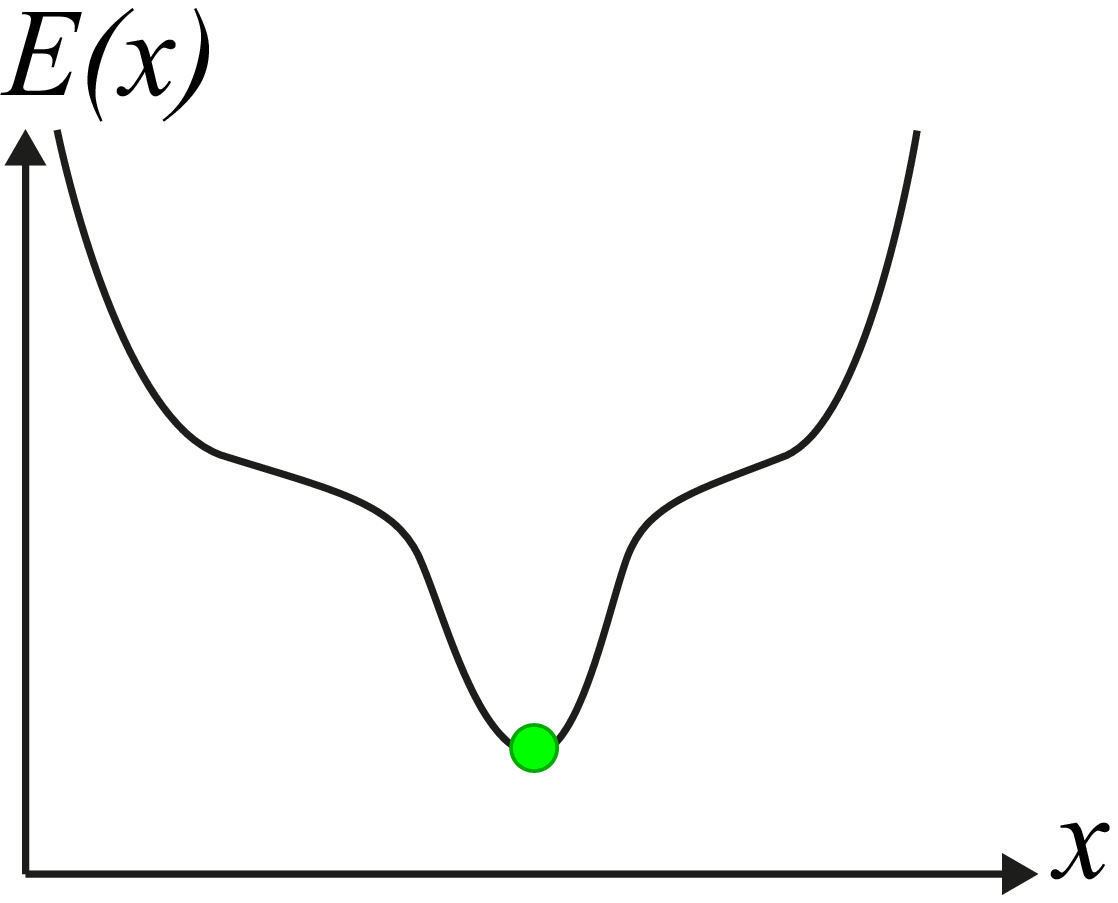}}~
\subfloat[Quasiconvex]{\includegraphics[trim=0mm 0mm 0mm 0mm,width=.25\linewidth,clip=true]{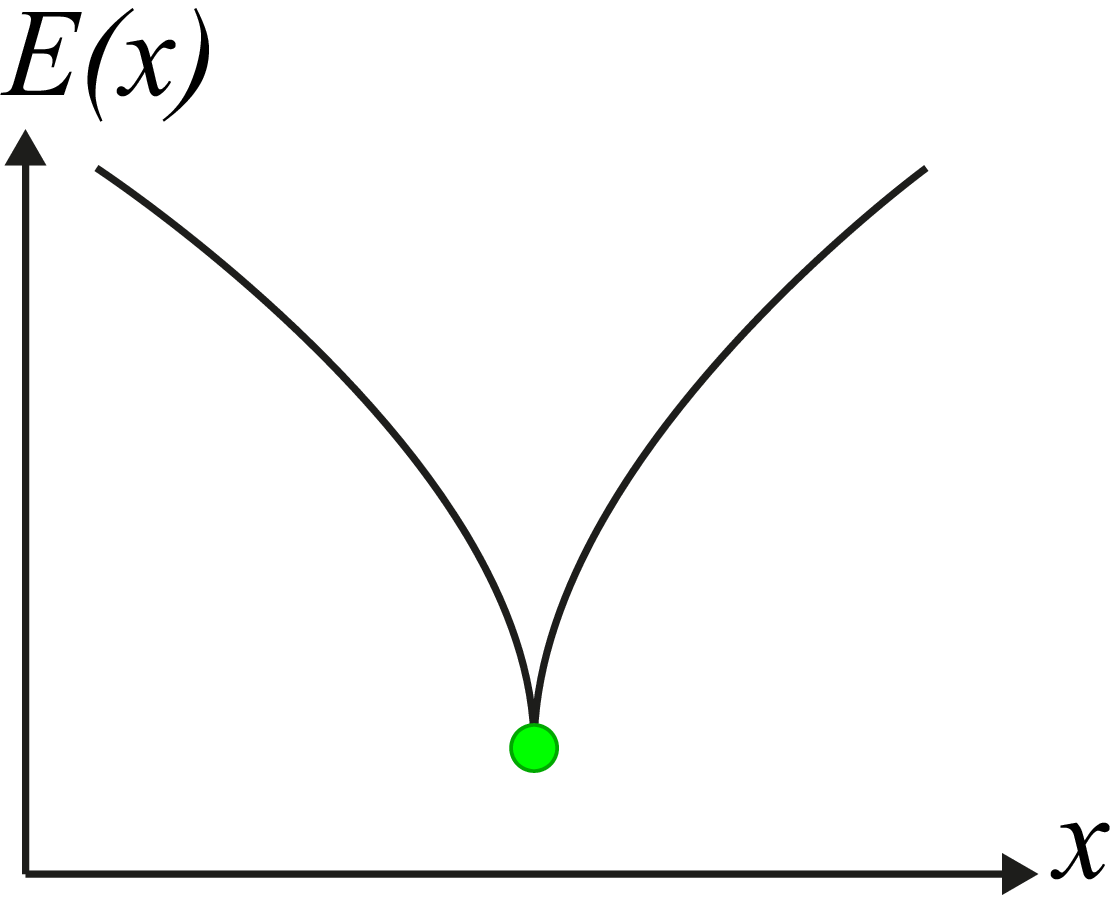}}
\end{center}
\caption{One dimensional example of a (a) non-convex, (b)    convex, (c) pseudoconvex and (d) a quasiconvex function. Red and green dots indicate the local minimum and the global solution, respectively. Red and green circles represent local and global optimum, respectively.}
\label{fig:1Dfunction}
\end{figure}

In the continuous domain of energy, a function may be classified as non-convex, convex, pseudoconvex or quasiconvex (Figure \ref{fig:1Dfunction}). Below, we define each of these terms mathematically. 

An energy function $E:\mathcal{S}\rightarrow\mathbb{R}$ is convex if
\begin{align}
&\bullet \text{the energy domain $\mathcal{S}$ (or the solution space) is a convex set and}\\ \notag
&\bullet\forall S_1,S_2\in\mathcal{S} \text{ and } 0\leq\lambda\leq 1\\ \notag
&~~~~E(\lambda S_1+ (1-\lambda)S_2)\leq \lambda E(S_1)+(1-\lambda)E(S_2). \notag
\end{align}
   A set $\mathcal{S}$ is a convex set if $S_1,S_2\in\mathcal{S}$ and $0\leq \lambda\leq 1\Rightarrow\lambda S_1+(1-\lambda)S_2 \in \mathcal{S}$. If $E$ is differentiable in $S_1\in\mathcal{S}$, $E$ is said to be pseudoconvex at $S_1$ if
   \begin{align}
   \nabla E(S_1)\cdot(S_2-S_1)\geq 0, S_2\in\Omega\Rightarrow E(S_2)\geq E(S_1).   
   \end{align}
 We call $E$ a quasiconvex function if
\begin{align}
&\bullet \text{the energy domain $\mathcal{S}$ is a convex set and}\\ \notag
& \bullet \text{the sub-level sets  $\bs{S}_\alpha=\{S\in\mathcal{S}|E(S)\leq\alpha\}$ are convex for all $\alpha$} \notag
\;.
\end{align}
Pseudoconvex functions share the property of convex functions in that, if $\nabla E(S)=0$, then $S$ is a global minimum of $E$. The pseudoconvexity is strictly weaker than convexity. In fact,  every convex function is pseudoconvex. For example, $E(S)=S+S^3$ is pseudoconvex and non-convex. Also, every pseudoconvex function is quasiconvex, but the relationship is not commutative, e.g. $E(S)=S^3$ is quasiconvex and not pseudoconvex.

%%%%%%%%%%%%%%%%%%%%%%%%%%%%%%%%%%%%%%%%
%%%%%%%%%%%%%%%%%%%%%%%%%%%%%%%%%%%%%%%%

In this paper we focus on convex and non-convex optimization problems; more details on quasiconvex problems can be found in \citep{dos1998quasiconvex}.
In the continuous domain, an optimization problem must meet two conditions to be a convex optimization problem: 1) the objective function must be convex, and 2) the feasible set must also be convex. The drawbacks associated with non-convex problems are that, in general, there is no guarantee in finding the global solution and results strongly depend on the initial guess/initialization. In contrast, for a convex problem, a local minimizer is actually a global minimizer and results are independent of the initialization. However, non-convex energy functional often give more accurate models (see Section  \ref{sec:FidOpt}).

The corresponding terminologies for convex and non-convex problems in the discrete domain are submodular and non-submodular (supermodular) problems, respectively. 
%A function $f:2^\Omega\rightarrow\mathbb{R}$ is submodular if it satisfies the following condition: $f(x)+f(y)\geq f(x\cup y)+f(x\cap y)$, where $2^\Omega$ denotes the power set of $\Omega$ and $x$ and $y$ are subset of $\mathcal{P}$
Let $E$ be a function of $n$ binary variables and $E(f_1,...,f_n)=\sum_i E^i(f_i)+\sum_{i<j}E^{ij}(f_i,f_j)$. Then the discrete energy functional $E$ is  submodular if the following condition holds:
\begin{align}
E^{ij}(0,0)+E^{ij}(1,1)<E^{ij}(0,1)+E^{ij}(1,0).
\end{align}
For higher order energy terms, e.g. $E^{ijk}(f_i,f_j,f_k)$, $E$ is submodular if all \emph{projections} \footnote{Suppose $E$ has $n$ binary variables. If
$m<n$ of these variables are fixed, then we get a new function $E'$ of $n-m$ binary variables; $E'$ is called a projection of $E$.} 
of $E$ of two variables are submodular \citep{kolmogorov2004energy}. %To define \emph{projection}, suppose $E$ has $n$ binary variables. If $m<n$ of these variables are fixed, then we get a new function $E'$ of $n-m$ binary variables; $E'$ is the projection of $E$.

Submodular energies can be optimized efficiently via graph cuts.  \cite{greig1989exact} were the first to utilize min-cut/max-flow algorithms to find the globally optimal solution for binary segmentation in 1989. Later in 2003,  \cite{ishikawa2003exact} generalized the graph cut technique to find the exact solution for a special class of multi-label problems (more detail on Ishikawa's approach in Section \ref{sec:multilabel}).

In recent years, many efforts have been made to bridge the gap between convex and non-convex optimization problems in the continuous domain through convex approximations of non-convex models. Historically, the two-region segmentation problem (foreground and background) was convexified in 2006 by \cite{chan2006algorithms} and the multi-region segmentation problem was convexified in 2008 by  \cite{chambolle2008convex} and  \cite{pock2008convex} for the first time (additional details on continuous multi-region segmentation problem in Section \ref{sec:multilabel}).

\subsection{Fidelity vs. Optimizibility }
\label{sec:FidOpt}
In energy-based segmentation problems there is a trade-off between \emph{fidelity} and \emph{optimizability} \citep{scia2011_g4a2011,mcintosh2012medial,ulen2013efficient,nosrati2014local}. Fidelity describes how faithful the energy function is to the data and how accurate it can model and capture intricate problem details. Optimizability refers to  how easily we can optimize the objective function and attain the global optimum. 

%%%%%%%%%%%%%%%%%%%%%%%%%%%%%%%%%%%%%%%%
\begin{figure}
\begin{center}
{\includegraphics[trim=0mm 0mm 0mm 0mm,width=.7\linewidth,clip=true]{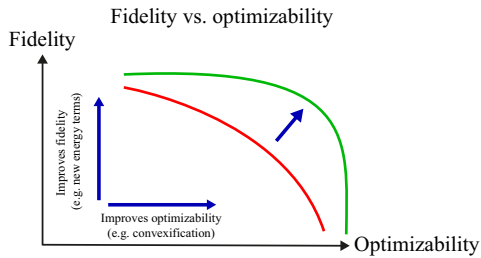}}
\end{center}
\caption{Fidelity vs. optimizibility. Ideally, an energy function is designed in a way that is faithful to the underlying segmentation problem and, at the same time,  easy to be optimized. }
\label{fig:fidVsopt}
\end{figure}
%%%%%%%%%%%%%%%%%%%%%%%%%%%%%%%%%%%%%%%%

Generally, the better the objective function models the problem, the more complicated it becomes and the harder it is to optimize. If we instead sacrifice fidelity to obtain a globally optimizable objective function, the solution might not be accurate enough for our segmentation purpose. 

In the image segmentation literature, many works have focused on increasing the fidelity and improving the modeling capability of objective functions by (i) adding new energy terms, e.g. edge, region, shape, statistical overlap and area prior terms \citep{gloger2012prior, shen2011active,andrews2011convex,Bresson2006variational,pluempitiwiriyawej2005stacs,ayed2009statistical,ayed2008area}; (ii) extending binary segmentation methods to multi-label segmentation \citep{vese2002multiphase,mansouri2006multiregion,rak2013adaptive}; (iii) modeling spatial relationships between labels, objects, or object regions \citep{felzenszwalb2010tiered,liu2008graph,rother2009minimizing,colliot2006integration,gould2008multi}; and (iv) learning objective function parameters \citep{alahari2010efficient,nowozin2010parameter,szummer2008learning,mcintosh2007single,kolmogorov2007applications}. 

Other works chose to improve optimizibility by approximating non-convex energies with convex ones \citep{lellmann2009convex,bae2011global,boykov2001fast,chambolle2008convex}. 

An ideal method improves both optimizibility and fidelity  without sacrificing either property (green contour in  Figure \ref{fig:fidVsopt}).

\subsection{Uncertainty and fuzzy / probabilistic vs. crisp labelling}
\label{sec:fuzzy}
In an MIS problem,  ideally, we are interested in finding an optimal ground truth labeling for an image, where each label represents a single structure of interest. However, as medical images are approximate representations of physical tissues and due to noise coming from the internal body  structures and/or imaging devices,  it is often  difficult to precisely define a ground truth labeling. Even the manual segmentation of an image by several experts have some degree of inter-expert (different experts) and intra-expert (same expert at different times)  variability due to ambiguities in the image. Therefore, it is beneficial to encode uncertainty into segmentation frameworks   \citep{koerkamp2010uncertainty}. This information  can be used to highlight the ambiguous image regions so to prompt users' attention to confirm or manually edit the segmentation of these regions. 

\begin{figure}
\begin{center}
\subfloat[Less certain solution]{\includegraphics[trim=0mm 0mm 0mm 0mm,width=.35\linewidth,clip=true]{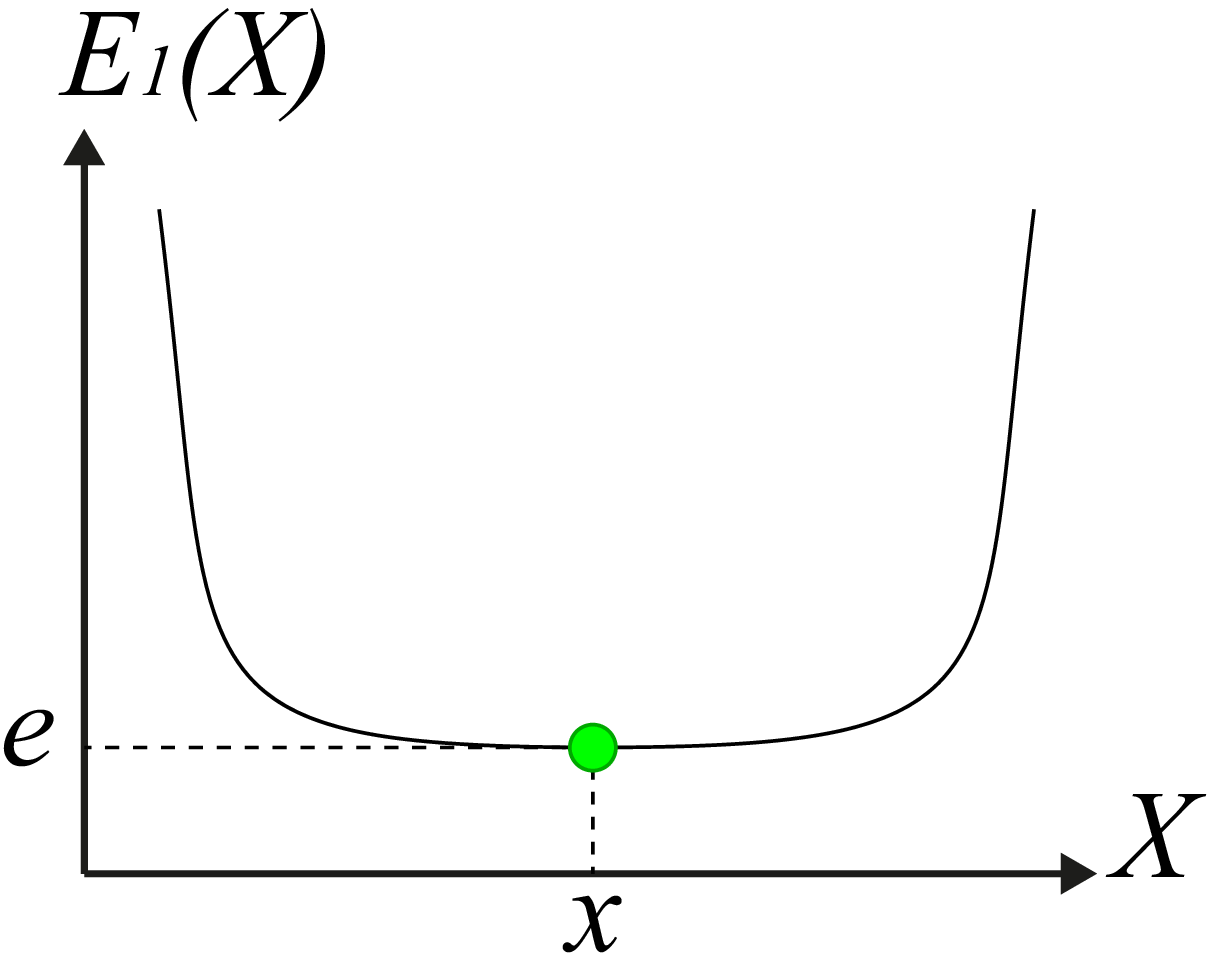}}~~~~~
\subfloat[More certain solution]{\includegraphics[trim=0mm 0mm 0mm 0mm,width=.35\linewidth,clip=true]{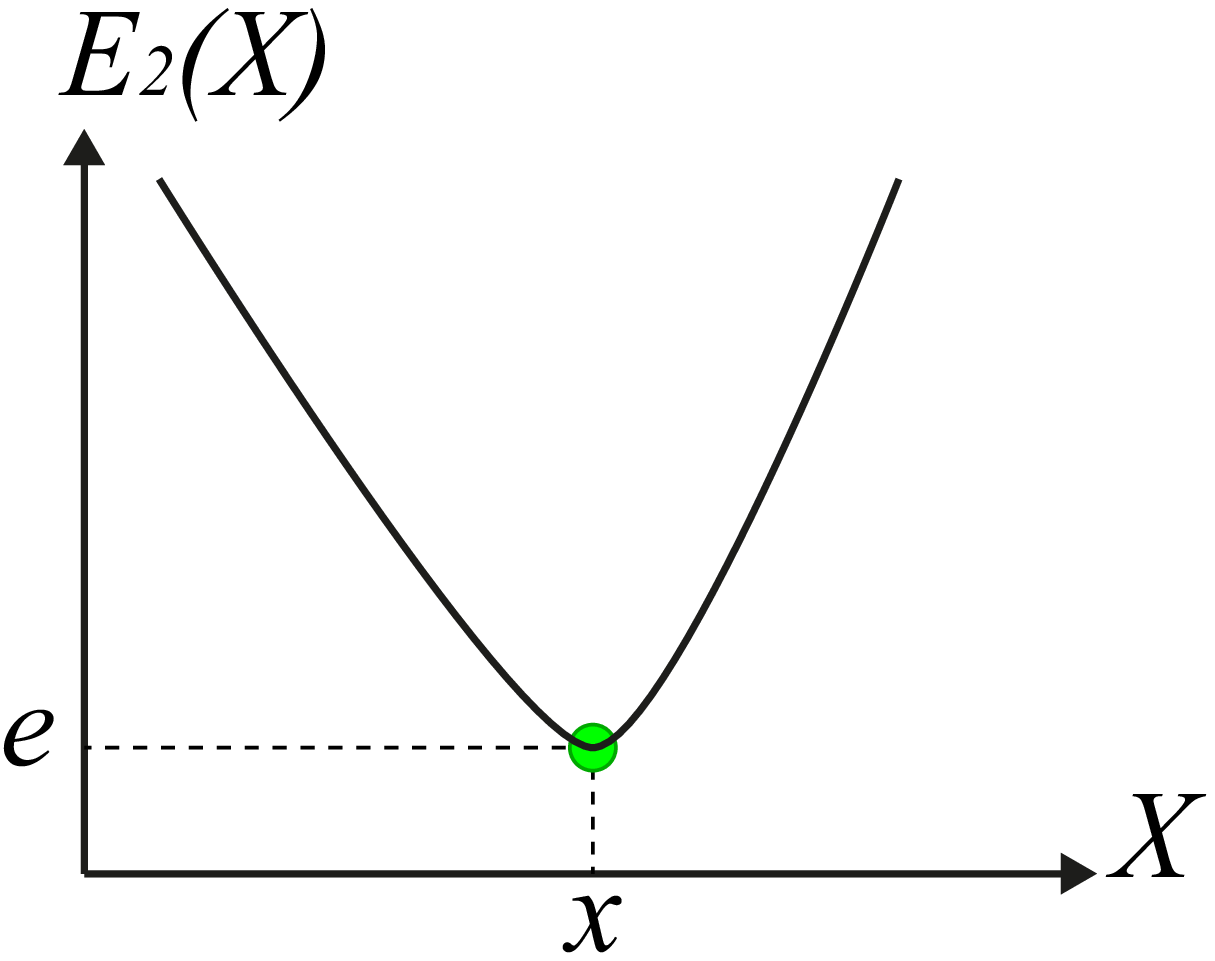}}
\end{center}
\caption{One dimensional example of two energy functions with (a) less vs. (b) more certain solutions.}
\label{fig:uncertainty}
\end{figure}

Uncertainty in object boundaries may arise from numerous sources, including graded composition \citep{udupa2005go}, image acquisition artifacts, partial volume effects. Therefore, various image segmentation methods have been intentionally designed to output probabilistic or fuzzy results to better capture uncertainty in segmentation solutions  \citep{grady2006random,zhang2001segmentation}. Figure \ref{fig:uncertainty} demonstrates an example of how uncertainty information can be observed in an energy function. $E_1$ and $E_2$ in Figure \ref{fig:uncertainty} are two 1-D energy functions with  the same optimal solution. However, segmentations near the minimal solution in $E_1$ have very similar energy values (high uncertainty) as opposed to solutions near the same optimal point in $E_2$ (less uncertainty/more certain). In fact, under the energy $E_1$, a small perturbation in the image (e.g.  additional noise) may change the segmentation result significantly. Given a probability distribution function over the label space, i.e. $P(x)$ in \eqref{eq2}, one way to calculate the uncertainty at pixel $x$ is to use Shannon's entropy as: $h(x)=-\sum P(x)\log_2(P(x))$. The entropy can be used as an energy term in a segmentation energy function. In this case, lower entropy corresponds to larger certainty and vice versa.

%%%%%%%%%%%%%%%%%%%%%%%%%%%%%%%%%%%%%%%%
%%%%%%%%%%%%%%%%%%%%%%%%%%%%%%%%%%%%%%%%

%\subsection{Fuzzy / probabilistic vs. crisp labelling}
%As mentioned in the previous section, 

As stated in Section \ref{sec:optimizationbased}, in addition to crisp labelling where each pixel is mapped to exactly one object label, two common ways to encode uncertainty into a segmentation framework are the adoption of probabilistic and fuzzy labelling. In probabilistic labelling, the probability of each label at each pixel is reported \citep{wells1996adaptive, grady2006random, saad2008kinetic, saad2010probexplorer, changizi2010probabilistic, andrews2011probabilistic, andrews2011convex}. In contrast, a partial membership of each pixel belonging to each class of labels by a membership function is reported in fuzzy labeling  \citep{bueno2004fuzzy,howing1997fuzzy}.

One important issue with probabilistic methods is that most standard techniques for statistical shape analysis (e.g. principal component analysis (PCA)) assume that the probabilistic data lie in the unconstrained real Euclidean space, which is not valid as the sample space for a probabilistic data is the unit simplex. Neglecting this unit simplex in statistical shape analysis may produce invalid shapes. \note{In fact, moving along PCA modes results in invalid probabilities that need to be projected back to the unit simplex. This projection also discards valuable uncertainty information.} To avoid the problem of generating invalid shapes,  \cite{pohl2007using} proposed a method based on the logarithm of  odds (LogOdds) transform that maps probabilistic labels to an unconstrained Euclidean vector space and its inverse maps a vector of real values (e.g. values of the signed distance map at a pixel) to a probabilistic label.  However, a shortcoming  of the LogOdds transform is that it is asymmetric in one of the labels, usually chosen as the background, and changes in this label's probability are magnified in the LogOdds space. This limitation was addressed by   \cite{changizi2010probabilistic}  and   \cite{andrews2014isometric} where the authors first use the isometric log-ratio (ILR) transformation to isometrically and bijectively map the simplex to the Euclidean real space and then analyzed the transformed data in the Euclidean real space, and finally  transformed the analysis results back to the unit simplex. More recently, \cite{andrews2015generalized} proposed a generalized log ratio transformation (GLR) that offers more refined control over the distances between different labels. 
%%%%%%%%%%%%%%%%%%%%%%%%%%%%%%%%%%%%%%%%
\begin{figure}[!t]
\begin{center}
\subfloat[Sub-pixel accuracy]{\includegraphics[trim=0mm 0mm 0mm 0mm,width=.3\linewidth,clip=true]{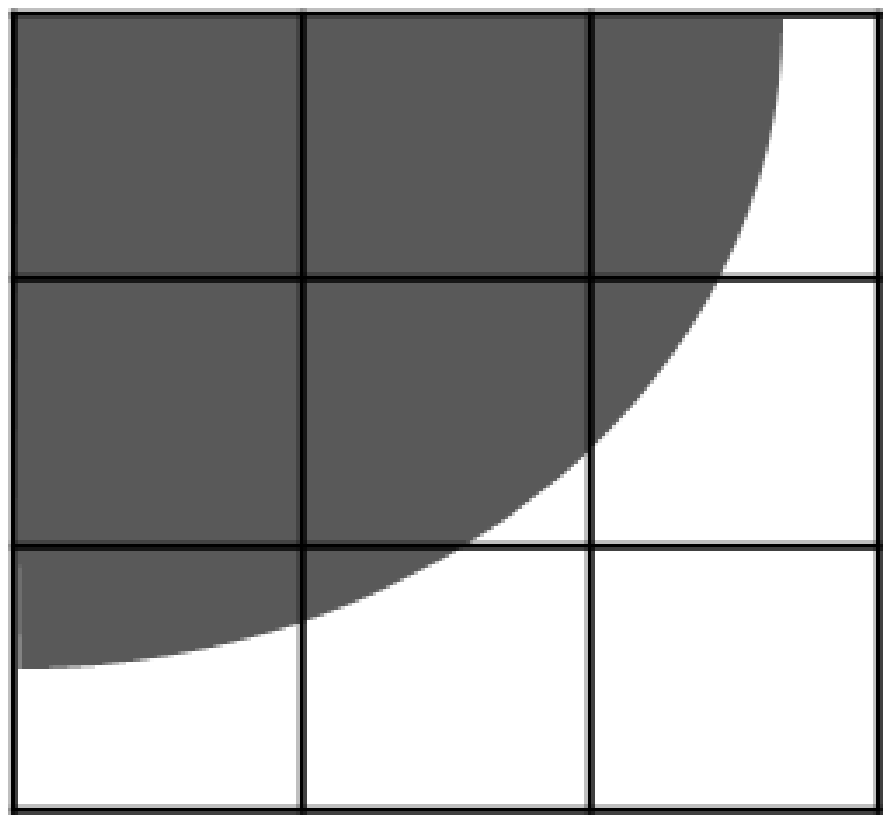}}~~
\subfloat[Grid artifact]{\includegraphics[trim=0mm 0mm 0mm 0mm,width=.3\linewidth,clip=true]{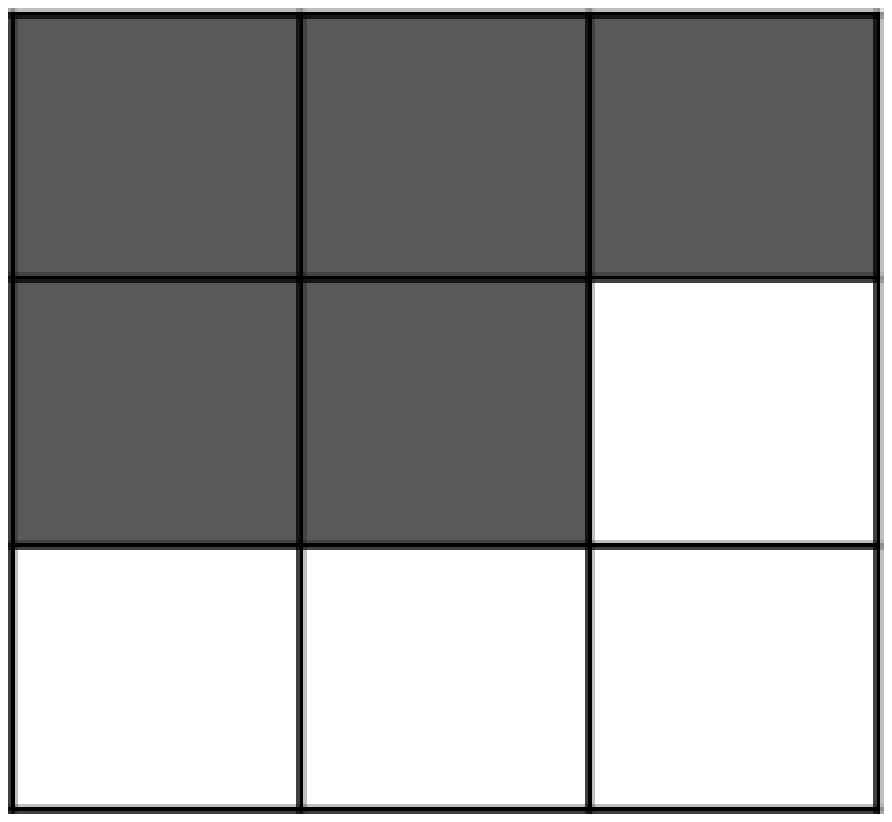}}~~
\subfloat[Fuzzy representation]{\includegraphics[trim=0mm 0mm 0mm 0mm,width=.3\linewidth,clip=true]{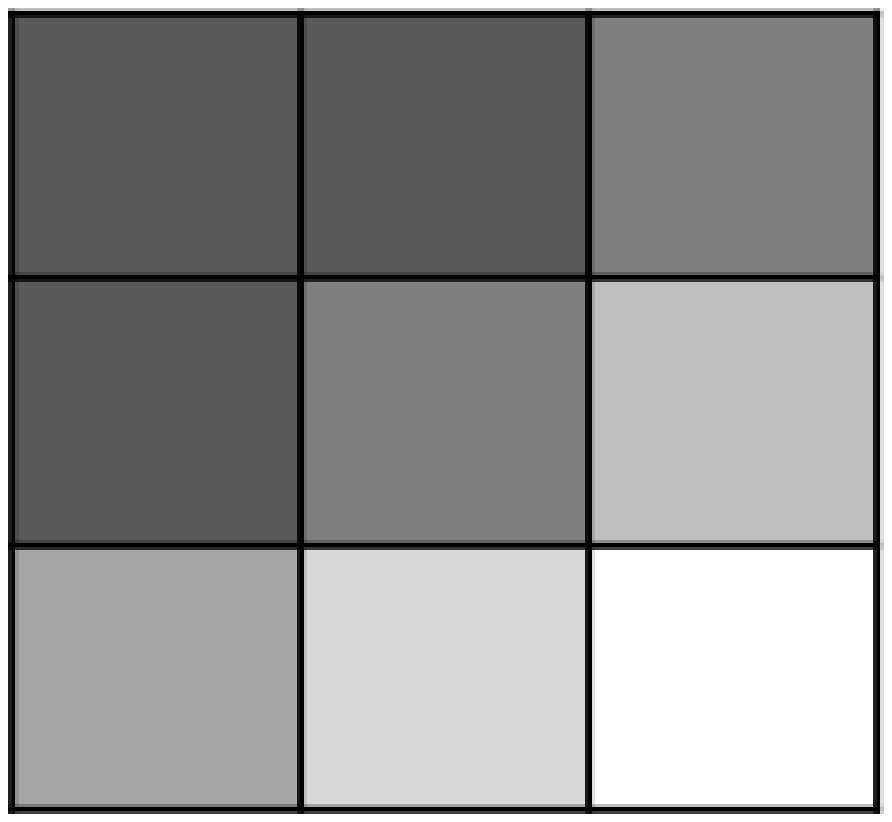}}
\end{center}
\caption{A sample segmentation (a) with and (b) without sub-pixel accuracy. (c) Representing sub-pixel accuracy using a fuzzy representation.}
\label{fig:subpixel}
\end{figure}
%%%%%%%%%%%%%%%%%%%%%%%%%%%%%%%%%%%%%%%%

\subsection{Sub-pixel accuracy}
In the spatially discrete setting, objects are converted into a discrete graph. This discretization causes loss of spatial information, which causes the object boundaries to align with the axes or graph edges as demonstrated in Figure \ref{fig:subpixel}(b). In contrast, the continuous domain does not have such shortcoming. In other words, sub-pixel accuracy allows for assigning a label to one part of a pixel and another label to the other part. This sub-pixel label assignment causes the segmentation accuracy to exceed the nominal pixel resolution of the image (Figure \ref{fig:subpixel}(a)). However, as images are digitalized in computers, the accuracy of a crisp segmentation is always limited to the image pixel resolution. One way to achieve sub-pixel accuracy is to use a fuzzy representation (see Section \ref{sec:fuzzy}) where at each pixel, its degree of membership to a label is proportional to the area covered by that label (Figure \ref{fig:subpixel}(c)).

We should emphasize that although in the continuous domain, image representation and energy formulations are continuous (Figure \ref{fig:contdiscEnergy}(a) and Figure \ref{fig:contdisc}(a)), implementation of these methods for image processing involves a discretization step (e.g. estimating a derivative by discrete forward difference). However, while the values of labels are discrete (e.g. integer values) in the discrete settings, label values in the continuous setting can be real-valued. Nevertheless, from the theoretical point of view,  continuous models correspond to the limit of infinitely fine discretization.

\section{Prior knowledge for targeted image segmentation } 
\label{sec:prior}
\begin{figure*}
    \centering
    {\includegraphics[trim=0mm 0mm 0mm 0mm,width=1\linewidth,clip=true,angle=0]{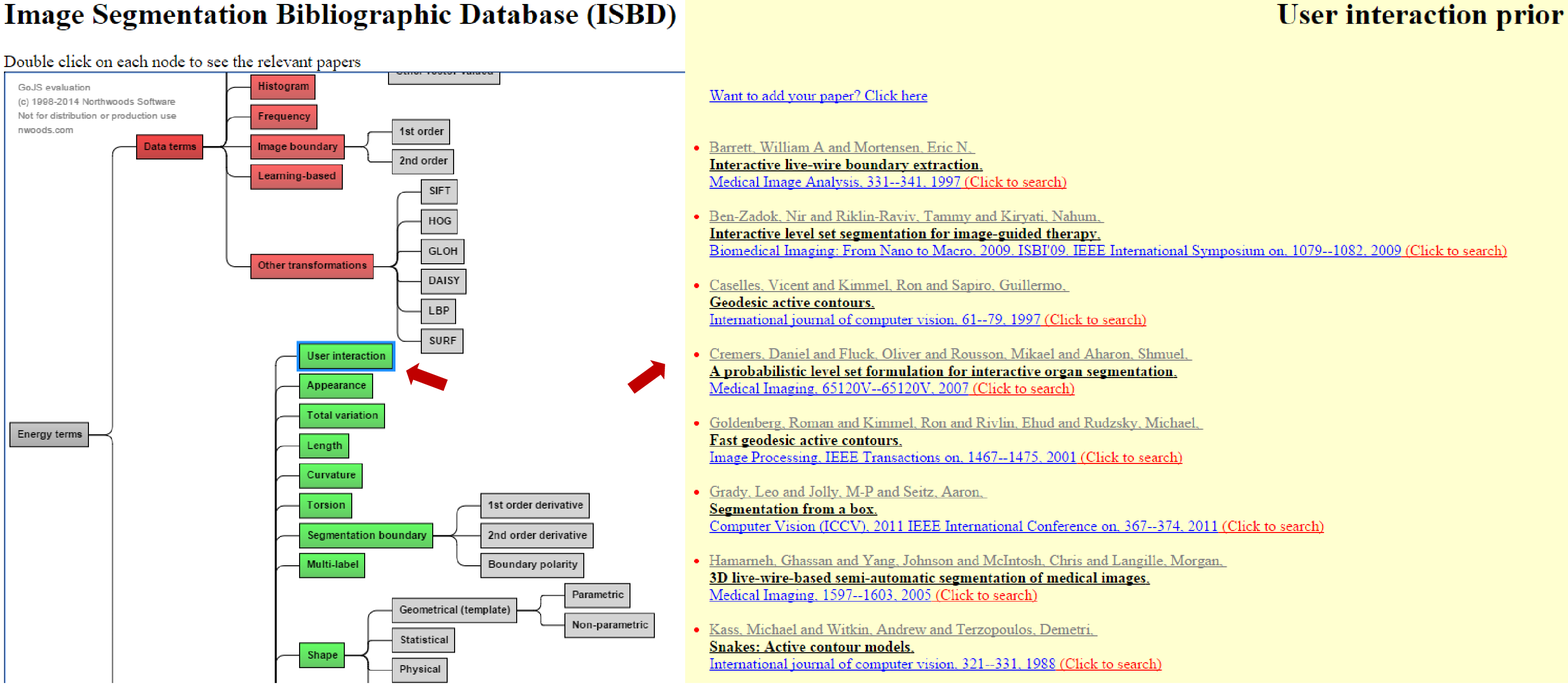}}
    \caption{Snapshot of our interactive online database of segmentation articles categorized by type of prior information devised in their framework (\url{http://goo.gl/gy9pyn}). Our online system allows users to update the records to ensure an up-to-date database.}
	\label{fig:tree2}
\end{figure*}

In this section, we review the prior knowledge information devised to improve image segmentation.  Table \ref{table1} presents some of these important priors  and compares them in terms of the nature of achievable solution due to a given formulation (i.e. globally vs. locally optimal),  metrication error, domain of action (continuous vs. discrete), and other properties. We also created an interactive online database to categorize existing works based on the type of prior knowledge they use. We made our website  interactive so that researchers can contribute to keep the database up to date. Figure \ref{fig:tree2} illustrates a snapshot of our online database showing different prior information that have been used in the literature for targeted image segmentation.
%%%%%%%%%%%%%%%%%%%%%%%%%%%%%%%%%%%%%%%%%%%%%%%%%%%

%\afterpage{%
 %   \clearpage% Flush earlier floats (otherwise order might not be correct)
  %  \thispagestyle{empty}% empty page style (?)
   % \begin{landscape}% Landscape page
 
%\begin{sidewaystable}
\begin{table*}
\footnotesize
\caption{Some important prior information for targeted image segmentation}\label{table1}\vspace{-3mm}
\begin{center}
%\begin{tabular}{"c"c"c"c"c"c"c"c|c|c"c|c|c"c|c|c"c"c"}
\begin{tabular}{|c|c|c|c!{\vrule width 2pt}c|c|c!{\vrule width 2pt}c|c|c!{\vrule width 2pt}c|c|c!{\vrule width 2pt}c|c|c||c|c|}
\hline
\textbf{Method}&
\textbf{\begin{sideways}Multi-object\end{sideways}}&
\textbf{\begin{sideways}Shape\end{sideways}}&
\textbf{\begin{sideways}Topology\end{sideways}}&
\multicolumn{3}{ c!{\vrule width 2pt} }{\textbf{\begin{sideways}Moments\end{sideways}}}&
\multicolumn{3}{ c!{\vrule width 2pt} } {\textbf{\begin{sideways}Geometrical/region interaction\end{sideways}}}&
\multicolumn{3}{ c!{\vrule width 2pt}  } {\textbf{\begin{sideways}Spatial distance\end{sideways}}}&
\textbf{\begin{sideways}Adjacency\end{sideways}}&
\textbf{\begin{sideways}No. of regions/labels\end{sideways}}&
\textbf{\begin{sideways}Model/Atlas\end{sideways}}&
\textbf{\begin{sideways}No grid artifact\end{sideways}}&
\textbf{\begin{sideways}Guarantees on global solution\end{sideways}}
%\textbf{\begin{sideways}Containment\end{sideways}}&
%\textbf{\begin{sideways}Exclusion\end{sideways}}&
%\textbf{\begin{sideways}Attraction\end{sideways}}&
%\textbf{\begin{sideways}Min. distance~~\end{sideways}}&
%\textbf{\begin{sideways}Max. distance~~\end{sideways}}&
%\multicolumn{3}{c|}{
%\textbf{\begin{sideways}Continuous\end{sideways}}
%\textbf{\begin{sideways}framework\end{sideways}}
\\ \cline{5-13}
%%%%%%%%%%%%%%%%%%%%%%%%%%%%%%%%%%%%%%%%%%%%%
&&&
&\begin{sideways}$0^{th}$ (Size/Area/Volume)~~ \end{sideways}
&\begin{sideways}$1^{st}$ (Location/Centroid) \end{sideways}
&\begin{sideways}Higher orders \end{sideways}
&\begin{sideways}Containment \end{sideways}
&\begin{sideways}Exclusion \end{sideways}
&\begin{sideways}Relative position \end{sideways}
&\begin{sideways}min. \end{sideways}
&\begin{sideways}max. (Centroid) \end{sideways}
&\begin{sideways}Attraction \end{sideways}
&&&&&\\ \hline
%%%%%%%%%%%%%%%%%%%%%%%%%%%%%%%%%%%%%%%%%%%%%
\citep{cootes1995active}& &  & 
&  & & 
&  &  &
&  &  &  
&  &  &  
&  &  \\
%%%%
\citep{cootes1995combining}& &  & 
&  & & 
&  &  &
&  &  &  
&  &  &  
&  &  \\
%%%%
\citep{rousson2002shape}&\tickno & \tickyes & \tickno
& \tickno &\tickno &\tickno 
& \tickno & \tickno &\tickno
& \tickno & \tickno & \tickno 
& \tickno & \tickno & \tickyes 
& \tickyes & \tickno \\
%%%%
\citep{chen2002using}& &  & 
&  & & 
&  &  &
&  &  &  
&  &  &  
&  &  \\
%%%%
\citep{tsai2003shape}& &  & 
&  & & 
&  &  &
&  &  &  
&  &  &  
&  &  \\ \hline
%%%%
\citep{slabaugh2005graph}& &  & 
&  & & 
&  &  &
&  &  &  
&  &  &  
&  &  \\
%%%%
\citep{zhu2007graph}&\tickno & \tickyes & \tickno
& \tickno &\tickno &\tickno 
& \tickno & \tickno &\tickno
& \tickno & \tickno & \tickno 
& \tickno & \tickno & \tickyes 
& \tickno & \tickno \\ \hline
%%%%
\citep{veksler2008star}& \tickno & \tickyes & \tickno
& \tickno &\tickno &\tickno 
& \tickno & \tickno &\tickno
& \tickno & \tickno & \tickno 
& \tickno & \tickno & \tickno 
& \tickno & \tickyes \\ \hline
%%%%
\citep{song2010simultaneous}& \tickyes & \tickyes & \tickno
& \tickno &\tickno &\tickno 
& \tickno & \tickno &\tickyes
& \tickno & \tickno & \tickno 
& \tickno & \tickno & \tickyes 
& \tickno & \tickyes \\ \hline
%%%%
\citep{andrews2011convex}& \tickyes & \tickyes & \tickno
& \tickno &\tickno &\tickno 
& \tickno & \tickno &\tickyes
& \tickno & \tickno & \tickno 
& \tickno & \tickno & \tickyes 
& \tickyes & \tickyes \\ \hline
%%%%
\citep{han2003topology}&  & &
&  & & 
&  &  &
&  &  &  
&  &  &  
&  & \\ 
%%%%
\citep{zeng2008topology}&\tickno & \tickno & \tickyes
& \tickno &\tickno &\tickno 
& \tickyes & \tickno &\tickno
& \tickno & \tickno & \tickno 
& \tickno & \tickno & \tickno 
& \tickno & \tickyes \\ \hline
%%%%
\citep{vicente2008graph}&\tickno & \tickno & \tickyes
& \tickno &\tickno &\tickno 
& \tickno & \tickno &\tickno
& \tickno & \tickno & \tickno 
& \tickno & \tickno & \tickno 
& \tickno & \tickno \\ \hline
%%%%
\citep{foulonneau2006affine}&\tickno & \tickyes & \tickno
& \tickyes &\tickyes &\tickyes 
& \tickno & \tickno &\tickno
& \tickno & \tickno & \tickno 
& \tickno & \tickno & \tickno 
& \tickyes & \tickyes \\ \hline
%%%%
\citep{ayed2008area}&\tickno & \tickno & \tickno
& \tickyes &\tickno &\tickno 
& \tickno & \tickno &\tickno
& \tickno & \tickno & \tickno 
& \tickno & \tickno & \tickno 
& \tickyes & \tickno \\ \hline
%%%%
\citep{klodt2011convex}&\tickno & \tickno & \tickno
& \tickyes &\tickyes &\tickyes 
& \tickno & \tickno &\tickno
& \tickno & \tickno & \tickno 
& \tickno & \tickno & \tickno 
& \tickyes & \tickyes \\ \hline
%%%%
\citep{limconstrained}&\tickno & \tickno & \tickno
& \tickyes &\tickyes &\tickyes 
& \tickno & \tickno &\tickno
& \tickno & \tickno & \tickno 
& \tickno & \tickno & \tickno 
& \tickno & \tickno \\ \hline
%%%%
\citep{wu2011region}&\tickno & \tickno & 
& \tickno &\tickno &\tickno 
& \tickyes & \tickno &\tickno
& \tickyes & \tickyes & \tickyes 
& \tickno & \tickno & \tickno 
& \tickno & \tickyes \\ \hline
%%%%
\citep{zhao1996variational}& \tickyes & \tickno & \tickno & \tickno &\tickno &\tickno 
& \tickno & \tickyes &\tickno 
& \tickno & \tickno & \tickno 
& \tickno & \tickno & \tickno 
& \tickyes & \tickyes \\ \hline
%%%%
\citep{samson2000level}& \tickyes & \tickno & \tickno 
&\tickno &\tickno &\tickno 
&\tickno & \tickyes &\tickno 
& \tickno & \tickno & \tickno 
& \tickno & \tickno & \tickno 
& \tickyes & \tickno \\ \hline
%%%%
\citep{li2006optimal}& \tickyes & \tickno & \tickno &\tickno &\tickno &\tickno 
&\tickyes & \tickyes &\tickno 
& \tickyes & \tickyes & \tickyes 
& \tickno & \tickno & \tickno 
& \tickno & \tickyes \\ \hline
%%%%
\citep{zeng1998volumetric}&\tickno & \tickno & \tickno &\tickno &\tickno &\tickno 
&\tickyes & \tickno &\tickno 
& \tickyes & \tickyes & \tickyes 
& \tickno & \tickno & \tickno 
& \tickyes & \tickno \\ \hline
%%%%
\citep{goldenberg2002cortex} & \tickno & \tickno & \tickno 
&\tickno &\tickno &\tickno 
&\tickyes & \tickno &\tickno 
& \tickyes & \tickyes & \tickyes 
& \tickno & \tickno & \tickno 
& \tickyes & \tickno \\ \hline
%%%%
\citep{paragios2002variational} & \tickno & \tickno & \tickno 
&\tickno &\tickno &\tickno 
&\tickyes & \tickno &\tickno 
& \tickyes & \tickyes & \tickyes 
& \tickno & \tickno & \tickno 
& \tickyes & \tickno \\ \hline
%%%%
\citep{vazquez2009multiphase}&\tickno & \tickno & \tickno 
&\tickno &\tickno &\tickno 
&\tickyes & \tickno &\tickno 
& \tickno & \tickno & \tickyes 
& \tickno & \tickno & \tickno 
& \tickyes & \tickno \\ \hline
%%%%
\citep{ukwatta2012efficient}&\tickno & \tickno & \tickno &\tickno &\tickno &\tickno 
&\tickyes & \tickno &\tickno 
& \tickyes & \tickno & \tickno 
& \tickno & \tickno & \tickno 
& \tickyes & \tickyes \\ \hline
%%%%
\citep{rajchl2012fast2}&\tickyes & \tickno & \tickno &\tickno &\tickno &\tickno 
&\tickyes & \tickyes &\tickno 
& \tickno & \tickno & \tickno 
& \tickno & \tickno & \tickno 
& \tickyes & \tickyes \\ \hline
%%%%
\citep{delong2009Globally}&\tickyes & \tickno & \tickno &\tickno &\tickno &\tickno 
&\tickyes & \tickyes &\tickno 
& \tickyes & \tickno & \tickyes 
& \tickno & \tickno & \tickno 
& \tickno & \tickyes \\ \hline
%%%%
\citep{ulen2013efficient}&\tickyes & \tickno & \tickno &\tickno &\tickno &\tickno 
&\tickyes & \tickyes &\tickno 
& \tickyes & \tickno & \tickyes 
& \tickno & \tickno & \tickno 
& \tickno & \tickyes \\ \hline
%%%%
\citep{schmidt2012hausdorff}&\tickyes & \tickno & \tickno &\tickno &\tickno &\tickno 
&\tickyes & \tickno &\tickno 
& \tickyes & \tickyes & \tickyes 
& \tickno & \tickno & \tickno 
& \tickno & \tickno \\ \hline
%%%%
\citep{nosrati2014local}&\tickyes & \tickno & \tickno &\tickno &\tickno &\tickno 
&\tickyes & \tickyes &\tickno 
& \tickyes & \tickyes & \tickyes 
& \tickno & \tickno & \tickno 
& \tickyes & \tickno \\ \hline
%%%%
\citep{nosrati2013bounded}&\tickyes & \tickno & \tickno &\tickno &\tickno &\tickno 
&\tickyes & \tickyes &\tickno 
& \tickyes & \tickno & \tickno
& \tickno & \tickno & \tickno 
& \tickyes & \tickyes \\ \hline
%%%%
\citep{nosrati2013segmentation}&\tickyes & \tickyes & \tickyes &\tickyes &\tickyes &\tickno 
&\tickyes & \tickyes &\tickyes 
& \tickyes & \tickyes & \tickno 
& \tickno & \tickno & \tickno 
& \tickno & \tickno \\ \hline
%%%%
\citep{liu2008graph}&\tickyes & \tickno & \tickno &\tickno &\tickno &\tickno 
&\tickno & \tickno &\tickno 
& \tickno & \tickno & \tickno 
& \tickyes & \tickno & \tickno 
& \tickno & \tickno \\ \hline
%%%%
\citep{felzenszwalb2010tiered}&\tickyes & \tickno & \tickno &\tickno &\tickno &\tickno 
&\tickno & \tickno &\tickno 
& \tickno & \tickno & \tickno 
& \tickyes & \tickno & \tickno 
& \tickno & \tickyes \\ \hline
%%%%
\citep{strekalovskiy2011generalized}& &  &  
& & & 
& &  & 
&  &  &  
&  &  &  
&  &  \\
%%%%
\citep{strekalovskiy2012nonmetric}&\tickyes & \tickno & \tickno 
&\tickno &\tickno &\tickno 
&\tickno & \tickno &\tickno 
& \tickno & \tickno & \tickno 
& \tickyes & \tickno & \tickno 
& \tickyes & \tickyes \\
%%%%
\citep{bergbauer2013morphological}& &  &  & & & 
& &  & 
&  &  &  
&  &  &  
&  &  \\ \hline
%%%%
\citep{zhu1996region}&  & &
&  & & 
&  &  &
&  &  &  
&  &  &  
&  & \\
%%%%
\citep{brox2006level}& \tickyes & \tickno & \tickno
& \tickno &\tickno &\tickno 
& \tickno & \tickno &\tickno
& \tickno & \tickno & \tickno 
& \tickno & \tickyes & \tickno
& \tickyes & \tickno \\ \hline
%%%%
\citep{ben2008region}&  & &
&  & & 
&  &  &
&  &  &  
&  &  &  
&  & \\ \hline
%%%%
\citep{delong2012fast}& \tickyes & \tickno & \tickno
& \tickno &\tickno &\tickno 
& \tickno & \tickno &\tickno
& \tickno & \tickno & \tickno 
& \tickno & \tickyes & \tickno
& \tickno & \tickno \\ \hline
%%%%
\citep{yuan2012continuous}& \tickyes & \tickno & \tickno
& \tickno &\tickno &\tickno 
& \tickno & \tickno &\tickno
& \tickno & \tickno & \tickno 
& \tickno & \tickyes & \tickno
& \tickyes & \tickyes \\ \hline
%%%%
\citep{iosifescu1997automated}&  &  & 
&  & & 
&  &  &
&  &  &  
&  &  & 
&  &  \\
%%%%
\citep{collins1997animal}& \tickyes & \tickyes & \tickno
& \tickno &\tickno &\tickno 
& \tickno & \tickno &\tickno
& \tickno & \tickno & \tickno 
& \tickno & \tickno & \tickyes
& \tickyes & \tickno \\
%%%%
\citep{prisacariu2012pwp3d}&  &  & 
&  & & 
&  &  &
&  &  &  
&  &  & 
&  &  \\ \hline
%%%%
\citep{sandhu2011nonrigid}&   &  & 
&  & & 
&  &  &
&  &  &  
&  &  & 
&  &  \\ 
%%%%
\citep{prisacariu2013simultaneous}& \tickno & \tickyes & \tickno
& \tickno &\tickno &\tickno 
& \tickno & \tickno &\tickno
& \tickno & \tickno & \tickno 
& \tickno & \tickno & \tickyes
& \tickyes & \tickno \\ \hline
%%%%
%%%%%%%%%%%%%%%%%%%%%%%%%%%%%%%%%%%%%%%%%%%%%
\end{tabular}
\end{center}
\end{table*}

\subsection{User interaction}
%In MIA area, no automated technique is currently reliable enough to be completely trusted by clinicians and a human expert must eventually review and correct the automated results. 
Incorporating user input into a segmentation framework may be an intuitive and easy way for the users to assist with characterizing  the desired object and obtain usable results. In an interactive segmentation system,  the user input is used to encode prior knowledge about the targeted object. The specific prior knowledge that the user is considering is unknown to the method, but only the implication of such prior knowledge (e.g. pixel $x$  must be part of the object) is passed on to the interactive algorithm.  Given a high-level intuitive user interactive system, the end-user does not need to know about the low-level underlying optimization and energy function details.

User input can be incorporated in several ways, such as through: mouse clicking (or even via eye gaze \citep{sadeghi2009hands}) and to provide seed points,  specifying the subsets of object boundary or  specifying sub-regions (bounding boxes) that contain the object of interest. The work proposed by \citep{kass1988snakes} is perhaps one of the early works to incorporate user interaction into the segmentation framework, where they enable users to  enforce spring-like forces between snake's control points to affect the energy functional and to push the snake out of a local minima into another more desirable location. 

The first form of user input (providing \emph{seeds}) involves user specifying labels of some pixels inside and outside the targeted object by mouse-clicking or brushing. This allows a user to enforce hard constraints on labeled pixels. For example, in a binary segmentation scenario in the discrete setting, one can enforce $f_p^{foreground}=1$ if $p\in foreground$ and $f_p^{background}=0$ if $p\in background$ in \eqref{discrete}. 

In  the continuous domain, \citep{paragios2003user}, \citep{cremers2007probabilistic} and \citep{ben2009interactive} have proposed level set-based methods in which a user can correct the solution interactively by clicking on incorrectly labelled pixels. Mathematically, let $\phi$ be the level set function (often is represented by the signed distance map of the foreground) where $\phi>0$ and $\phi<0$  represent inside and outside regions of the object of interest, respectively. \cite{cremers2007probabilistic} proposed to add the following user interaction term to their energy functional consisting of other data and regularization terms:
\begin{align}
E_{user}(\phi)=-\int_\Omega L(\bs{x})	\text{sign}(\phi(\bs{x}))d\bs{x}
\;,
\end{align}
where $L:\Omega\rightarrow\{-1,0,+1\}$ reflects the user input and is defined as:
\begin{align}
L(\bs{x})=
\begin{cases}
+1 & \text{if }\bs{x} \text { is marked as 'object'}\\
-1& \text{if }\bs{x} \text{ is marked as 'background'}\\
0& \text{if }\bs{x} \text{ is not marked}
\end{cases}
\; .
\end{align}
\cite{ben2009interactive} also used a similar energy functional similar to \citep{cremers2007probabilistic}. Assuming that $\{x_i\}_{i=1}^n$ denotes the set of user input, which indicates the incorrectly labelled  regions, they defined $M:\Omega\rightarrow\{0, 1\}$ as:
\begin{align}
M(z)=\sum_{i=1}^n\delta(z-x_i)
\;,
\end{align}
where $\delta$ is the Dirac delta function. The function $L:\Omega\rightarrow\mathbb{R}$ is defined as:
\begin{align}
L(\bs{x})=H(\phi(\bs{x}))+\left(1-2H(\phi(\bs{x}))\right)\int_{z\in\mathcal{N}}M(z)dz
\;,
\end{align}
where $H$ is the Heaviside step function and $\mathcal{N}$ is the neighbourhood of the coordinate $x$. $L(x)=0$ if the user's click is within the segmented region and $L(x)=1$ if the click is on the background. $L(x)=H(\phi(x))$ if $x$ is not marked. The user interaction term proposed by \cite{ben2009interactive} is then defined as:
\begin{align}
E_{user}(x)=\int_{x\in\Omega}\int_{x'\in\Omega}\big(L(x')-H(\phi(x))\big)^2K(x,x')dx'dx
\;,
\end{align}
where $K(x,x')$ is a Gaussian kernel.

Another form of user input is \emph{object boundary specification} where all or part of the object boundary is roughly specified by drawing a contour (in 2D) or initializing a surface (in 3D) around the object boundary. This form of user input is more suitable for 2D images as providing manual rough segmentations in 3D images (which is often the case in many medical image analysis problems) is not straightforward. Examples that require the user to provide an initial guess close to objects' boundary include \citep{wang2007soft} in the discrete setting, and edge-based active contours (e.g. gradient vector field (GVF)  \citep{xu1997gradient,xu1998generalized} and geodesic active contour  \citep{goldenberg2001fast}) in the continuous setting. Live-wire, proposed by \citep{barrett1997interactive}, is another effective tool for 2D segmentation that benefits from user-defined seeds on the boundary of the desired object. The 2D live-wire uses the gradient magnitude, gradient direction, and canny edge detector to build cost terms. After providing an initial seed point on the boundary of the object, live-wire calculates the local cost for each pixel starting from the provided seed and finds the minimal path between the initial seed point $p$ and the next point $q$ chosen by the user. The 2D live-wire was extended to 3D by  \cite{hamarneh20053d}.

Another form of user input, and probably the most convenient way for a user, is the \emph{sub-region specification} where a user is asked to draw a box around the targeted object. This bounding box can be provided automatically using machine learning techniques in object detection. In the discrete setting, GrabCut proposed by 
 \cite{rother2004grabcut} is one of the most well-known  methods with this kind of initialization.
 \cite{lempitsky2009image} proposed a method which shows how a bounding box is used to impose
a powerful topological prior that prevents the solution
from excessively shrinking and splitting,  and ensures that the solution is sufficiently close to each of the sides of the
bounding box. 
 \cite{grady2011segmentation} performed a user study and showed that a single box input is in fact enough for segmenting the targeted object.
In the continuous setting, this kind of user input (sub-region specification) is taken into account by methods like geodesic active contours \citep{caselles1997geodesic} \note{in which the user initializes the active contour around the object of interest.}

Similar interaction is utilized in 3D live-wire \citep{hamarneh20053d} as implemented in the TurtleSeg software\footnote{www.turtleseg.org} \citep{top2011active,top2011spotlight}. In 3D live-wire, few slices  in different orientations of a 3D volume are segmented using 2D live-wire. Then, the segmented 2D slices are used to segment the whole 3D volume by generating additional contours on new slices automatically. The new contours are obtained by calculating optimal paths connecting the points of intersection between the new slice planes and the original contours provided semi-automatically by the user.

\cite{saad2010exploration}   proposed another type of interactive image analysis in which a user is able to examine the uncertainty in the segmentation  results and improve the results, e.g. by changing the parameters of their segmentation algorithm.
For an expanded study on interaction in MIS, interested readers may refer to \citep{saad2010probexplorer,saad2010exploration}.

\subsection{Appearance prior}
\label{sec:appearance}
\begin{figure}[!t]
\centering
\subfloat[]{\includegraphics[trim=0mm 0mm 0mm 0mm,width=.5\linewidth,clip=true]{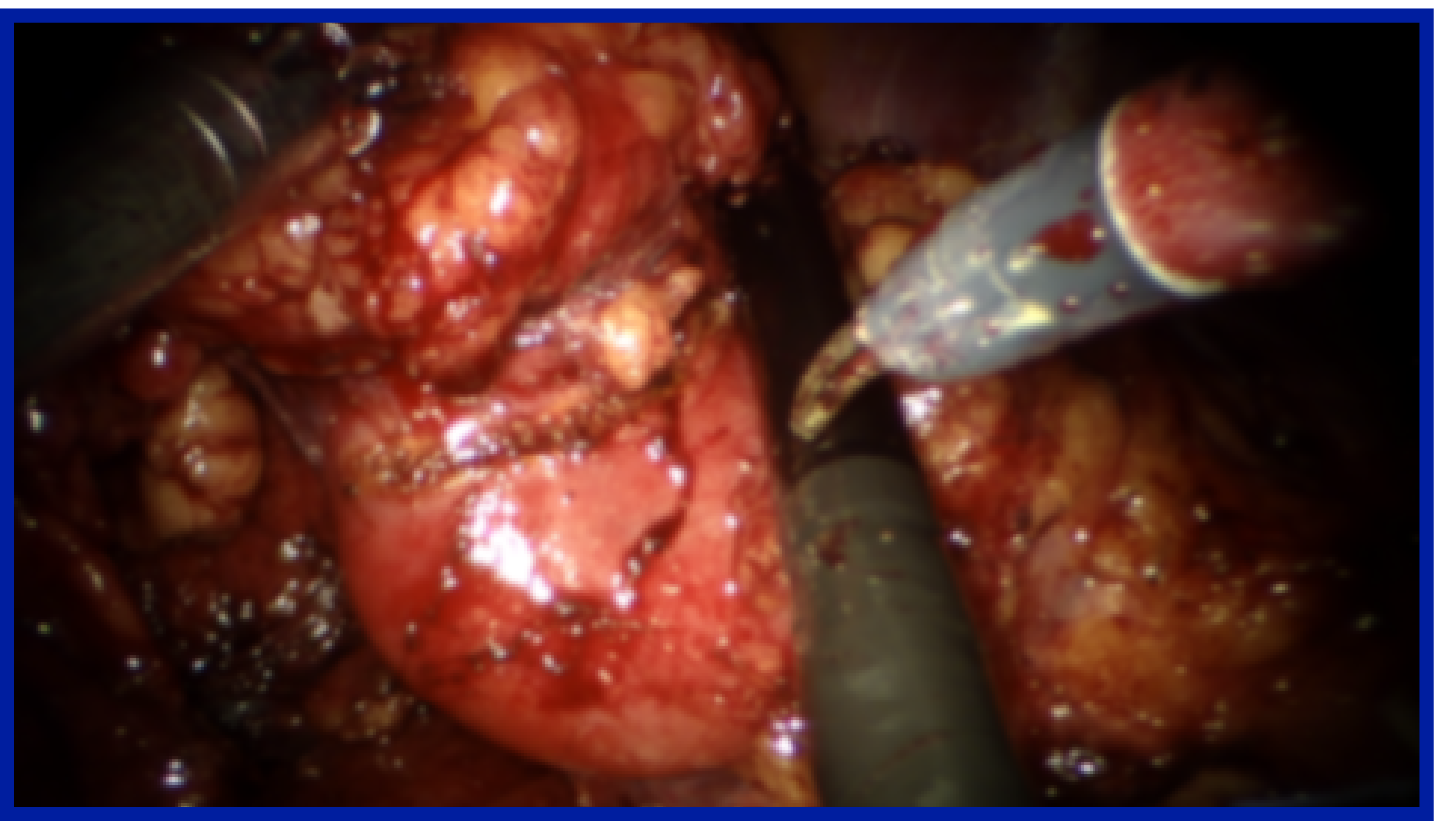}}~
\subfloat[]{\includegraphics[trim=0mm 0mm 0mm 0mm,width=.5\linewidth,clip=true]{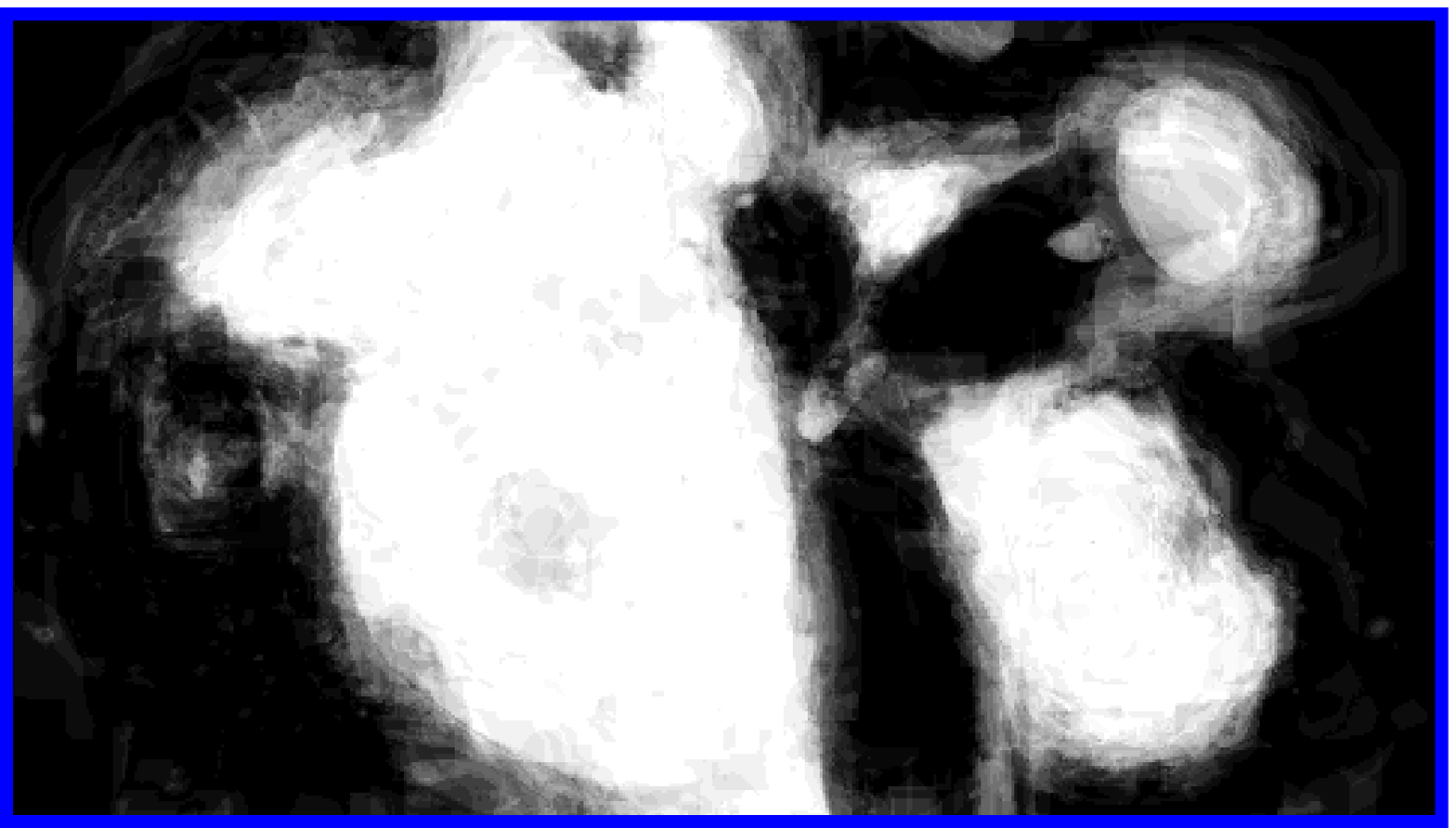}}\\
\subfloat[]{\includegraphics[trim=0mm 0mm 0mm 0mm,width=.5\linewidth,clip=true]{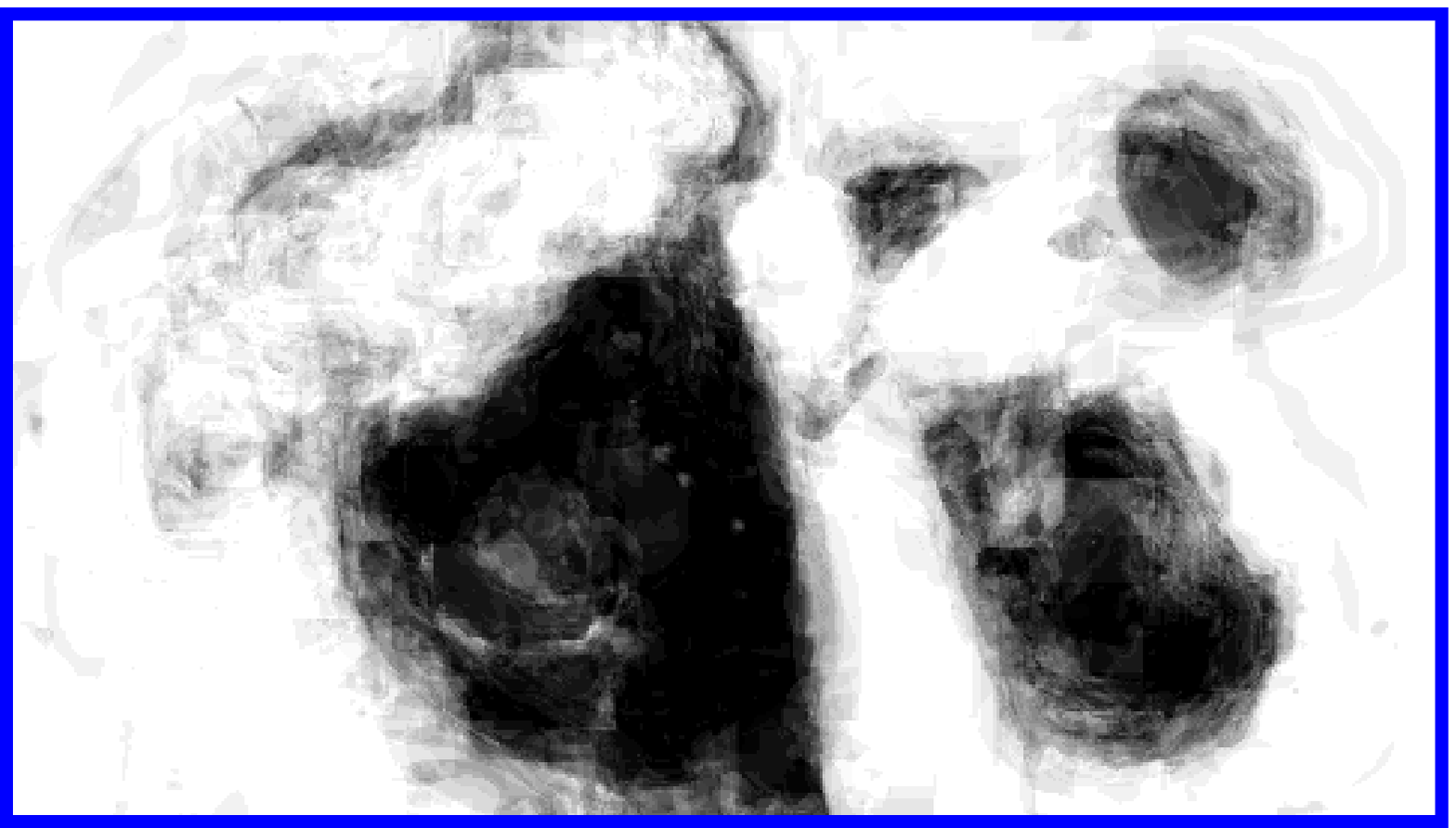}}~
\subfloat[]{\includegraphics[trim=0mm 0mm 0mm 0mm,width=.5\linewidth,clip=true]{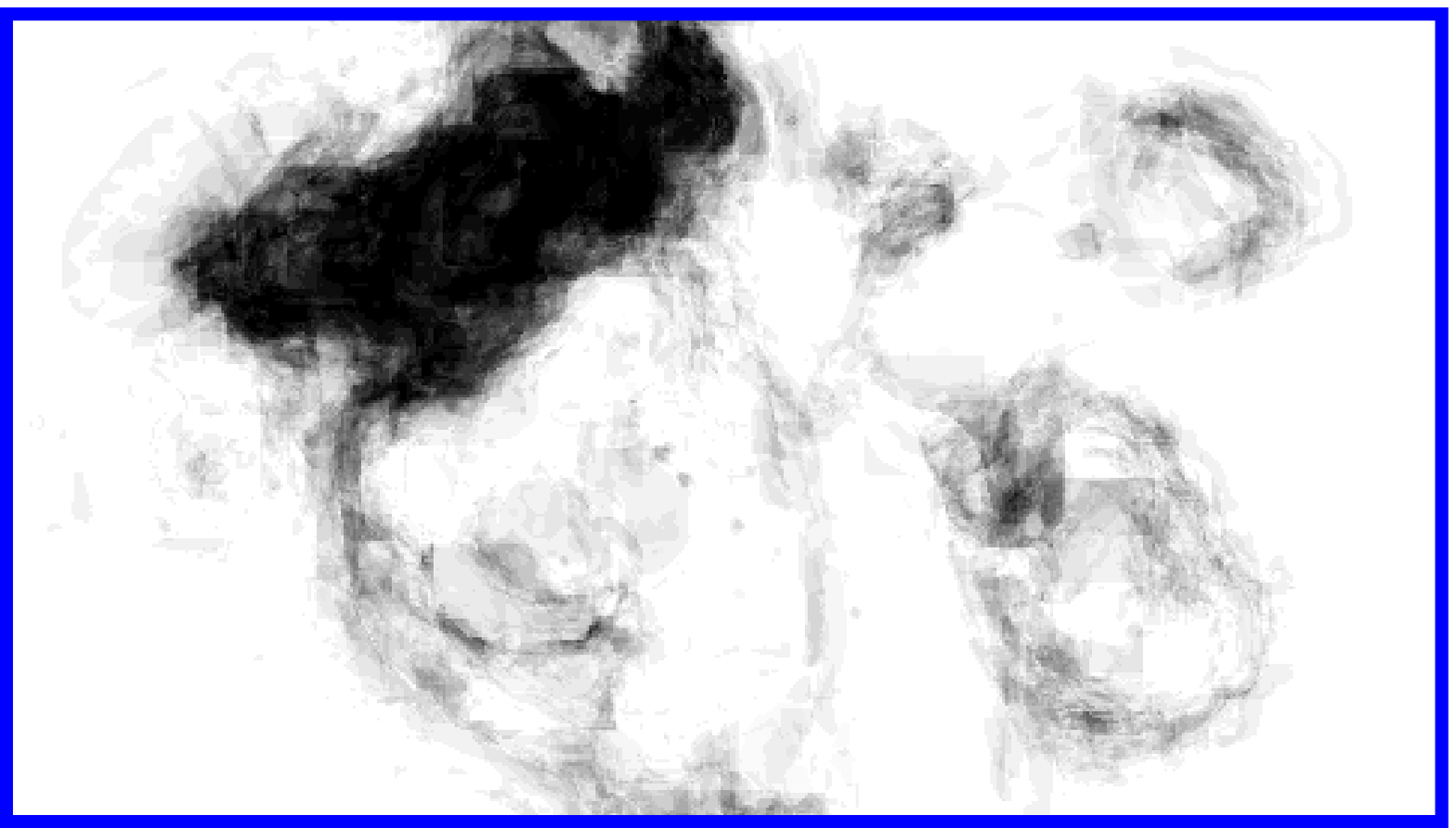}}
\caption{Examples of regional probabilities. (a) Original endoscopic image. (b-d) Probability of background, kidney and tumour for the frame shown in (a). A lower intensity in (b-d) corresponds to higher probability.}
\label{fig:click}
\end{figure}

Appearance is one of the most important visual cues to distinguish between different structures in an image. Appearance is described by studying the distribution of different features such as intensity values in gray-scale images, color, and texture inside each object. 
In most cases, appearance models are incorporated into the data term in \eqref{eq0} and \eqref{discrete}.
The purpose of incorporating appearance prior is to fit the appearance distribution of the segmented objects to the distribution of objects of interest, e.g. using Gaussian mixture model (GMM)  \citep{rother2004grabcut}. 
In the literature, there are two ways to model the appearance: 1) adaptively learning the appearance during the segmentation procedure, and 2) knowing the appearance model prior to performing segmentation (e.g. by observing the appearance distribution of the training data).
In the former case, the appearance model is learned as the segmentation is performed \citep{vese2002multiphase} (computed online).
 In the second case, it is assumed that the probability of each pixel belonging to particular label is known, i.e. if $F_i(\bs{x})$ represents a particular set of feature values (e.g. intensity/color) associated with each image location  for $i^{th}$ object, then it is assumed that $P(\bs{x}|F_i(\bs{x}))$ is known (or pre-computed offline). This probability is usually learned and estimated from the distribution of features inside small samples of each object.
  Figure \ref{fig:click} illustrates the probability of different structures (the kidney, the tumour, and the background) in an endoscopic scene.  A lower intensity in Figures \ref{fig:click}(b-d) corresponds to higher probability. 

  To fit the segmentation appearance distribution to the prior  distribution, a dissimilarity measure $d$ is usually needed where $d(g_i, \hat{g}_i)$ measures the difference between the appearance distribution of $i^{th}$ object ($g_i$) and its corresponding prior distribution $\hat{g}_i$. This dissimilarity measure can be encoded into the energy functional \eqref{eq0} directly as the data term or via a probabilistic formulation. %, i.e. maximising $P(g_i|S_i,\hat{g}_i)\propto e^{-d(g_i,\hat{g}_i)}$ assuming a Gaussian distribution for $P(g_i)$ or, equivalently, minimizing $-\log P(g_i|S_i,\hat{g}_i)$.  
  For example, consider the appearance prior of an object in a scalar-valued image $I$, then $\hat{g}_i$ would be the mean ($\mu_i$) and variance ($\sigma_i^2$) of the intensities of the targeted object. Then, assuming a Gaussian approximation of the object's intensity $I$, the corresponding probability distribution will be:
 \begin{align}
 P(\bs{x}|\hat{g}_i)=\frac{1}{\sqrt{2\pi}\sigma_i^2}e^{-\frac{(I(\bs{x})-\mu_i)^2}{2\sigma_i^2}}
 \;.
 \label{Pgi}
 \end{align}
Other than scalar-valued medical images such as MR \citep{pluempitiwiriyawej2005stacs} and US \citep{noble2006ultrasound}), appearance models can be extracted from other types of images like color image (e.g. skin \citep{celebi2009lesion}, endoscopy \citep{figueiredo2010variational}, microscopy \citep{nosrati2013segmentation}), other vector-valued images (dynamic postiron emission tomography, dPET,   \citep{saad2008kinetic}), and tensor-valued or manifold-valued images \citep{feddern2003level,wang2004affine,weldeselassie2007dt}.
For the vector-valued images, one can use multivariate Gaussain density as an appearance model. The formulation is similar to \eqref{Pgi} with the use of  the covariance matrix $\Sigma_i$ instead of $\sigma_i^2$. Regarding the tensor-valued images, several distance measures in the space of tensors have been proposed such as:
\begin{itemize}
\item \emph{Log-Euclidean tensor distance} is defined as:
\[
d_{LE}(T_i,\hat{T}_i)=\sqrt{\text{trace}\bigg(\big(\text{logm}(T_i)-\text{logm}(\hat{T}_i)\big)^2\bigg)}
,
\]
where $T_i$ and $\hat{T}_i$ are a tensor from $i^{th}$ region  and its corresponding prior tensor model, respectively.
\item The symmetrized Kulback-Leibler (SKL) divergence (also known as J-convergence) \citep{wang2004affine} is  defined as: 
\[d_{SKL}(T_i,\hat{T}_i)=\frac{1}{2}\sqrt{trace(T_i^{-1}\hat{T}_i+T_i\hat{T}_i^{-1})-2n}\;,\] where $n$ is the size of the tensor $T_i$ and $\hat{T}_i$ ($n=3$ in DT-MRI). This measure is affine invariant.
\item The Rao distance \citep{lenglet2004segmentation} is  defined as: \\ \[d_R(T_i,\hat{T}_i)=\sqrt{\frac{1}{2}\big(\log^2(\lambda_i)+\log^2(\lambda_j)\big)}\;,\] where $\lambda_i$ denotes the eigenvalues of $T_i^{-1/2}\hat{T}_iT_i^{-1/2}$.
\end{itemize}

Intensity and color information are not always sufficient to distinguish different objects. Hence, several methods proposed to model objects with more complex appearance using texture information as a complementary feature  \citep{huang2005deformable,malcolm2007graph2,santner2009interactive}.

\cite{bigun1991multidimensional} introduced a simple texture feature model consists of the Jacobian matrix convolved  by a Gaussian kernel ($K_\sigma$)  that results in three different feature channels, i.e. in case of a 2D image the features are $K_\sigma*(I_x^2,I_xI_y, I_y^2)$. However, these features ignore the non-textured object that might be of interest. Therefore, \cite{rousson2003active} proposed to use the following texture features in order to segment objects with and without texture: $(I,\frac{I_x^2}{|\nabla I|},\frac{I_xI_y}{|\nabla I|}, \frac{I_y^2}{|\nabla I|})$.

More advanced texture features such as those based on   Haar and  Gabor filter banks have shown many successes in medical image segmentation \citep{huang2005deformable,malcolm2007graph2,santner2009interactive}.
\cite{koss1999abdominal} and \cite{frangi1998multiscale} are two works that utilized advanced features to segment abdominal organs and to measure vesselness, respectively. In \citep{frangi1998multiscale}, the eigenvalues of the image Hessian matrix are used for measuring the vesselness of pixels in images. This measure is used for liver vessel segmentation both in a variational framework  \citep{freiman2009variational} and in a graph-based framework \citep{esneault2010liver}.
 %Note that the shape information discussed in this section is completely different from the shape prior discussed in Section \ref{sec:shape} as the use of shape information here is to  estimate $p(g_i |x\in i)$ via a filter response, while the shape model in Section \ref{sec:shape} is used to fit a top-down model to objects of interest. 
Statistical overlap prior is another strong appearance prior that has been proposed by \cite{ayed2009statistical}. Their method embeds  statistical information (e.g. histogram of intensities) about the overlap between the distributions within the object and the background in a variational image segmentation framework. They used the Bhattacharyya coefficient  measuring the amount of overlap between two distributions, i.e. $d_B(g_i(z), \hat{g}_i)=\sum_{z}\sqrt{g_i(z)\hat{g}_i(z)}\;\;\;\forall z\in\mathbb{Z}$ if $I:\Omega\rightarrow\mathbb{Z}$.
 \cite{ben2009embedding} used this strong prior to segment left ventricle in MR images.

  Other features such as frequency, bag of visual words, gradient location and orientation histogram (GLOH) \citep{mikolajczyk2005performance}, DAISY \citep{tola2008fast}, GIST (spatial envelop)  \citep{oliva2001modeling}, local binary pttern (LBP) \citep{heikkila2009description}, SURF \citep{bay2006surf}, histogram of oriented gradient (HOG) \citep{dalal2005histograms}, and scale invariant feature transform (SIFT) \citep{lowe2004distinctive}  are sometimes helpful as appearance features \citep{bosch2007image}.

Sometimes the appearance of structures is too complicated that regular features cannot describe them  accurately. To extract the appearance characteristics of such structures different machine learning techniques have been proposed. These machine learning techniques learn the appearance either by combining several features like texture, color, intensity, HOG, etc., and feed the combined feature vectors to a classifier like random decision forest (RF) or support vector machine (SVM) \citep{tu2006learning,nosrati2014efficient}, or by learning  a dictionary which describes the object of interest \citep{mairal2008discriminative,nieuwenhuis2014co,nayak2013classification}.

In general, appearance features can be extracted in the following domains based on the type of the medical data:
\begin{itemize}
\item \textbf{spatial domain:} several methods have been developed to segment 2D or 3D static images \citep{chan2000active,cootes2001active,vese2002multiphase,feddern2003level,wang2004affine,huang2005deformable,malcolm2007graph2,santner2009interactive};

\item \textbf{time domain:} in dynamic medical images, it is beneficial to consider the temporal dimension along with the spatial dimensions. For example, extracting appearance features in temporal direction would be very informative in dynamic positron emission tomography (dPET) images, where each pixel in the image represents a time activity curve (TAC) that describes the metabolic activity of a tissue as a result of tracer uptake \citep{saad2008kinetic}. Other examples include \citep{mirzaei2013decision} where spatio-temporal features are used to distinguish tumour regions in  4D lung CT (3D+time) and \citep{amir2014auto} where the likelihood of vessel regions are calculated based on temporal and frequency analysis.

\item \textbf{scale domain:} for some objects with more complex texture, it is useful to estimate the appearance model in different scales for more accurate results and ensure that the model is scale invariant \citep{han2009image,mirzaalian2010vessel}.
\end{itemize}

Regardless of where the appearance information comes from, it is encoded through a data energy term ($\mathcal{D}$)  that assigns each pixel a probability of belonging to each class of objects \eqref{eq3}.

\subsection{Regularization}
\label{sec:regularization}
The regularization term corresponds to priors on the space of feasible solutions. %As an example, the regularization term in \eqref{eq1} and \eqref{discrete} ensures that the region boundaries are smooth. 
 Several regularization terms have been proposed in the literature. The most famous one is the Mumford-Shah model \citep{mumford1989optimal} that penalizes the boundary length of different regions in a spatially continuous domain\note{, i.e. $\sum_{i=1}^n|\partial S_i|$.} %Let $\Omega$ be the image domain and each region/segment denotes by $\Omega_i, i=1,\cdots,n$, the regularization term used in \citep{mumford1989optimal} is defined as $\sum_{i=1}^n|\partial\Omega_i|$. 
The corresponding regularization model in the discrete domain is Pott's model that  penalizes any appearance discontinuity  between neighbouring pixels and is defined as $\sum_{p,q\in\mathcal{N}}w_{pq}\delta(I_p\neq I_q)$. 

The regularization term formulated in the discrete setting is biased by the discrete grid and favours curves to orient along with the grid, e.g. in horizontal and vertical or diagonal directions in a 4-connected lattice of pixels. As mentioned in Section \ref{compare}, this produces grid artifacts, also known as metrication error (Figure \ref{fig:metrication}). On the other hand, the regularization term in the continuous settings allows one to  accurately represent geometrical entities such as curve length (or surface area) without any grid bias. %On the other hand, discrete optimization problems are easier to analyse and tend to have higher optimizability.

Some other regularization terms in continuous domain, written in the level set notation ($\phi$), are listed as follows:
\begin{itemize}
\item Length regularization: $\int_\Omega|\nabla H(\phi(\bs{x}))|d\bs{x}$, where $H(.)$ is the Heaviside step function.
\item Total variation (TV): $\int_\Omega|\nabla\phi(\bs{x})|d\bs{x}$,  which only smooths the tangent direction of the level set curve. This term is used especially when a single function $\phi$ is used to segment multiple regions, i.e. $\phi$ is not necessarily a signed distance function.
It is worth mentioning that there are two variants of the total variation term: the isotropic variant using $\ell_2$ norm,
\begin{align}
\int_\Omega|\nabla\phi(\bs{x})|_2 d\bs{x}=\int_\Omega\sqrt{|\phi_{x_1}|^2+\cdots+|\phi_{x_N}|^2} \;d\bs{x}
\;,
\end{align}
and the anisotropic variant using $\ell_1$ norm,
\begin{align}
\int_\Omega|\nabla\phi(\bs{x})|_1d\bs{x}=\int_\Omega|\phi_{x_1}|+\cdots+|\phi_{x_N}|d\bs{x}\;.
\end{align}
The anisotropic version is not rotationally invariant and therefore favours results that are aligned along the grid system. The isotropic version is typically preferred but cannot be properly handled by discrete optimization algorithms \note{as the derivatives are not available in all directions in the discrete settings}.
\item $\mathcal{H}^1$ norm: $\int_\Omega|\nabla\phi(\bs{x})|^2d\bs{x}$, which applies  a purely isotropic smoothing at every pixel $\bs{x}$.
\end{itemize}
A comparison of the above mentioned regularization terms can be found in \citep{chung2009image}.

Higher order regularization terms were also proposed to encode more constraints on the optimization problem. For example, \cite{duchenne2011tensor} introduced the ternary term (along with unary and pairwise terms in the standard MRF)  for graph matching (and not image segmentation) application and \cite{delong2012minimizing} proposed an efficient optimization framework to optimize sparse higher order energies in the discrete domain.

Curvature regularization is another useful type of regularization that has been shown to be capable of capturing thin and elongated structures \citep{schoenemann2009curvature,el2010fast}. In addition, there is evidence that cells in the visual cortex are responsible for detecting curvature \citep{dobbins1987endstopped}.

\begin{figure}
\begin{center}
\subfloat[]{\includegraphics[trim=0mm 0mm 0mm 0mm,width=.6\linewidth,clip=true]{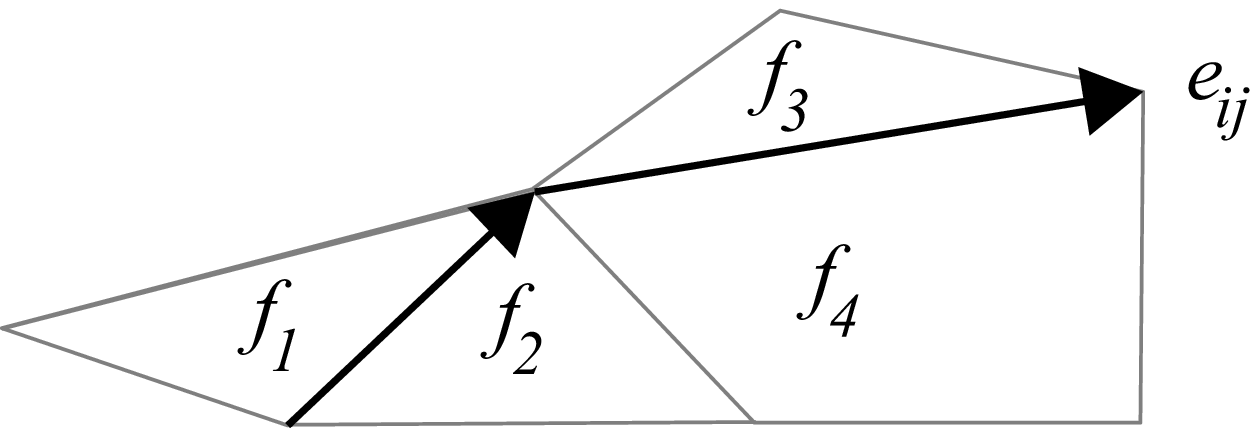}}~~
\subfloat[]{\includegraphics[trim=0mm 0mm 0mm 0mm,width=.3\linewidth,clip=true]{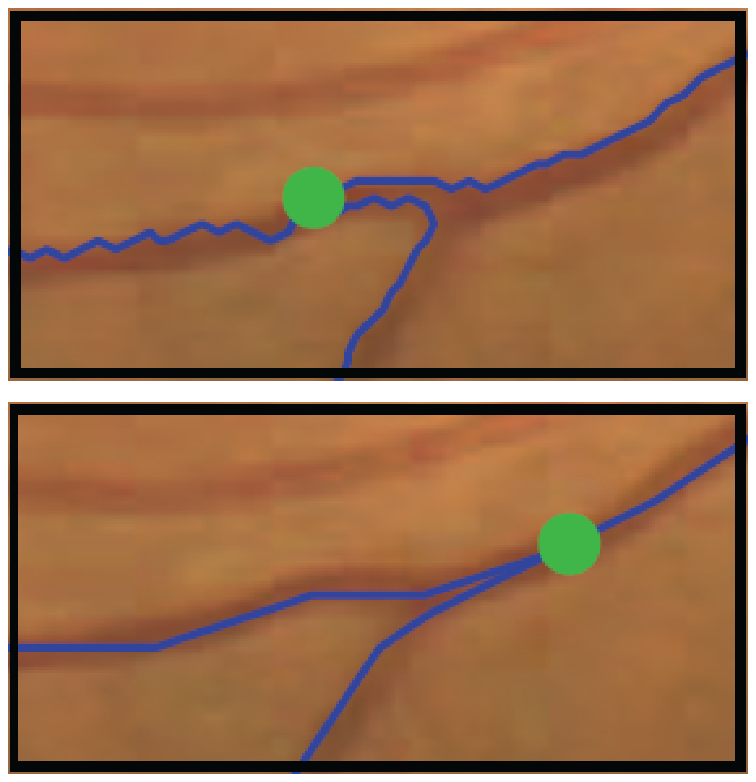}}
\end{center}
\caption{Curvature regularization term. (a) Approximating the curvature term by tesselating the image domain into a cell complex. $e_{ij}$ is the boundary variable and $f_1$, $f_2$, $f_3$ and $f_4$ are its corresponding  region variables. (b) Vessel segmentation \textbf{top:} without curvature and \textbf{bottom:} with curvature regularization term. \small(Image (b) adopted from \citep{strandmark2013shortest})}
\label{fig:curvature}
\end{figure}

While curvature regularization term can be easily formulated in the local optimization frameworks, e.g. in a level set formulation \citep{leventon2000level} and Snakes' model \citep{kass1988snakes}, it is much more difficult to incorporate such prior in a global optimization framework. \cite{strandmark2011curvature} proposed several improvements to approximate curvature regularization within  a global optimization framework. They defined the curvature term as: $\int_{\partial R} k(\bs{x})^2 d\bs{x}$, where $\partial R$ is the boundary of the foreground region and $k$ is the curvature function. They  approximated the above mentioned curvature term with  discrete computation techniques by tessellating the image domain into a cell complex, e.g. hexagonal mesh, which is a collection of non-overlapping basic regions whose union gives the whole domain. They recast the problem as an integer linear program (along with a data term and length/area regularization terms) and optimized the total energy  via linear programming (LP) relaxation. Figure \ref{fig:curvature}(a) shows how \cite{strandmark2011curvature} discretized the image domain by cells. If  $f_i$, $i=1,\cdots,m$ denotes binary variables associated to each cell region and $e_{i}$ denotes the boundary variable, then the curvature regularization term is written as a linear function: $\sum_{i,j}b_{ij}e_{ij}$, where $e_{ij}$ denotes the boundary pairs and
\begin{align}
b_{ij}=\min\{l_i,l_j\}\left(\frac{\alpha}{\min\{l_i,l_j\}}\right)^2
\;,
\end{align}
where $l_i$ is the length of edge $i$ and $\alpha$ is the angle difference between two lines. %To impose the consistency between edges and their corresponding regions, the authors introduced some constraints. Assume $y_{ij}$ corresponds to its four incident regions $x_1$, $x_2$, $x_3$ and $x_4$ (Figure \ref{fig:curvature}(a)). If $x_1=x_2=1$ or $x_1=x_2=0$, $y_{ij}$ should not be activated (similarly for $x_3$ and $x_4$). These constraints can be linearly encoded as:
%\begin{align}
%x_1+x_2+y_{ij}+y_{ji}&\leq 2\\
%-x_1-x_2+y_{ij}+y_{ji}&\leq 0
%\;.
%\end{align}
Later, \cite{strandmark2013shortest} extended their previous work \citep{strandmark2011curvature} and proposed a globally optimal shortest path method that minimizes general functionals of higher-order curve properties, e.g. curvature and torsion. Figure \ref{fig:curvature}(b) illustrates the usefulness of curvature prior on vessel segmentation.

\subsection{Boundary information}
\label{sec:boundary}
Boundary and edge information is a powerful feature for delineating the objects of interest in an image. To incorporate such information, it is often assumed that the object boundaries are more likely to pass between pixels with large intensity/color contrast or, more generally, regions with different appearance (as captured by any of the measures in Section \ref{sec:appearance}). As object boundaries are locations where we expect discontinuities in the labels, this information is usually linked to the regularization term in \eqref{eq0} such that the regularization penalty is decreased in high contrast regions (most likely objects' boundaries) to allow for discontinuity in labels. The functions $w_{ij}=exp(-\beta\|I_i-I_j\|^2_2)$ and $w'_{ij}=1/1+\beta\|I_i-I_j\|^2_2$ are two examples of a boundary weighting function where $I_i$ and $I_j$ represent the intensity/color value associated with pixels $i$ and $j$ in image $I$, respectively \citep{grady2012targeted}. These boundary weights are used as  multiplication factors along with the regularization terms mentioned in Section \ref{sec:regularization}. Geodesic active contour \citep{caselles1997geodesic}, normalized-cut  \citep{shi2000normalized}, and random walker \citep{grady2006random} are three examples that employed such boundary weighting technique.

Boundary and edge information can also be linked to the data term in \eqref{eq0} via the use of edge detectors, which typically involve first and second order spatial differential operators. Several methods have been proposed to calculate first and second order differences in scalar images \citep{canny1986computational,frangi1998multiscale} and color images \citep{shi2008quaternion,tsai2002adaptive}. However, some medical images are manifold-valued (e.g. DT MRI). To address this, \cite{nand2011detecting} extended the first order differential as $g(\boldsymbol{x})=\sqrt{\lambda}\hat{e}$ where  $\lambda$ and $\hat{e}$ are respectively the largest eigenvalue and eigenvector of $\boldsymbol{S}(\boldsymbol{x})=\boldsymbol{J}(\boldsymbol{x})^T\boldsymbol{J}(\boldsymbol{x})$ and $\boldsymbol{J}(\boldsymbol{x})$ is the Jacobian matrix generalizing the gradient of a scalar field to the derivatives of the 3D DT image. Similarly, the authors extended the second order differential as $G'(\boldsymbol{x})=\frac{G(\boldsymbol{x})+G(\boldsymbol{x})^T}{2}$ where $G(\boldsymbol{x})$ is the Jacobian matrix of $g(\boldsymbol{x})$, i.e. $G_{ij}=\frac{\partial g_{i}}{\partial x_j}$.  Similar approach has been proposed for boundary detection in color images, e.g. in color snakes  \citep{sapiro1997color} and in detecting boundaries of oral lesions in color images  \citep{chodorowski2005color}.

\begin{figure}[!t]
\begin{center}
\subfloat[]{\includegraphics[trim=0mm 0mm 0mm 0mm,width=.3\linewidth,clip=true]{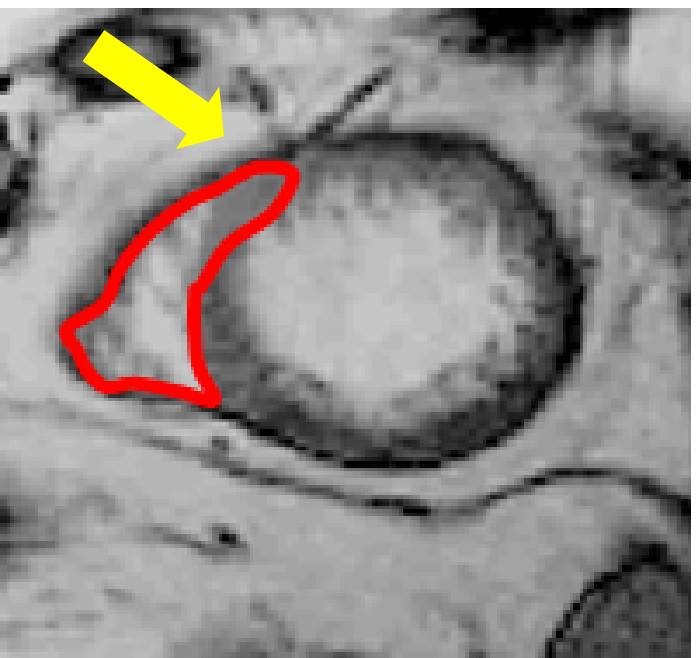}}~~~~~~~
\subfloat[]{\includegraphics[trim=0mm 0mm 0mm 0mm,width=.3\linewidth,clip=true]{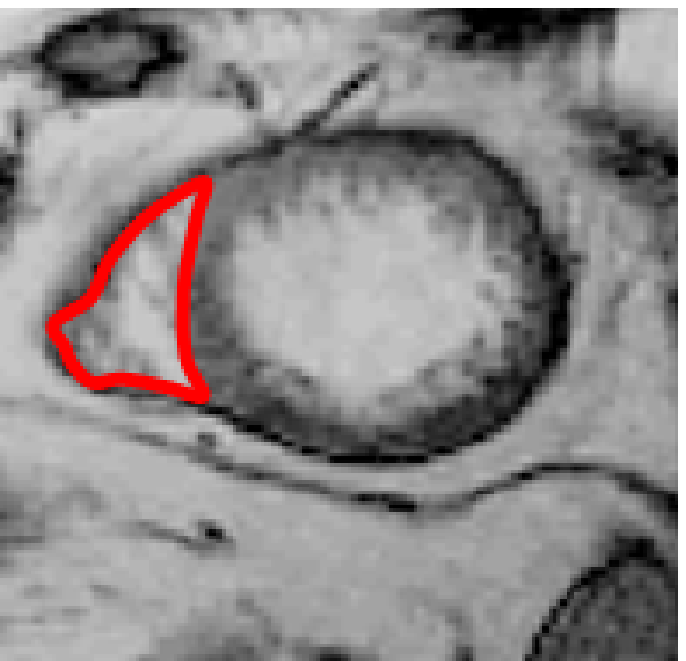}}
\end{center}
\caption{\note{Cardiac right ventricle segmentation (a) without encoding edge polarity and (b) with encoding edge polarity by specifying the bright to dark edges as the desired ones.} Note how the incorrect boundary transition (the yellow arrow)  in (a) has been corrected in (b) by specifying boundary polarity.}
\label{fig:boundarypolarity}
\end{figure}

\textbf{Boundary polarity:} A problem with the aforementioned boundary models is that they describe a boundary point that passes between two pixels with high image contrast without accounting for the direction of the transition \citep{boykov2006graph,grady2012targeted}.   \cite{singaraju2008interactive} considered the transition direction in boundary detection. For example, it is possible to distinguish between boundaries from bright to dark and from dark to bright (boundary polarity). This boundary polarity is incorporated into a graph-based framework by replacing each undirected edge, $e_{ij}$, by two directed edges, $e_{ij}$ and $e_{ji}$, with edge weight calculated as:
\begin{align}
&w_{ij}=
\begin{cases}
\exp(-\beta_1\| I_i-I_j\|^2_2) & \text{if} ~I_i>I_j\\
\exp(-\beta_2\| I_i-I_j\|^2_2) & \text{otherwise}
\end{cases}
\label{boundarypolarity}
\;,
\end{align}
where $\beta_1>\beta_2$. In \eqref{boundarypolarity}, boundary transition from bright to dark is less costly than boundary transition from dark to bright. \note{One example of encoding boundary polarity is shown in Figure \ref{fig:boundarypolarity}, where the boundary ambiguity is resolved by specifying the boundary polarity, i.e. in this example,  bright to dark boundary.}

The assumption of high contrast in objects' boundaries might not be always  valid in many medical images, e.g. soft tissue boundaries in CT images. In addition, the two proposed  contrast models, $w_{ij}$ and $w'_{ij}$, are suitable for objects with smooth appearance and not for textured objects. One possible way to address these aforementioned issues (low contrast image and textured objects) is to utilize the piecewise constant case of Mumford-Shah model \citep{mumford1989optimal} and replace $I_i$ with $\tau(I_i)$, where $\tau$ is a function that maps the pixel content to a transformed space where the object appearance is relatively constant \citep{grady2012targeted}. The Mumford-Shah model segments the image into  a set of pairwise disjoint regions with minimal appearance variance and minimal boundary length. Among the most popular methods that adopted the Mumford-Shah model is the active contours without edges (ACWOE) method proposed by  \cite{chan2001active}. As an example \citep{sandberg2002level} proposed a level set-based active contour algorithm to segment textured objects. Another example is the work proposed by  \cite{paragios2002geodesic} where  boundary and region-based segmentation modules were exploited and unified into a geodesic active contour model to segment textured objects.

\subsection{Extending binary to multi-label segmentation}
\label{sec:multilabel}
In many medical image analysis problems, we are often  interested in segmenting multiple objects (e.g. segmenting retinal layers from optical coherence tomography \citep{yazdanpanah2011segmentation}). Unlike a large class of binary labeling problems that can be solved globally, multi-label problems, on the other hand, cannot be globally minimized in general. In 2001,  \cite{boykov2001fast} proposed two algorithms ($\alpha$-expansion and $\alpha$-$\beta$ swap) based on graph cuts that efficiently find a local minimum of a multi-label problem. They consider the following energy functional
\begin{align}
E(\bs{f})=\sum_{p\in\mathcal{P}} D_p(f_p)+\sum_{\{p,q\}\in\mathcal{N}}V_{pq}(f_p,f_q)
\;,
\label{ishikawa}
\end{align}
where $\mathcal{P}$ is the set of all pixels, $\bs{f}=\{f_p|p\in\mathcal{P}\}$ is a labeling of the image,  $D_p(f_p)$ measures how well label $f_p$ fits pixel $p$ and $V_{pq}$ is a penalty term for every pair of neighbouring  pixels $p$ and $q$ and encourages neighbouring pixels to have the same label. The second term ensures that the segmentation boundary is smooth. The methods proposed in  \citep{boykov2001fast} require $V_{pq}$ to be either a \emph{metric} or \textit{semimetric}. $V$ is a metric on the space of labels $\mathcal{L}$ if it satisfies the following three conditions:
\begin{align}
V(\alpha,\beta)=0 \Leftrightarrow \alpha=\beta \label{metric1}\\
V(\alpha,\beta)=V(\beta,\alpha)>0\label{metric2}\\
V(\alpha,\beta)\leq V(\alpha,\gamma)+V(\gamma,\beta)
\;,
\end{align}
for any labels $\alpha$, $\beta$,$\gamma\in\mathcal{L}$. If $V$ only satisfies 	\eqref{metric1} and  \eqref{metric2} then $V$ is a semimetric. \cite{boykov2001fast} find the local minima by swapping a pair of labels ($\alpha$-$\beta$-swap) or expanding a label ($\alpha$-expansion) and evaluate the energy using graph cuts iteratively. Later in 2003, Ishikawa \citep{ishikawa2003exact} 
showed that, if $V_{pq}(f_p,f_q)$ is convex and symmetric in $f_p-f_q$, one can compute the exact solution of the multi-label problem. 
 Ishikawa used the following formulation:
\begin{align}
E(f)=\sum_{p\in\mathcal{P}}D(f_p)+\sum_{(p,q)\in\mathcal{N}}g(\ell(f_p)-\ell(f_q))\;,
\label{ishikawa2}
\end{align}
where $D(.)$ in the first term (data term) is any bounded function that can be non-convex, $g(.)$ is a convex function, and $\ell$ is a function that gives the index of a label, i.e. $\ell(label~i)=i$.  The term $g(\ell(f_p)-\ell(f_q))$ expresses that there is a linear order among the labels and the regularization depends only on the difference of their ordinal number. Ishikawa  showed that if $g(.)$  is convex in terms of a linearly ordered label set, the problem of \eqref{ishikawa2} can be exactly optimized by finding the min-cut over a specially constructed multi-layered graph in which each layer corresponds to one label.
%where $U=\{u:\mathcal{P}\rightarrow\mathcal{L}\}$ is the set of all feasible  labelling functions, $\rho$ in the first term (data term) is any bounded function that can be non-convex and $g(x)$ is a convex function. The term $g(u_v-u_w)$ expresses that there is a linear order among the labels and the regularization depends only on the difference of their ordinal number. Ishikawa  showed that if $g(x)$  is convex in terms of a linearly ordered label set, the problem of \eqref{ishikawa} can be exactly optimized by finding the min-cut over a specially constructed multi-layered graph in which each layer corresponds to one label.}

\begin{figure}
\begin{center}
\subfloat[]{\includegraphics[trim=0mm 0mm 0mm 0mm,width=.25\linewidth,clip=true]{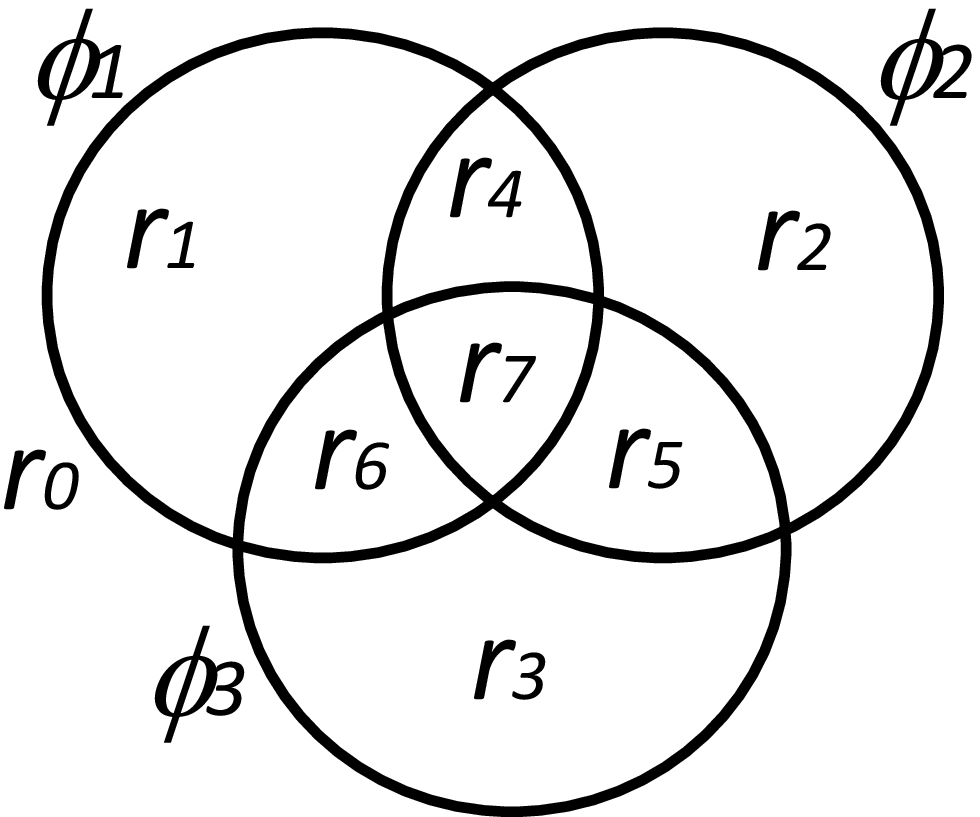}}~
\subfloat[]{\includegraphics[trim=0mm 0mm 0mm 0mm,width=.25\linewidth,clip=true]{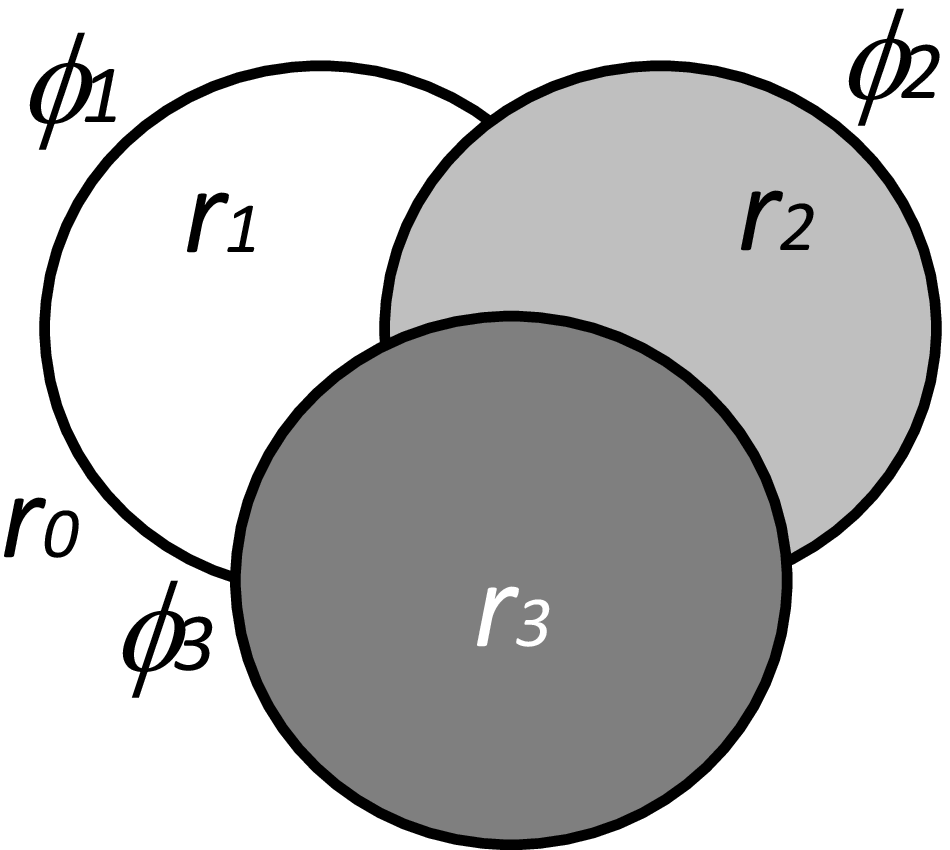}}~
\subfloat[]{\includegraphics[trim=0mm 0mm 0mm 0mm,width=.3\linewidth,clip=true]{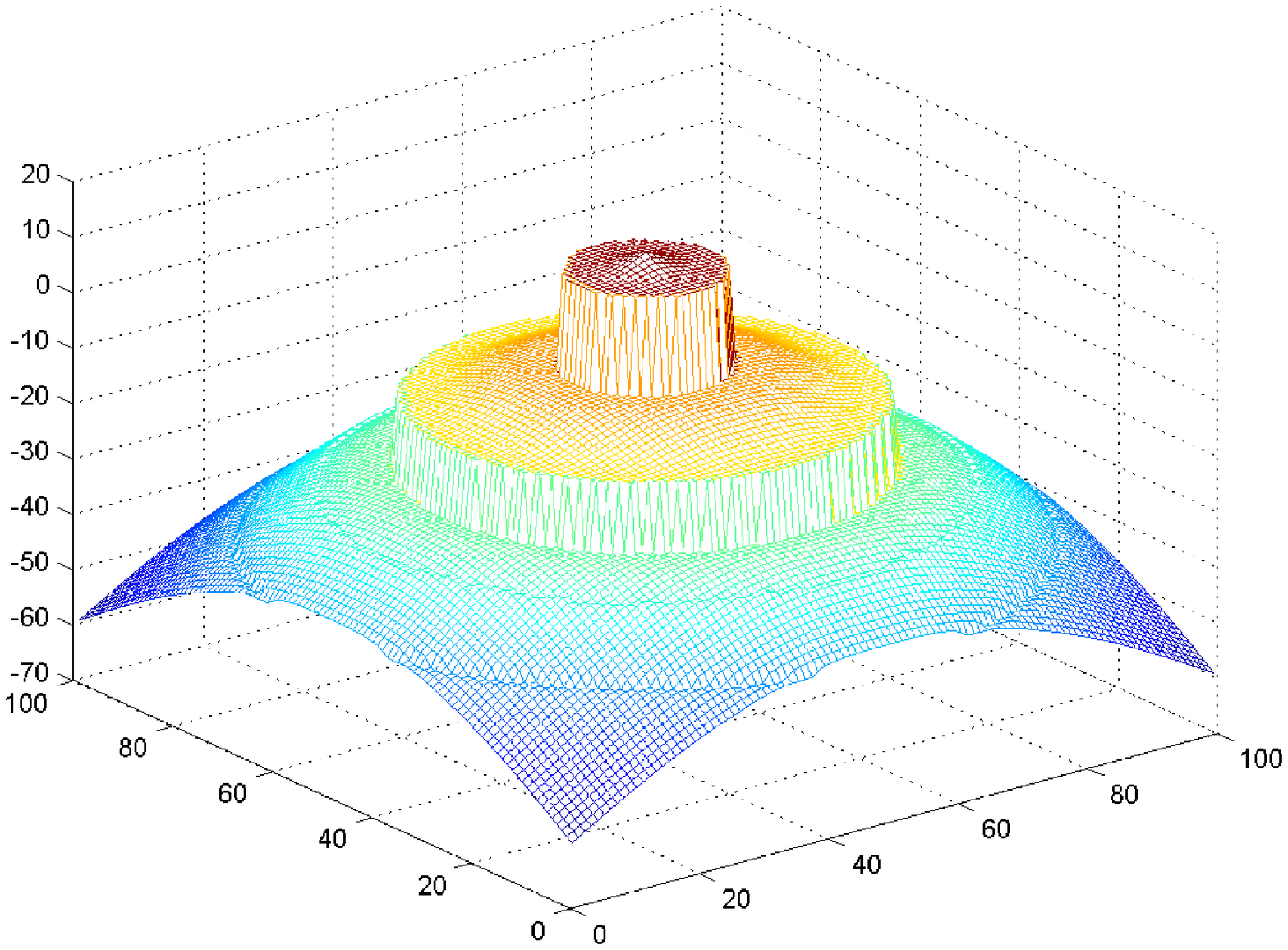}}
\end{center}
\caption{Multi region level set methods proposed by (a) \cite{vese2002multiphase}, (b) \cite{mansouri2006multiregion}, and (c) \cite{chung2009image}.}
\label{fig:levelset}
\end{figure}

In the continuous domain, \cite{vese2002multiphase} extended their level set-based method to multiphase level sets. To segment $N$ objects, their method 
needs $\lceil \log_2 N\rceil$ level set functions. The number of regions is upper-bounded by a power of two (Figure \ref{fig:levelset}(a)). \note{Therefore, the actual number of regions the method yields is sometimes not clear as it depends on the image and the regularization weights. This issue happens specifically when the number of regions of interest is less than $2^{\lceil \log_2 N\rceil}$.}  \cite{mansouri2006multiregion} proposed to assign an individual level set function to each object of interest (excluding the background), i.e. their method needs $N-1$ non-overlapping level set functions to segment $N$ objects (Figure \ref{fig:levelset}(b)).
\cite{chung2009image} proposed another method that uses a single level set function for multi-object segmentation. They proposed to use different layers (or levels) of a level set function to represent different regions as opposed to just using the zero level set (Figure \ref{fig:levelset}(c)). None of the aforementioned continuous methods guarantee a globally optimal solution for multi-label problems. 
 \cite{pock2008convex} proposed a spatially continuous formulation of Ishikawa's multi-label problem. In their method, the non-convex variational problem is reformulated  as a convex variational problem via a technique they called \emph{functional lifting}. They used the following energy functional
\begin{align}
E(u)=\int_\Omega\rho(u(\bs{x}),\bs{x})d\bs{x}+\int_\Omega|\nabla u(\bs{x})|d\bs{x}
\;,
\label{pock}
\end{align}
which can be seen as the continuous version of Ishikawa's formulation \eqref{ishikawa}. $u:\Omega\rightarrow\Gamma$ in \eqref{pock} is the unknown labeling function and $\Gamma=[\gamma_{min},\gamma_{max}]$ is the range of $u$. The first term in \eqref{pock} is the data term, which can be a non-convex function, and the second term is the total variation regularization term which is a convex term. In the functional lifting technique, the idea is to transfer the original problem formulation to a higher dimensional space by representing $u$ in terms of its super level sets $\phi$ defined as:
\begin{align}
\phi(\bs{x},\gamma)=
\begin{cases}
1 & \text{if }u(\bs{x})>\gamma\\
0& \text{otherwise}\\
\end{cases}
\; .
\label{superlevelset}
\end{align}
Now, \eqref{pock} can be re-written in terms of the super level set function as
\begin{align}
E(\phi)=\int_\Sigma\rho(\bs{x},\gamma)|\partial_\gamma\phi(\bs{x},\gamma)|d\Sigma+\int_\Sigma|\nabla\phi(\bs{x},\gamma)|d\Sigma
\; ,
\end{align}
which is convex in $\phi$ and $\Sigma=[\Omega\times\Gamma]$. The minimization of $E(\phi)$ is not a convex optimization problem since $\phi:\Sigma\rightarrow\{0,1\}$. Hence, $\phi$ is relaxed to vary in $[0,1]$. We emphasize that the method of Pock et al. cannot always guarantee the globally optimal solution of the original problem (before $\phi$ is relaxed and when $\phi$ is binary).  \cite{brown2009convex} utilized functional lifting technique proposed by \citep{pock2008convex} and proposed a dual formulation for the multi-label problem. Their method guarantees a globally optimal solution. 
Recently, inspired by Ishikawa,  \cite{bae2011fast} proposed a continuous max-flow model for multi-labeling problem via convex relaxed formulations. Not only can their continuous max-flow formulations obtain exact and global optimizers to the original problem, but they also showed that their method is significantly faster than the primal-dual algorithm of \cite{pock2008convex}.

\subsection{Shape prior}
\label{sec:shape}
Shape information is a powerful semantic descriptor for specifying targeted objects in an image. In our categorization, shape prior can be modelled in three ways: geometrical (template-based), statistical, and physical. %The general idea is to find the shape probability and encode it into the segmentation energy functional, e.g. using Mahalanobis distance measure as is done in \citep{andrews2011convex}, to constrain the segmentation to follow global shape consistency while capturing local deformations.

\subsubsection{Geometrical model (template)}
Sometimes the shape of the targeted object is known a priori  (e.g. ellipse or cup-like shape). In this case, the shape can be modelled either by parametrization (e.g. an ellipse can be parametrized by its center coordinate, major and minor radius and  orientation) or by a non-parametric way (e.g. by its level set representation) and incorporated into a segmentation framework.

One way to incorporate a geometrical shape model into a segmentation framework is to penalize any deviation from the model. In the continuous domain, given two shapes represented by their signed distance functions $\phi_1$ and $\phi_2$, a simple way to calculate the dissimilarity between them is given by $\int_\Omega (\phi_1-\phi_2)^2d\bs{x}$. The problem with this measure is that it depends on $\Omega$, i.e. as the size of $\Omega$ is increased, the difference becomes larger. An alternative is to constrain the integral to the domain of $\phi_1$, i.e. $\int_\Omega(\phi_1-\phi_2)^2H(\phi_1)d\bs{x}$, as proposed in  \citep{rousson2002shape}.
The aforementioned formulas are usable if the pose of the object  of interest (location, rotation and scale) is known. If the pose of an object is unknown, one can include the pose parameters into the shape energy term and optimize the energy functional with respect to both pose parameters and the level set.  For example, the authors in \citep{chen2002using} imposed the shape prior on the extracted contour after each iteration of their level set-based algorithm. \cite{pluempitiwiriyawej2005stacs} also described the shape of an ellipse with five parameters that include its pose parameters and optimized their energy functional by iterating between optimizing the shape energy term and the regional term.

\begin{figure}
\begin{center}
\subfloat[]{\includegraphics[trim=0mm 0mm 0mm 0mm,width=.5\linewidth,clip=true]{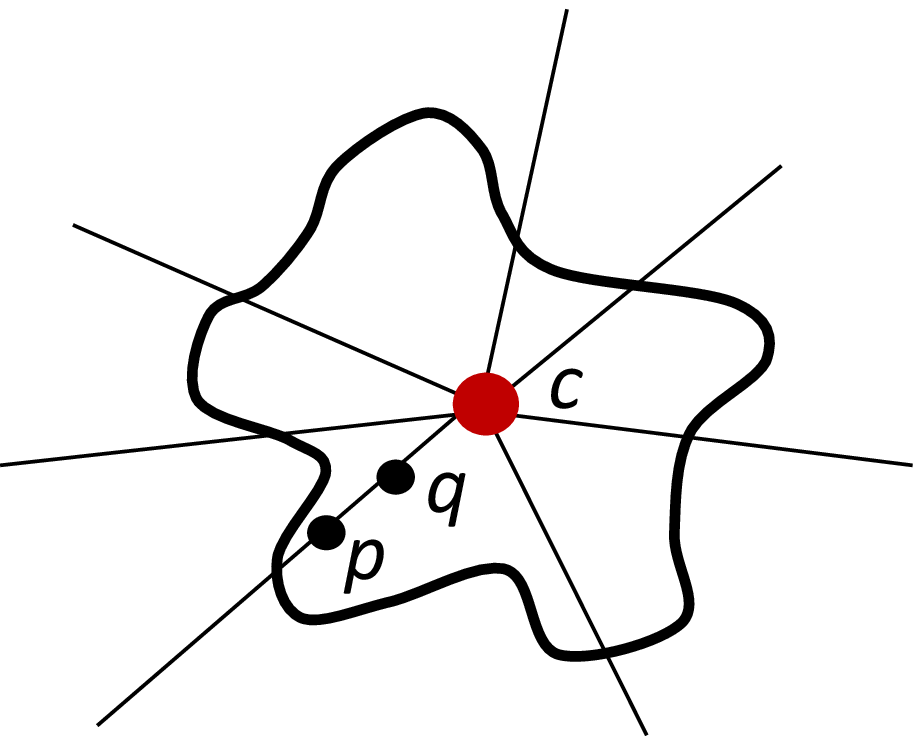}}
\end{center}
\caption{A star shape with a  given center $c$. $p$ an $q$ are two points on the line passing through $c$. If $p$ is labeled as object, then $q$ must also be labeled as the object.}
\label{fig:starshape}
\end{figure}

In the discrete domain, the method of \cite{slabaugh2005graph} is one of the primary works to incorporate an explicit shape model into a graph-based segmentation framework. They proposed the following extra term (in addition to data and regularization terms) that constrained the segmentation to return an elliptical object:
\begin{align}
E_{ellipse}(f,\theta)=\sum_{i\in\mathcal{P}}|M^\theta_i-f_i| \;,
\label{GCellipse}
\end{align}
where $M^\theta$ is the mask of an ellipse parametrized by $\theta$. As minimizing  such a term is not straightforward, the authors optimize the energy functional iteratively, i.e. by finding the best $f$ for a fixed $\theta$ and then optimizing $\theta$ for a fixed $f$.
For complex shapes that are hard to parametrize, an alternative approach is to fit a shape template to the current segmentation as proposed in  \citep{freedman2005interactive}. 
\cite{veksler2008star} proposed to incorporate a more general class of shapes, known as \emph{star shapes}, into graph-based segmentation. In Veksler's work, it is assumed that the center point ($c$) of the object is given. According to their definition, ``an object has a star shape if for any point $p$ inside the object, all points on the straight line between the center $c$ and $p$ also lie inside the object'' (Figure \ref{fig:starshape}). The following pairwise term was introduced to impose the star shape prior:
\begin{align}
E^{Star}_{pq}(f_p,f_q)=
\begin{cases}
0 & \text{if } f_p=f_q\\
\infty & \text{if } f_p=1~~ \text{and}~~ f_q=0\\
\beta & \text{if } f_p=0 ~~\text{and}~~ f_q=1
\end{cases}
\;.
\label{starshape}
\end{align}
This prior is	particularly useful for segmentation of convex objects, e.g. optic cum and disc segmentation \citep{bai2014graph}.
\subsubsection{Statistical model}
In many practical applications, objects of the same class are not identical or rigid. For example, in medical images,  the shape of organs vary from one subject to another or even over time and so, assuming a fixed shape template may be inappropriate.  A typical way to capture the intra-class variation of shapes  is to build   a shape probability model, i.e. $P(shape)$. Now, two questions have to be investigated: 
1) how to represent a shape; explicitly (e.g. point cloud), implicitly (e.g. level set), boundary-based (e.g. surface mesh) or medial-based (e.g. m-reps \citep{pizer2003deformable}), and  2)  what probability distribution model to adopt, e.g. Gaussian distribution,  Gaussian mixture model, or kernel density estimation (KDE).
%, and wavelet coefficients are some shape representation examples that have been used in the literature. Gaussian distribution,  Gaussian mixture model (GMM), and kernel density estimation (KDE) are some examples of probability models that can be used in building a shape probability.} To capture the intra-class variation of shapes and generate a compact shape representation, 

\cite{cootes1995active} generated a compact shape representation and performed PCA (assuming Gaussian distribution) on a set of training shapes to obtain the main modes of variation. The idea is to model the plausible deformations of object's shape ($S$) by its principal modes of variation:
\begin{align}
S=\overline{S}+\sum_{i=1}^k w_iP_i
\;,
\label{stat}
\end{align}
where $\overline{S}$ is the average shape, $P_i$ is the $i^{th}$ principal component and $w_i$ is  its corresponding weight (or shape parameter). \cite{cootes1995active} used object's coordinates to represent $S$. %The shape $S$ can be represented by, for example, object's coordinates or signed distance map of an object's surface.  
 Given an initial estimation of the position of an object, the segmentation is performed by directly optimizing an energy functional over the weights $w_i$. This model is later improved by  \cite{tsai2001model,tsai2003shape, leventon2000statistical} and  \cite{van2002active}. For example, \cite{leventon2000statistical} represent $S$ by its level sets to automatically handle topological changes during the contour evolution. \cite{tsai2003shape} used the same level set-based shape representation as  \cite{leventon2000statistical} and incorporated the shape prior in a region-based energy functional as opposed to an edge-based energy proposed in \citep{cootes1995active}. \cite{van2002active} proposed to use a general set of local image structure descriptors including the moments of local histograms instead  of the  normalized first order derivative profiles used in \cite{collins1995automatic}.
 %The shape $S$ can be represented by, for example, object's coordinates or signed distance map of an object's surface.  Given an initial estimation of the position of an object, the segmentation is performed by directly optimizing an energy functional over the weights $w_i$. This model is later improved by  \cite{tsai2001model,tsai2003shape, leventon2000statistical} and  \cite{van2002active}.

Similar to \cite{tsai2003shape} in the continuous domain, \cite{zhu2007graph} employed an iterative approach that accounts for shape variability in a graph-based setting. At each iteration, they optimize  the weights of principal modes of variations and the set of rigid transformation parameters given a tentative segmentation. Then, the segmentation is updated given the fitted shape template by minimizing an energy functional consisting of a regional term. The procedure is repeated until convergence. Recently, \cite{andrews2014isometric} proposed a probabilistic framework and incorporated shape prior to segment multiple anatomical structures. They utilized PCA in the isometric log-ratio space as PCA assumes that the probabilistic data lie in the unconstrained real Euclidean space. This is not a valid assumption as the sample space for a probabilistic data is the unit simplex and PCA may generate invalid probabilities, and hence, invalid shapes.

In the above mentioned methods based on PCA, aligning the shapes before computing the principal modes of variation is necessary and to perform this alignment, it is often needed to provide point-to-point correspondences between landmarks of different subjects. This might be a tedious task. Hence, some methods proposed to capture shape variations in the frequency domain by representing shapes with the coefficients of its discrete cosine transform (DCT) \citep{hamarneh2000statistically}, Fourier transform  \citep{staib1992deformable} or spherical wavelet transform \citep{nain2006shape}.

While PCA is a popular linear dimensionality reduction technique, it has the restrictive assumption that the input data is drawn from a Gaussian distribution. \note{If the shape variation does not follow a Gaussian distribution, we might end up with invalid shapes or unable to represent valid shapes. In this case, a more accurate estimation of shape parameters might be obtained by Gaussian mixture models  as proposed in  \citep{cootes1999mixture}}. In addition, PCA is only capable of describing global shape variations, i.e. changing a parameter corresponding to one eigenvector deforms the entire shape, which makes it difficult to obtain a proper local segmentation. 
To control the statistical shape parameters locally,  \cite{davatzikos2003hierarchical}  presented a hierarchical formulation of active shape models using the wavelet transform. Their wavelet-based encoding of deformable contours is followed by PCA analysis. The statistical properties extracted by PCA are then used as priors on the contour's deformation, some of which  capture  global shape characteristics of the object boundaries while others capture local and high-frequency shape characteristics.  \cite{hamarneh2004medial} also proposed a method to locally control the statistical shape parameters. They used the medial-based profile for shape representation and developed spatially-localized feasible deformations using hierarchical (multi-scale) and regional (multi-location) PCA and deform the medial profile at certain locations and scales.  \cite{uzumcu2003ica} proposed to use independent component analysis (ICA) instead of PCA which does not assume a Gaussian distribution of the input data and can better  capture  localized shape variations. However, ICA representation for shape variability is not as compact as PCA. \cite{ballester2005adequacy} proposed to use principal factor analysis (PFA) as an alternative to PCA. PFA represents the observed $D$-dimensional data $\bs{O}$ as a linear function $\bs{\mathcal{F}}$ of an $L$-dimensional ($L<D$) latent variable $z$ and an independent Gaussian noise $\bs{e}$ as: $\mathcal{F}(\bs{O})=\bs{\Lambda} z+\bs{\mu} + err$, where $\bs{\Lambda}$ is the $D\times L$ \emph{factor loading matrix} defining the linear function $\mathcal{F}$, $\bs{\mu}$ is a $D$-dimensional vector representing the mean of the distribution of $\bs{O}$, and $err$ is a $D$-dimensional vector representing the noise variability associated with each of the $D$ observed variables. As PFA models covariance between variables and generates ``interpretable''modes of variation, while  PCA determines the factors that account for the total variance, \citep{ballester2005adequacy} argued that PFA is not only adequate for the study of shape variability but also gives better ``interpretability'' than PCA, and thus conclude that  PFA is better suited  for medical image analysis.%Authors in \citep{ballester2005adequacy} concluded that PFA is adequate for the study of shape variability and provides better ``interpretability'' than PCA as PFA models covariance between variables  whereas  PCA determines the factors which account for the total variance.

Nevertheless, both PCA and ICA  are linear factor analysis techniques, which make them difficult to model non-linear shape variations. Techniques such as kernel PCA \citep{scholkopf1998nonlinear} and kernel density estimation (KDE) are two alternatives to describe non-linear data.  The works proposed by \citep{cremers2006kernel,kim2007nonparametric,lu2012simultaneous}  are examples that used non-linear dimensionality reduction techniques (e.g. kernel PCA and  KDE) to incorporate shape priors into image segmentation frameworks. For more information about other linear and non-linear factor analysis techniques, we refer to \citep{fodor2002survey,bowden2000non}.

In addition to representing shapes as a set of points (as usually done in e.g. PCA cf. \eqref{stat}), shapes can be described by distance and angle information between different anatomical landmarks  \citep{wang20103d,nambakhsh2013left}. For example,  \cite{wang20103d} proposed a scale-invariant shape description by measuring the relative distances between pair of landmarks in a triplet, while \cite{nambakhsh2013left} model the left ventricle (LV) shape in the cardium by calculating the distance between each point on the surface of the LV and a reference point in the middle of the LV provided by a user. More reviews on statistical shape models for 3D medical image segmentation can be found in \citep{heimann2009statistical}.

Beside the aforementioned statistical methods,  some methods employed learning algorithms to impose a shape model into segmentation \citep{zhang2012deformable,kawahara2013augmenting}. In \citep{zhang2012deformable}, authors proposed a deformable segmentation method based on sparse shape composition and dictionary learning.  In another work, \cite{kawahara2013augmenting} augmented the auto-context method \citep{tu2010auto} and trained sequential classifiers for segmentation. Auto-context \citep{tu2010auto} is an iterative learning framework that jointly learns the appearance and regularization distributions where the predicted labels from the previous iteration are used as input to the current iteration.  \citep{kawahara2013augmenting} used auto-context to learn what shape-features (e.g. volume of a segmentation) a good segmentation should have.

\subsubsection{Physical model}
In some medical applications, the biomechanical  characteristics of tissues can be estimated so that the  physical characteristics of tissues can be modeled in a  segmentation framework as additional prior information, thereby leading to more reliable segmentations.  

The incorporation of material elasticity property into  image segmentation was first introduced in 1988 by  \cite{kass1988snakes} in which spring-like forces between snake's points is enforced. Following Kass' snakes model, several researchers also examined ways to extract  vibrational (physical) modes of shapes based on finite element method (FEM); these include methods proposed by  \cite{karaolani1989finite},  \cite{nastar1993non} and  \cite{pentland1991closed}. In these frameworks, an object is modelled based on its vibrational modes similar to \eqref{stat} where $P_i$ is a vibrational mode rather than a statistical mode.

When the physical characteristics of a tissue are known and several samples from the same tissue are available, one can take advantage of both statistical and physical models to obtain more accurate segmentation, as done by \cite{cootes1995combining} and  \cite{hamarneh2008simulation} where   statistical and vibrational modes of variation are combined into a single objective function.

\cite{schoenemann2007globally} encoded an elastic shape prior into a segmentation framework by combining the shape matching and segmentation tasks. Given a shape template, they proposed an elastic shape matching energy term that maps the points of the evolving shape to the template based on two criteria: 1) points of similar curvature should be matched, and 2) a curve piece of the evolving shape should be matched to a piece of the template of equal size. Their method achieves  globally optimal solutions.
%framework . Given a deformable shape $S_1:[0,l(S_1)]\rightarrow\mathbb{R}^2$ to be mapped to a shape  $S_2:[0,l(S_2)]\rightarrow\mathbb{R}^2$, their framework aims to find an optimal matching function $m:[0,l(S_1)]\rightarrow[0,l(S_2)]$ such that the point $S_1(s)$ and $S_2(m(s))$ correspond to each other. Their framework is based on two aspects: 1) points of similar curvature should be matched, and 2) a curve piece of $S_1$ should be matched to a piece of $S_2$ of equal size which implies an ideal derivative of $m'(s)=1$. The following energy function is proposed to penalize deviations from above aspect:
%\begin{align}
%\int_0^{l(S_1)}|k_{S_1}(s)-m'(s)k_{S_2}(s)|ds+\lambda\int_0^{l(c)}\psi(m'(s))ds
%\;,
%\end{align}
%where $l(.)$ denotes the length of the curves, and $k_{S_1}(s)$ and $k_{S_2}(s)$ denote the curvature of $S_1$ at $S_1(s)$ and $S_2$ at $S_2(s)$, respectively. $\psi(.)$ is defined as:
%\begin{align}
%&\psi(m')=
%\begin{cases}
%m'-1 & \text{if } m'\geq 1,\\
%\frac{1}{m'}-1& \text{otherwise}
%\end{cases}
%\;
%\end{align}
%and offers the advantage of symmetry, i.e. comparing $S_1$ to $S_2$ and $S_2$ to $S_1$ gives the same value.
\\

\subsection{Topological prior}
Many anatomical objects in medical images have a specific topology that has to be preserved after segmentation in order to obtain plausible results.  There are two types of topology specification in the literature: \emph{connectivity} and \emph{genus}. Connectivity specification ensures that the segmentation of a single object is connected\footnote{Formally, a segmentation $S$ is connected if $\forall x,y\in S,\;\exists Path_{xy}$, s.t. if $z\in Path_{xy}$, then $z\in S$.}. The genus information ensures that the final segmentation does not have any void region (if the object is known to be connected) or incorrectly fill void regions when the object is known to have internal holes \citep{grady2012targeted}. For example, a doughnut-shape initial segmentation should keep its shape (doughnut) during the segmentation process.

\cite{han2003topology} proposed a level set-based method for segmenting objects with topology preservation. Their method is based on the \emph{simple point} concept from digital topology \citep{bertrand1994simple}. \note{A \emph{simple point} is one that does not change the topology of the segmentation when it is added or removed from a segmentation.}  Specifically, the proposed method checks the topological number at each iteration to detect topological changes during the contour evolution. If the segmentation algorithm adds or removes only \emph{simple points} from an initial segmentation, then the new segmentation will have the same genus as before. 

\begin{figure}
\begin{center}
\subfloat[]{\includegraphics[trim=0mm 0mm 0mm 0mm,width=.35\linewidth,clip=true]{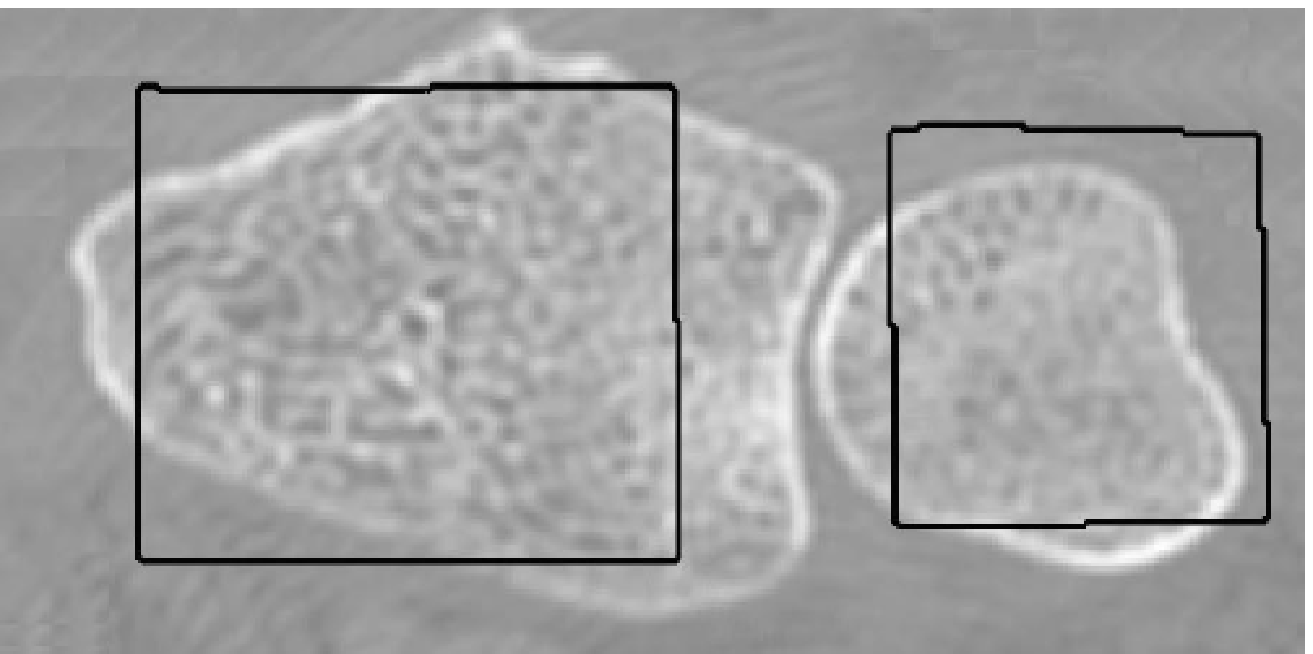}}~
\subfloat[]{\includegraphics[trim=0mm 0mm 0mm 0mm,width=.35\linewidth,clip=true]{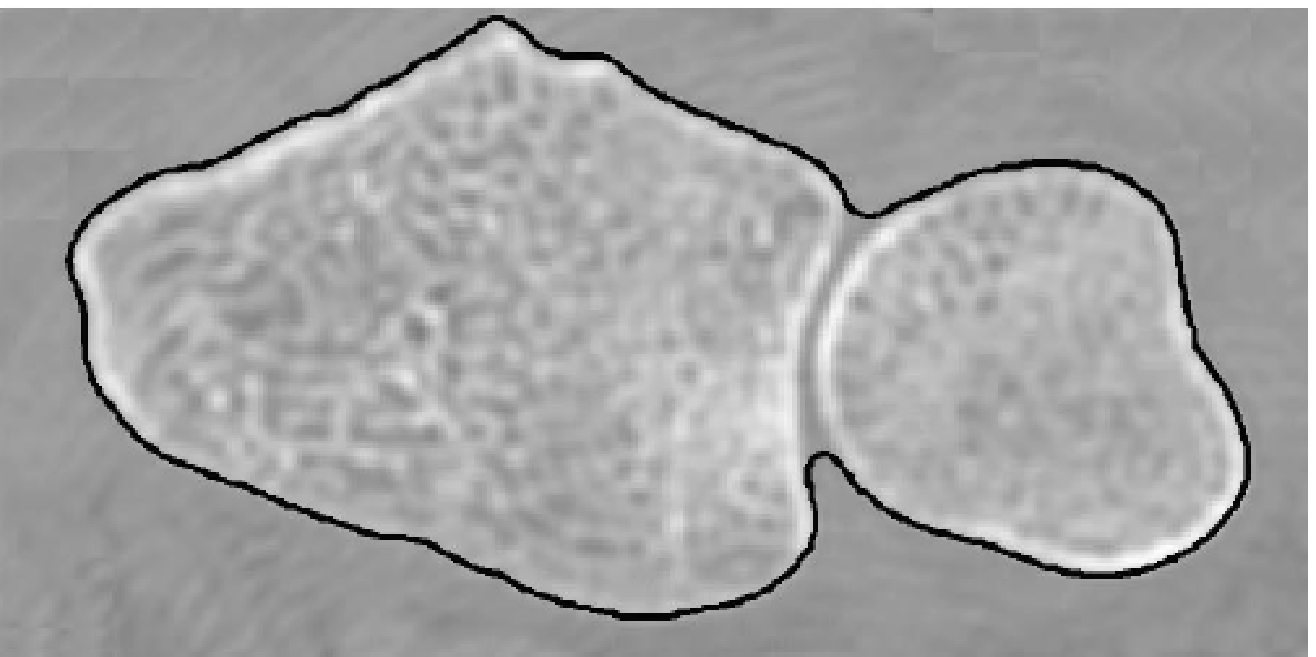}}~
\subfloat[]{\includegraphics[trim=0mm 0mm 0mm 0mm,width=.35\linewidth,clip=true]{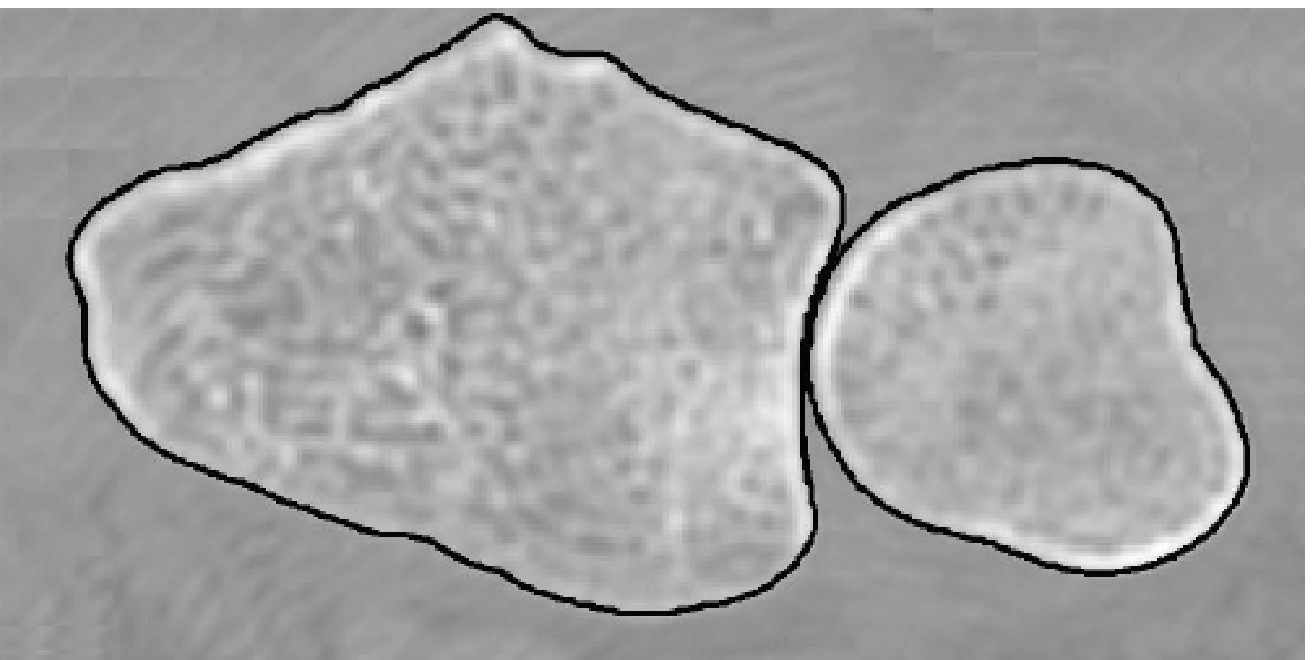}}\\
\subfloat[]{\includegraphics[trim=0mm 0mm 0mm 0mm,width=.35\linewidth,clip=true]{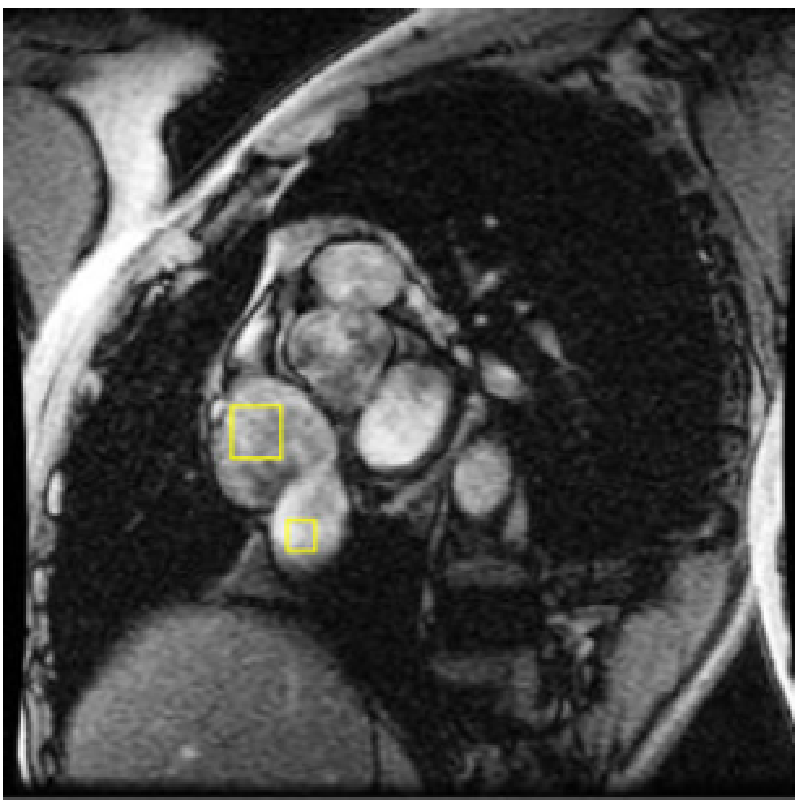}}~
\subfloat[]{\includegraphics[trim=0mm 0mm 0mm 0mm,width=.35\linewidth,clip=true]{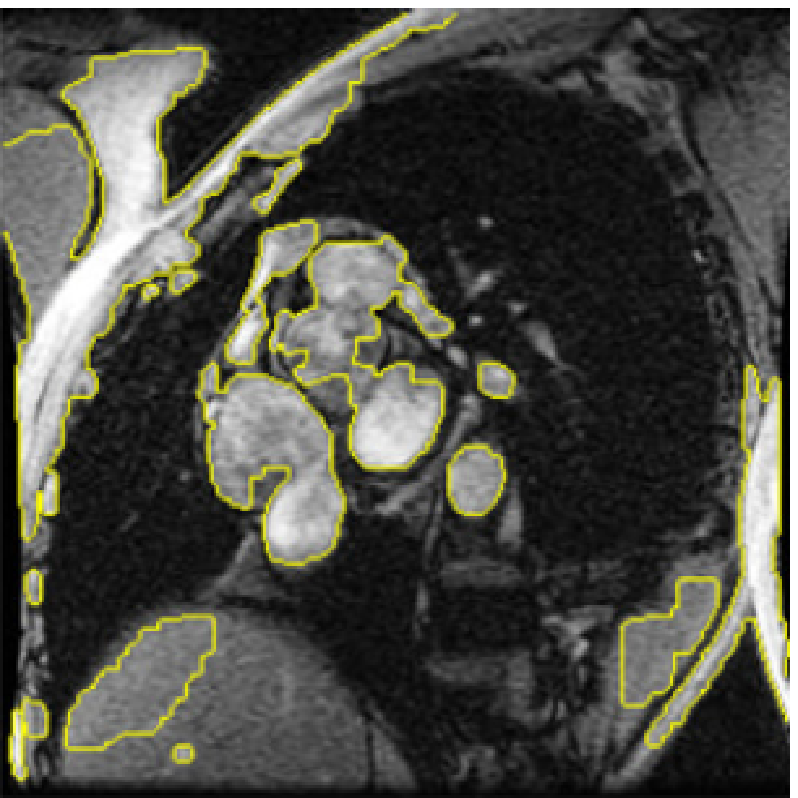}}~
\subfloat[]{\includegraphics[trim=0mm 0mm 0mm 0mm,width=.35\linewidth,clip=true]{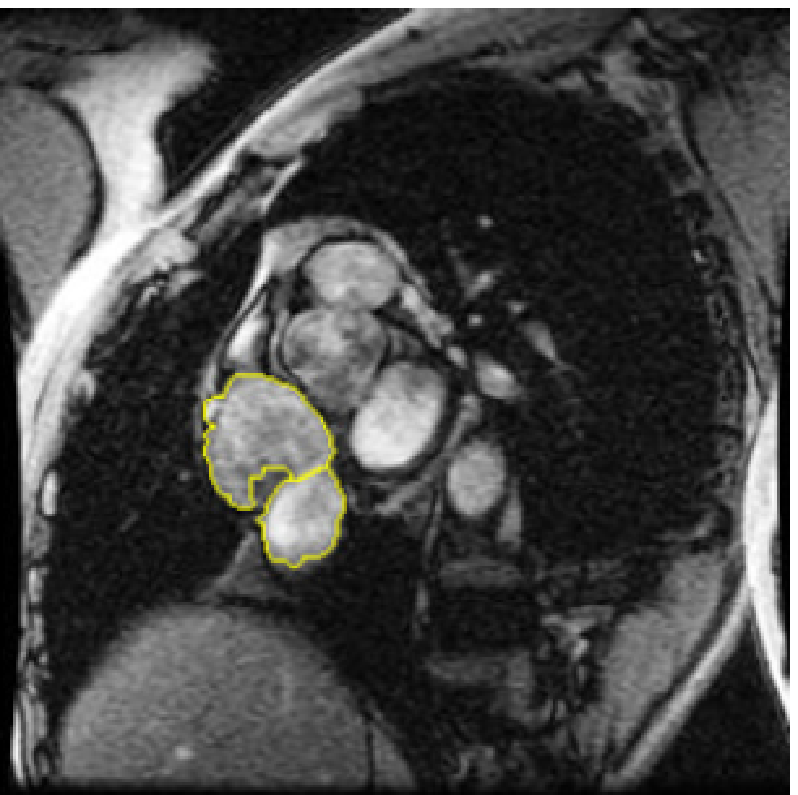}}
\end{center}
\caption{The effect of incorporating topological constraints into medical image segmentation frameworks. \textbf{Top row:} Segmentation of carpal bones in a CT image \citep{han2003topology}. \textbf{Bottom row:} Segmentation of cardiac ventricles in a MR image  \citep{zeng2008topology}. 
(a) Initialization. (b,c) Segmentation without (b) and with (c) topological constraints.}
\label{fig:topological}
\end{figure}

Inspired by \cite{han2003topology}, \cite{zeng2008topology} introduced \emph{topology cuts} and  cast the formulation of \cite{han2003topology} in a  discrete setting. They showed that the optimization of their energy functional with topology-preservation is NP-hard.
In another work, \cite{vicente2008graph} proposed an interactive method in the discrete domain to segment objects with topology-preservation. Their algorithm guarantees the connectivity between two designated points.  Further, the authors showed that their method can sometime find the global optimum under some conditions. Figure \ref{fig:topological} shows examples of encoding topological constraint in segmenting capral bones and cardiac ventricles.

\subsection{Moment prior}
In most segmentation methods that impose shape prior,  deviations of the observed shape from training shapes are usually suppressed by the shape prior imposed. This is undesirable in medical image segmentation where pathological cases occur (i.e. abnormal cases that deviate from the training shapes of healthy organs). Lower-order moment constraints can be an alternative to avoid  this limitation.

\begin{itemize}
\item \textbf{$0^{th}$ order moment (size/area/volume):}
The $0^{th}$ order moment corresponds to the size of an object. \cite{ayed2008area} proposed to add the area prior into the level set framework to speed up the curve evolution and to prevent leakage in the final segmentation. Given an image $I$ and the approximate area value of the targeted object ($\mathcal{A}$), their area energy term is defined as:
\begin{align}
E^{Area}(\bs{x})=\frac{1}{\mathcal{A}^2}\left(\int_{\Omega_{in}}d\bs{x}-\mathcal{A}\right)^2\int_{\Omega_{in}}g(I(\bs{x}))d\bs{x}
\;,
\end{align}
where $\Omega_{in}$ is the region inside the current segmentation and $g(.)$ attracts the evolving contour toward the high gradient regions (object boundaries).

\item \textbf{$1^{st}$ order moment (location/centroid):}
In case of having some rough information about the centroid of the targeted object,  this valuable information can be encoded into a segmentation framework using the $1^{st}$ order moment as proposed in \citep{klodt2011convex} (see below for more details).

\item \textbf{Higher-order moment:}
Generally, we can impose moment constraints of any order to refine the segmentation and capture fine-scale shape details. \cite{foulonneau2006affine} proposed to encode higher-order moments into a level set framework using a local optimization scheme. Recently,  \cite{klodt2011convex} proposed a convex formulation to encode moment constraints. They used the objective function in the form of
\begin{align}
E(u)=\int_\Omega \rho(\bs{x}) u(\bs{x})d\bs{x}+\int_\Omega g(\bs{x})|Du(\bs{x})|d\bs{x}
\;,
\label{Eu}
\end{align}
where $u\in BV:\mathbb{R}^d\rightarrow\{0,1\}$ is the labeling function and $Du$ is the distributional derivative ($Du(\bs{x})=\nabla u(\bs{x})$ for a differentiable $u$). Relaxing $u$ to vary between $0$ and $1$,   \eqref{Eu} becomes a convex optimization problem over the convex set $BV:\mathbb{R}^d\rightarrow [0,1]$. The global minimizer of the original problem ($E(u)$ before relaxing $u$) is obtained by finding the global minimum of the relaxed energy functional, $u^*$, and thresholding $u^*$ by a value $\mu\in(0,1)$.

\cite{klodt2011convex} imposed the $0^{th}$ order moment (i.e. area constraint in a 2D image) by bounding the area of $u$ between $c_1$ and $c_2$ where $c_1\leq c_2$ such that $u$ lies in the set
\begin{align}
\mathcal{C}_0=\left\{u\big|c_1\leq\int_\Omega ud\bs{x}\leq c_2\right\}.
\label{moment_1}
\end{align}
The exact area prior can be imposed by setting $c_1=c_2$. The $1^{st}$ moment (i.e. centroid constraint) is imposed by constraining the solution $u$ to be in the set $\mathcal{C}_1$ as:
\begin{align}
\mathcal{C}_1=\left\{u\big|\mu_1\leq \frac{\int_\Omega \bs{x}ud\bs{x}}{\int_\Omega ud\bs{x}}\leq \mu_2\right\}
\;,
\end{align}
where $\mu_1,\mu_2\in\mathbb{R}^d$. The set $\mathcal{C}_1$ ensures that the centroid of the segmented object lies between $\mu_1$ and $\mu_2$. The centroid is fixed when $\mu_1=\mu_2$.

In general, the $n^{th}$ order moment constraint is imposed  as:
\begin{align}
\mathcal{C}_n=\left\{u\big|A_1\leq \frac{\int_\Omega (x_1-\mu_1)^{i_1}\cdots(x_d-\mu_d)^{i_d} ud\bs{x}}{\int_\Omega ud\bs{x}}\leq A_2\right\}
\;,
\label{moment_n}
\end{align}
where $i_1+\cdots+i_d=n$, $A_1$, $A_2\in \mathbb{R}^{d\times d}$ are symmetric matrices and $A_1\leq A_2$ element wise. \cite{klodt2011convex} proved that all these sets are convex.  In their work, the above constraints \eqref{moment_1}-\eqref{moment_n}  are all hard constraints. Alternatively, all of the aforementioned constraints can be enforced as soft constraints by including them into the energy functional using Lagrange multipliers.  \cite{klodt2011convex} mentioned that, in practice,  imposing moments of more than the order of $2$ is not very useful as users cannot interpret these moments visually and the improvements are very small.
\end{itemize}

In the discrete settings, \cite{limconstrained} encode area, centroid and covariance ($2^{nd}$ order constraint) constraints into a graph-based method. While their method does not guarantee a globally optimal solution,  their method can impose non-linear combinations of the aforementioned constraints  as opposed to \citep{klodt2011convex}. 

\begin{figure}
\begin{center}
\subfloat[]{\includegraphics[trim=0mm 0mm 0mm 0mm,width=.3\linewidth,clip=true]{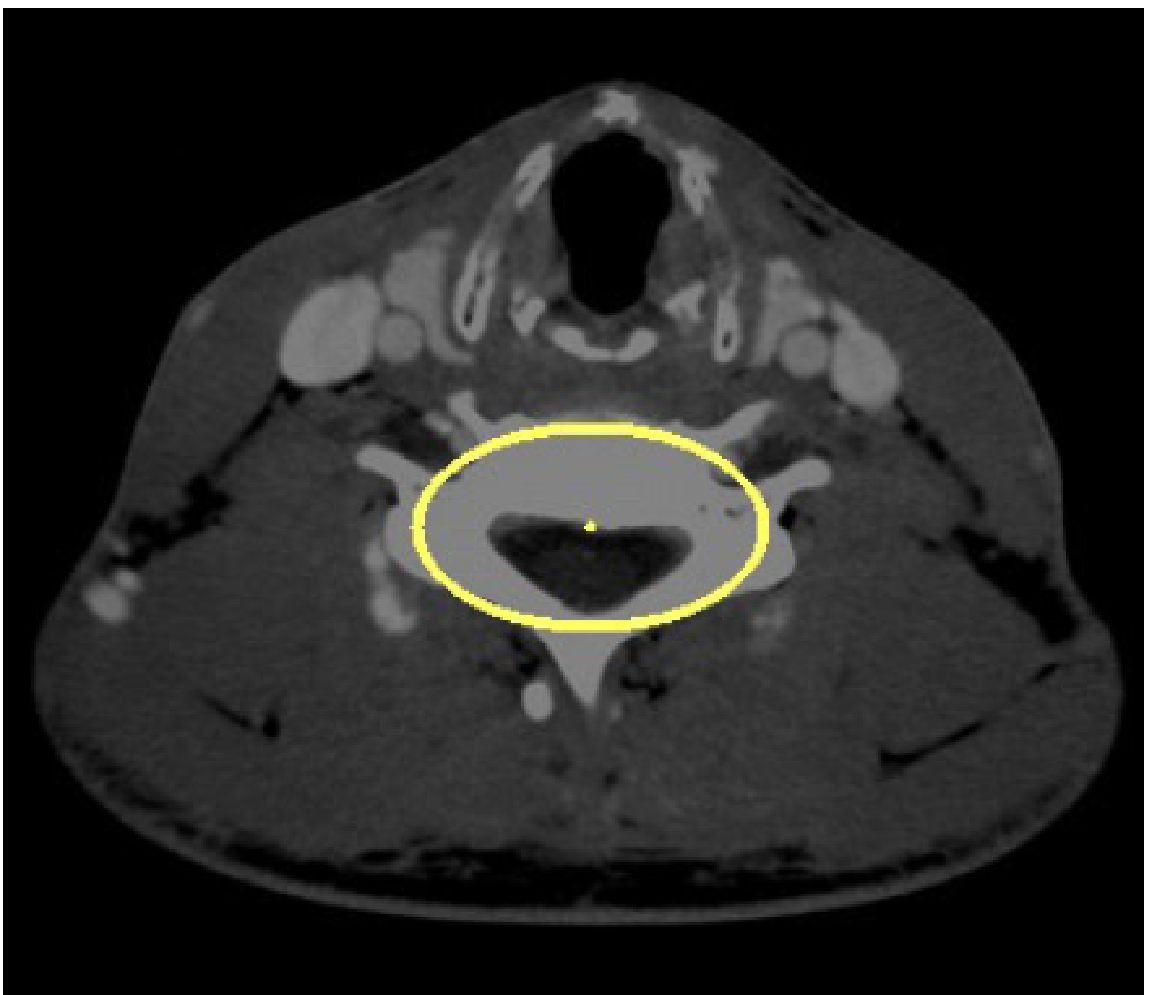}}~
\subfloat[]{\includegraphics[trim=0mm 0mm 0mm 0mm,width=.3\linewidth,clip=true]{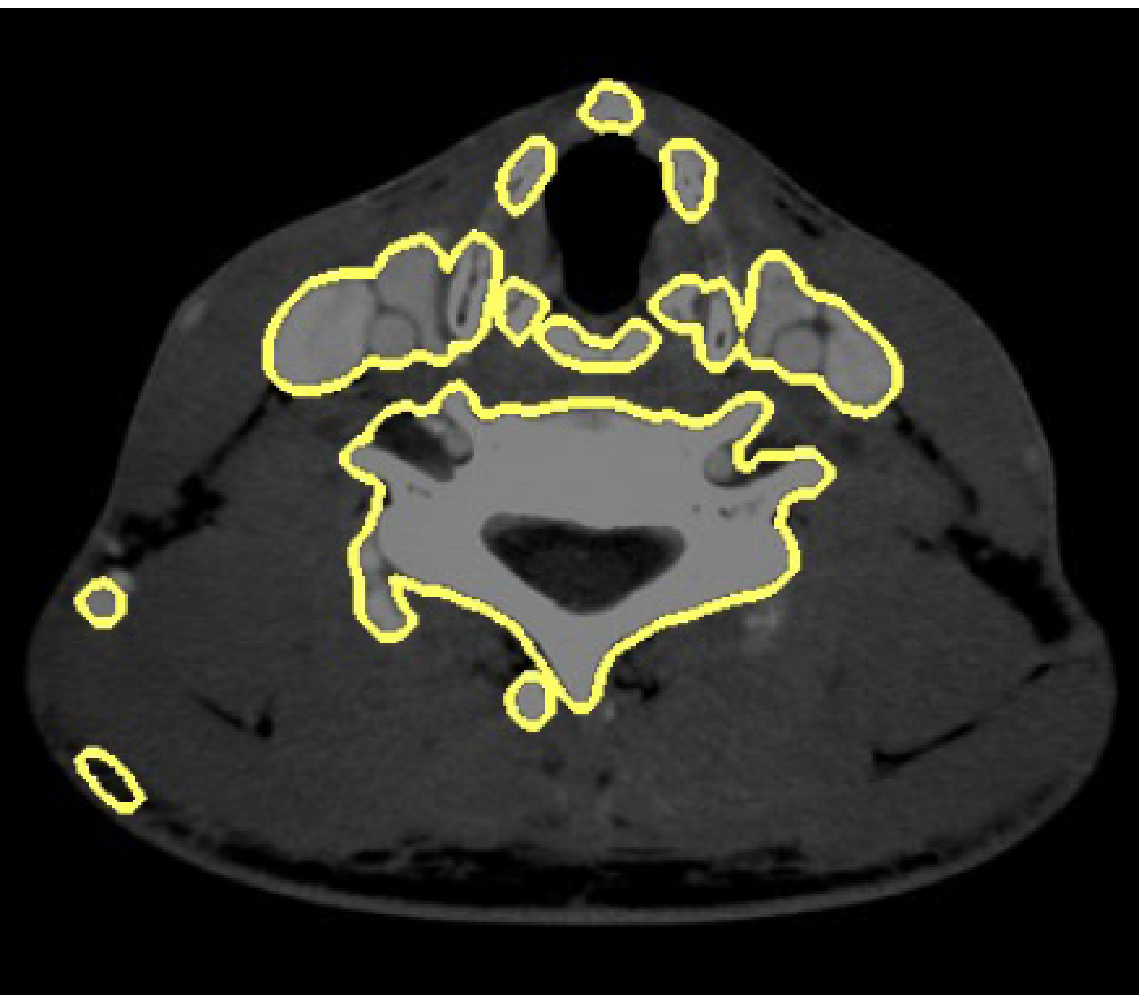}}~
\subfloat[]{\includegraphics[trim=0mm 0mm 0mm 0mm,width=.3\linewidth,clip=true]{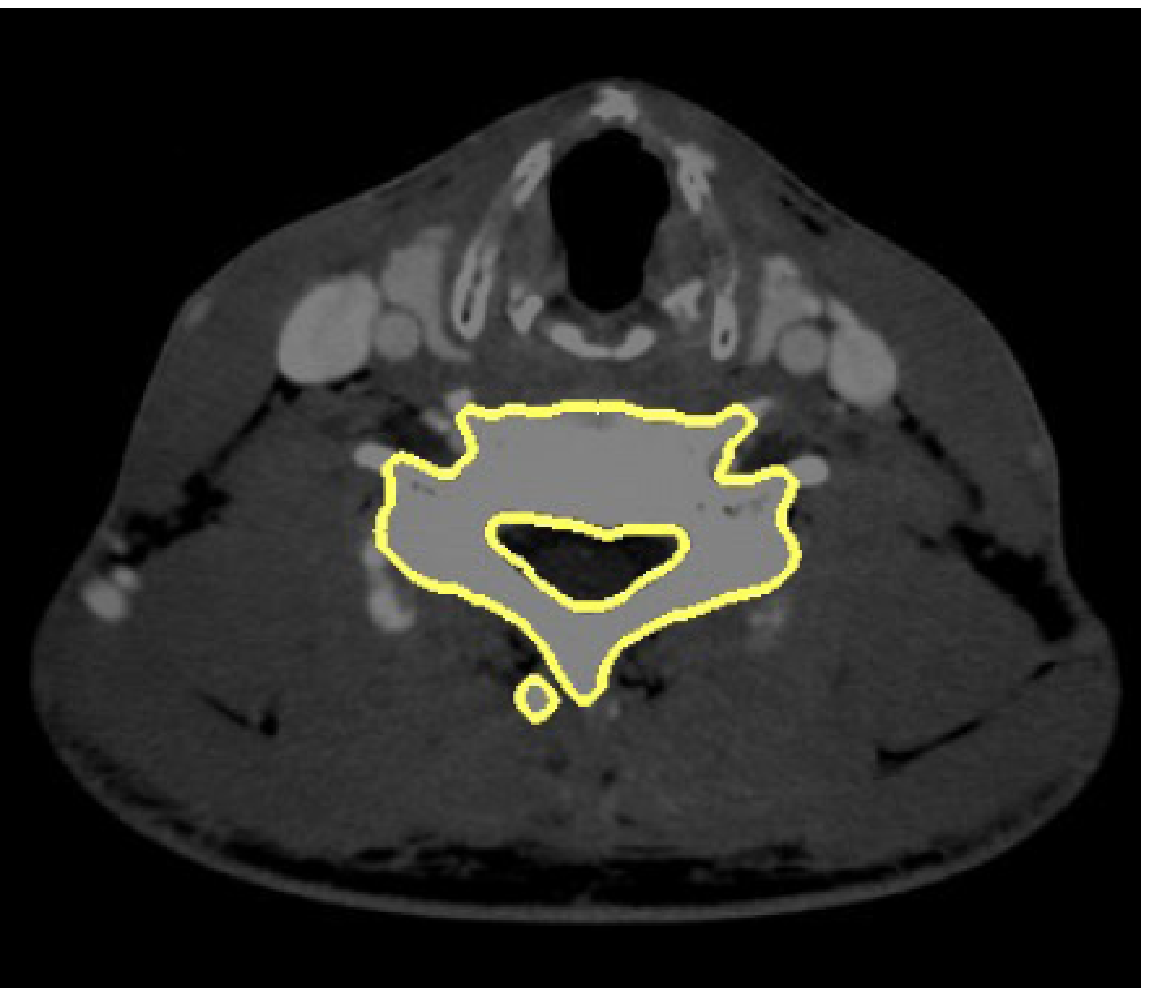}}
\end{center}
\caption{CT segmentation (b) without and (c) with moments constraints. The area constraint limits the segmentation to 
the size of the ellipse that was clicked by the user (a) that generates more accurate results.
(Images adopted from \cite{klodt2011convex})}
\label{fig:moments}
\end{figure}

Figure \ref{fig:moments} illustrates an example application of using moment constraints in CT segmentation.

\subsection{Geometrical and    region interactions prior}
Anatomical objects often consist of multiple regions, each
with a unique appearance model, and each has meaningful geometrical relationships or interactions with other regions of the object. Over the past decade, much attention has been given to
incorporating geometrical constraints into the segmentation
objective function. 

In the continuous domain,  several methods have been proposed based on coupled surfaces propagation to segment a single object in an image \citep{zeng1998volumetric,goldenberg2002cortex,paragios2002variational}.  \cite{vazquez2009multiphase} defined elastic coupling between multiple level set functions to model ribbon-like partitions. However, their approach was not designed to handle interactions between more than two regions. 

\begin{figure}[!t]
\begin{center}
\captionsetup[subfloat]{font=footnotesize,labelformat=empty}
\subfloat[(a) Original image]
{\includegraphics[trim=0mm 0mm 0mm 0mm,width=.25\linewidth,clip=true]{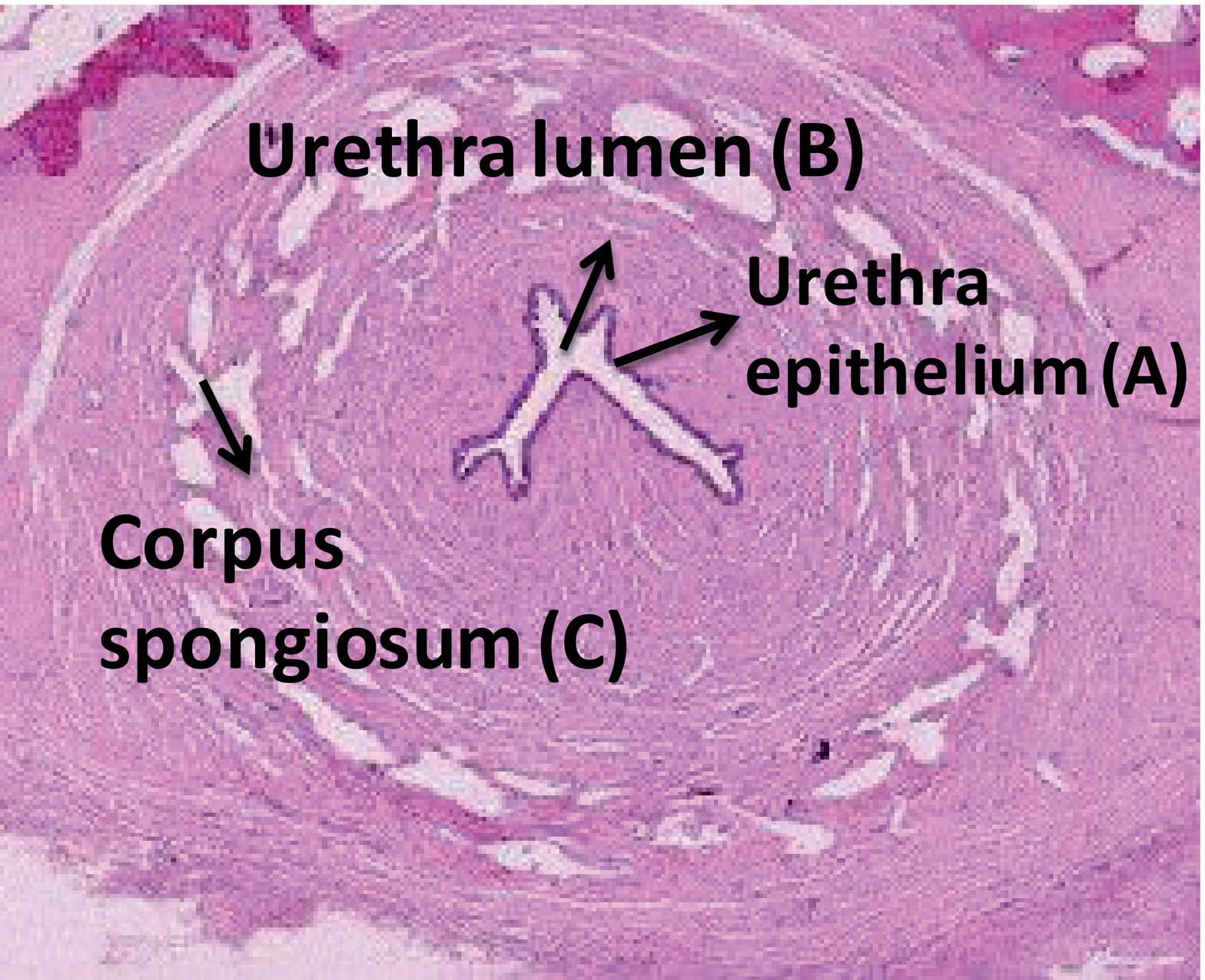}}~~
\subfloat[(b) Initialization]
{\includegraphics[trim=0mm 0mm 0mm 0mm,width=.25\linewidth,clip=true]{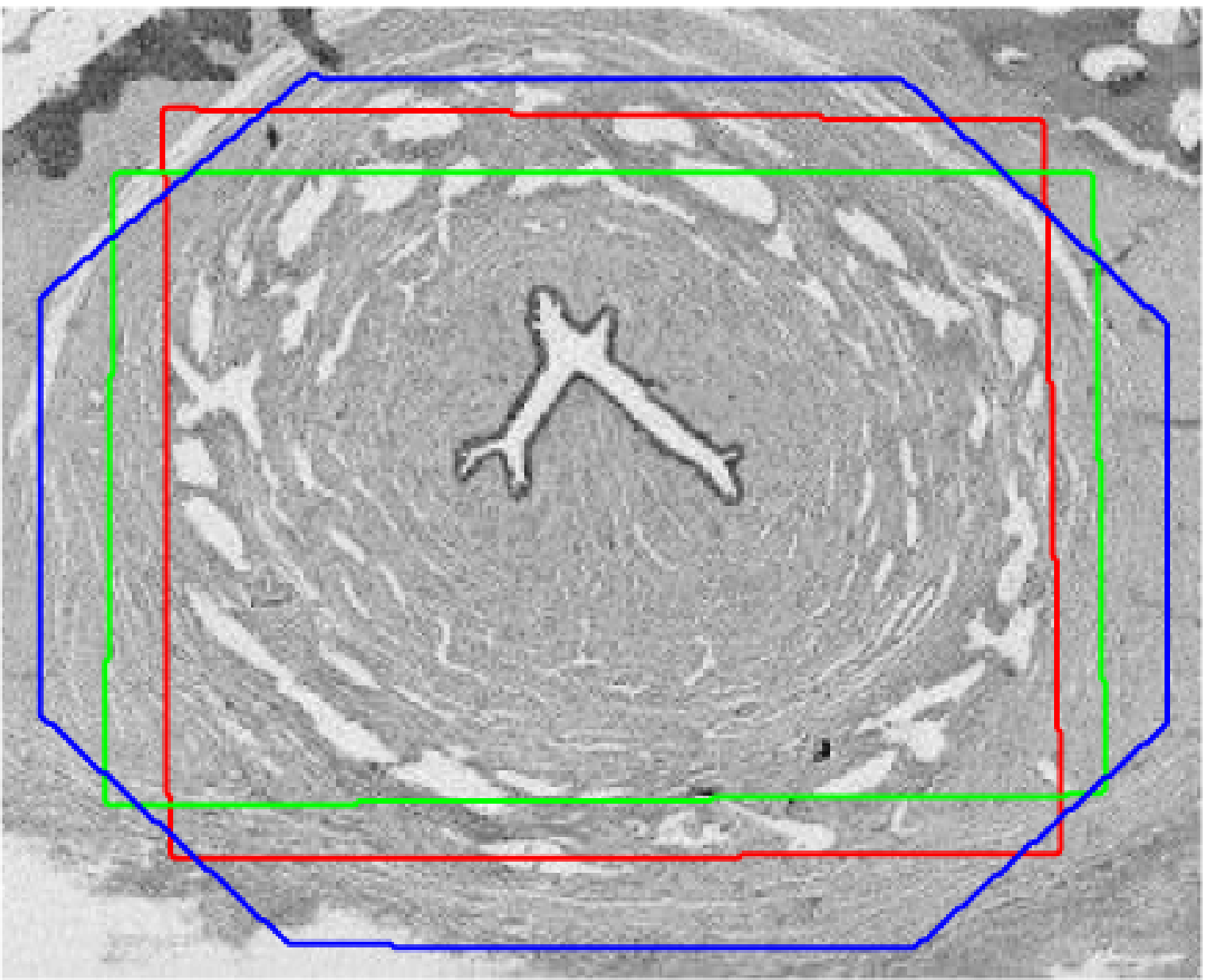}}~~
\subfloat[(c) No geometric constraint]
{\includegraphics[trim=0mm 0mm 0mm 0mm,width=.25\linewidth,clip=true]{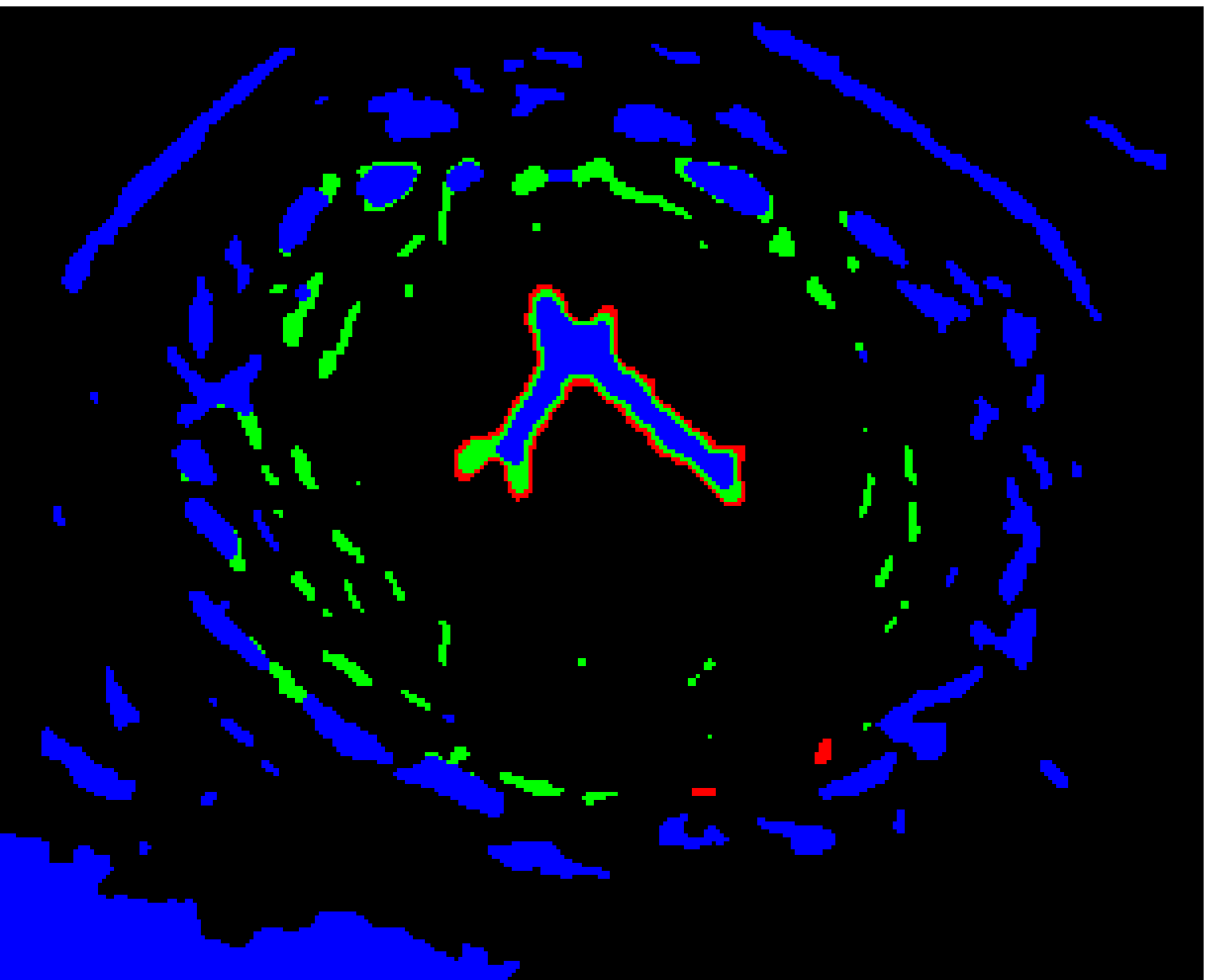}}~~
\subfloat[(d) With geometric constraints]
{\includegraphics[trim=0mm 0mm 0mm 0mm,width=.25\linewidth,clip=true]{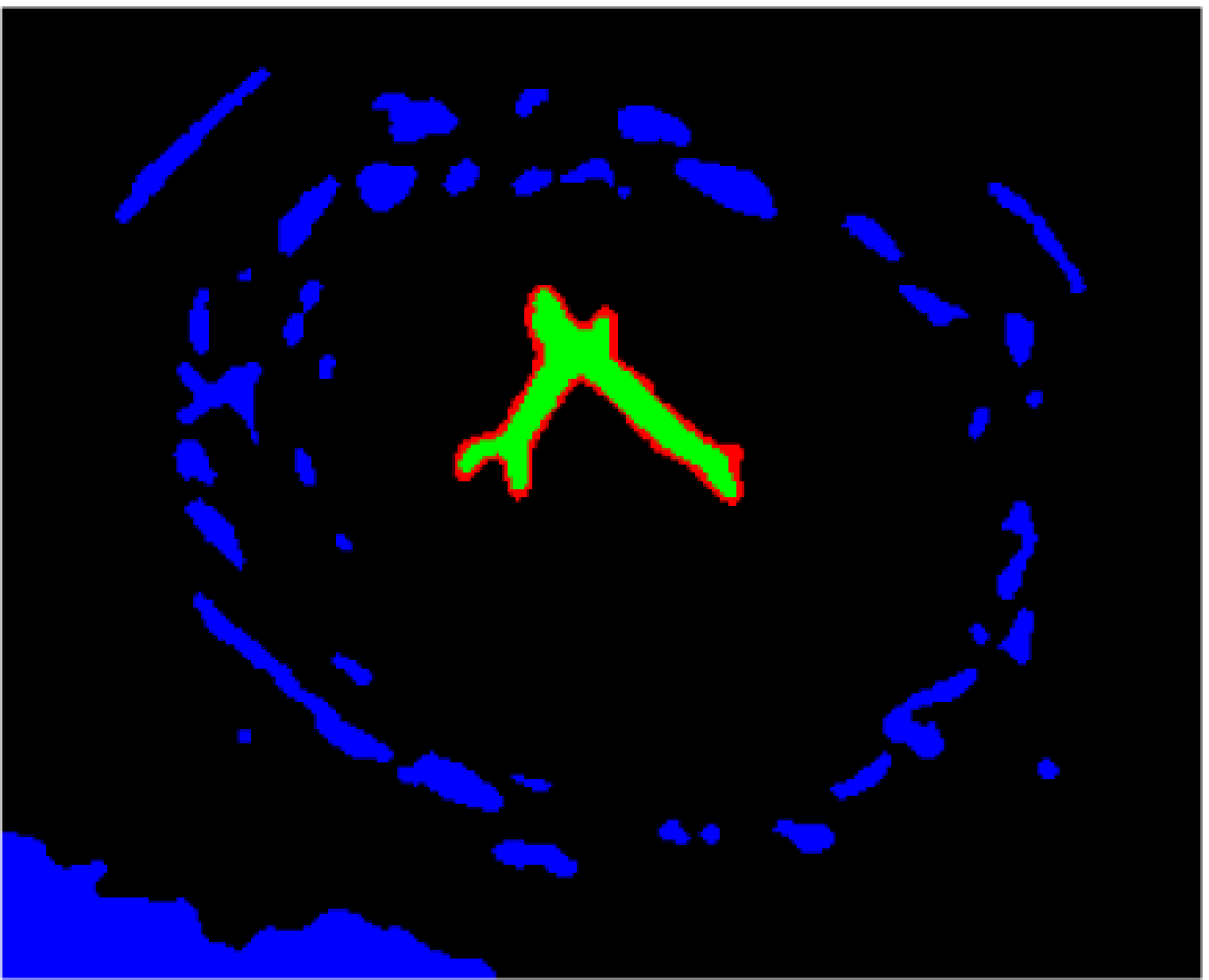}}\vspace{-2mm}
\end{center}
\caption{Urethra segmentation in a histology image. The constraint is set such that the urethra epithelium (A) contains the urethra lumen (B) and excludes the other regions with similar intensity with B, i.e. the corpus spongiosum (C). Here A, B and C are represented by red, green and blue colors, respectively. (Images adopted  from \citep{nosrati2014local})}
\label{fig:male}
\end{figure}

\begin{figure}[!t]
\begin{center}
\subfloat[]{\includegraphics[trim=0mm 0mm 0mm 0mm,width=.2\linewidth,clip=true,angle=90]{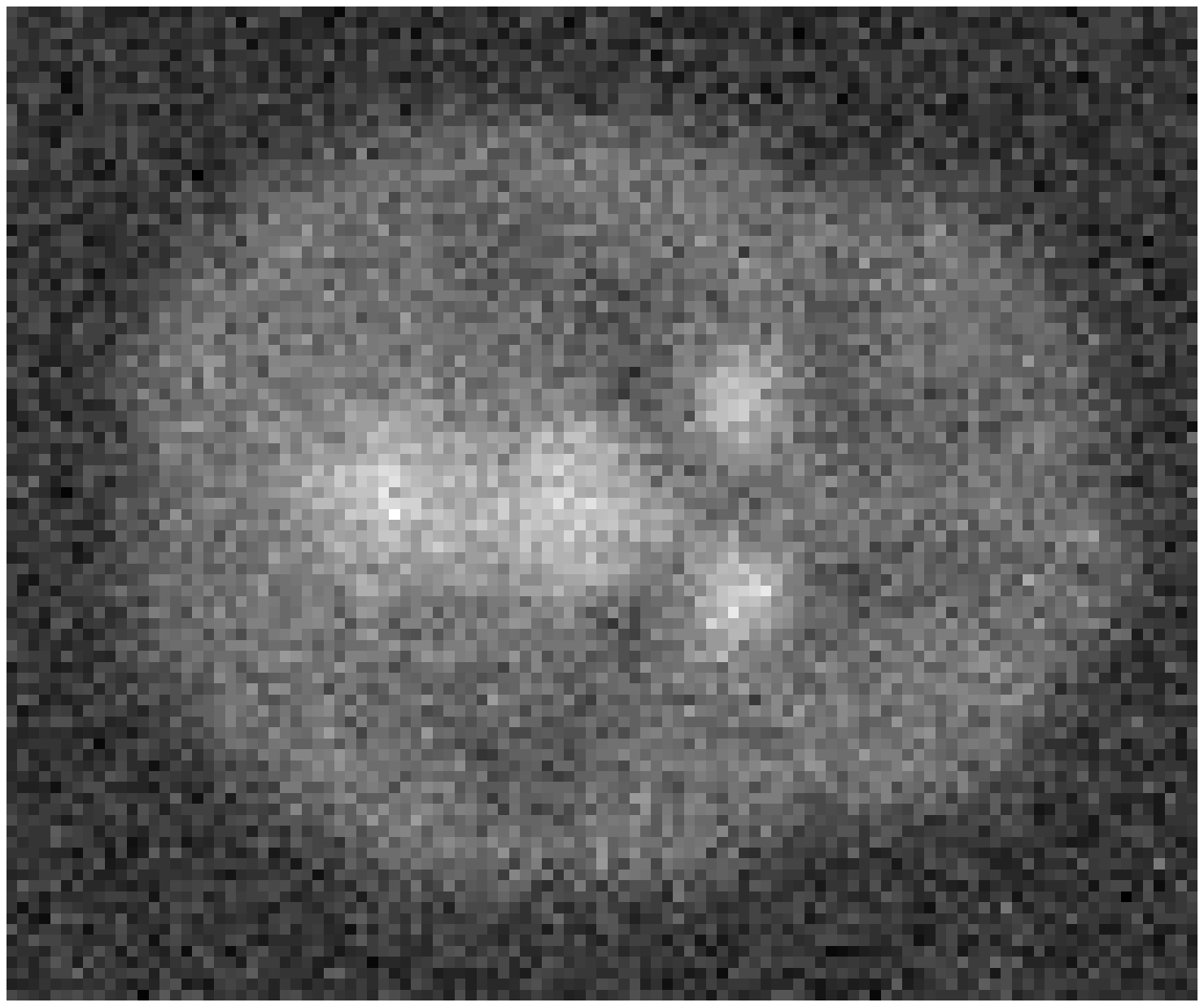}}~
\subfloat[]{\includegraphics[trim=0mm 0mm 0mm 0mm,width=.2\linewidth,clip=true,angle=90]{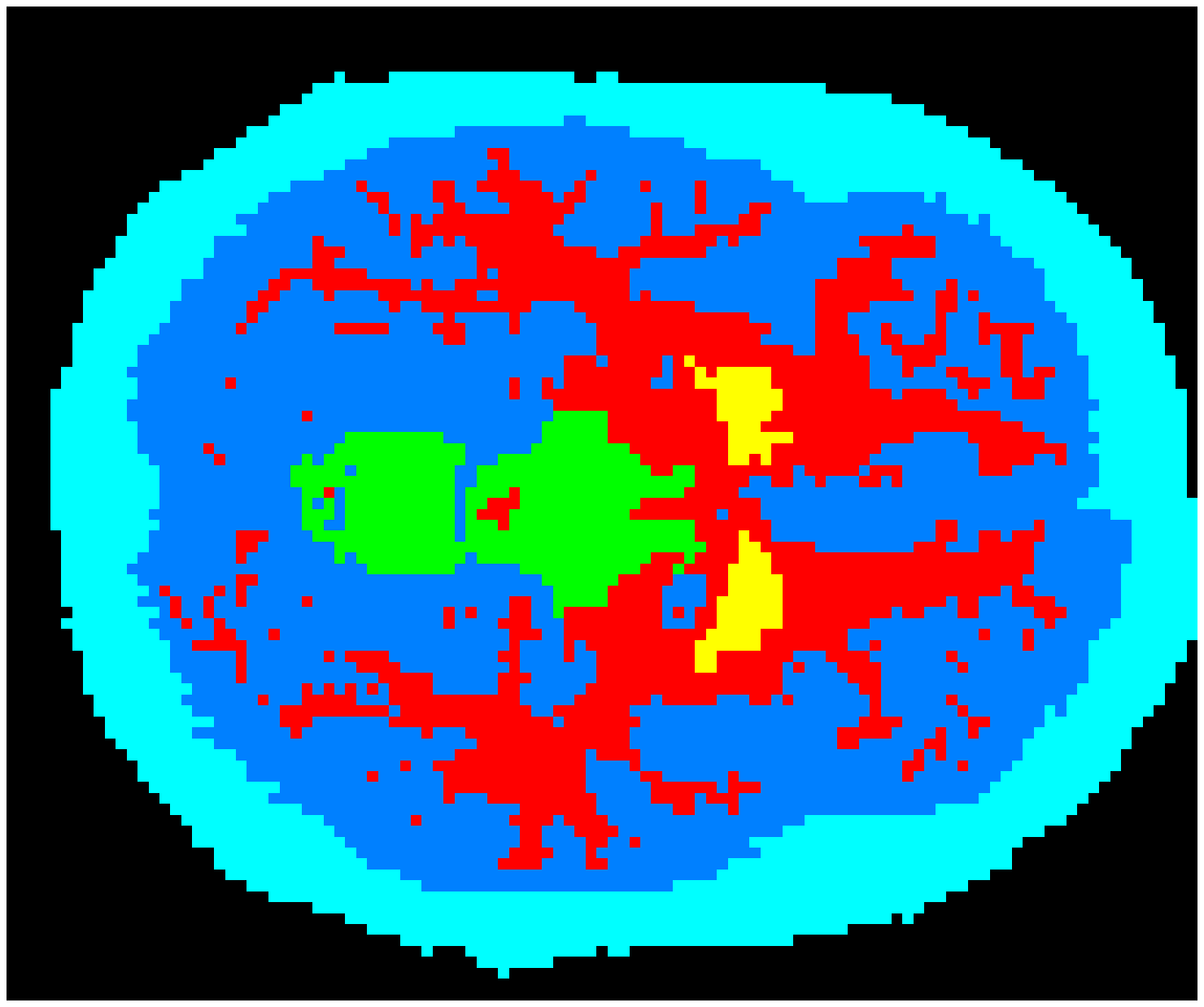}}~
\subfloat[]{\includegraphics[trim=0mm 0mm 0mm 0mm,width=.2\linewidth,clip=true,angle=90]{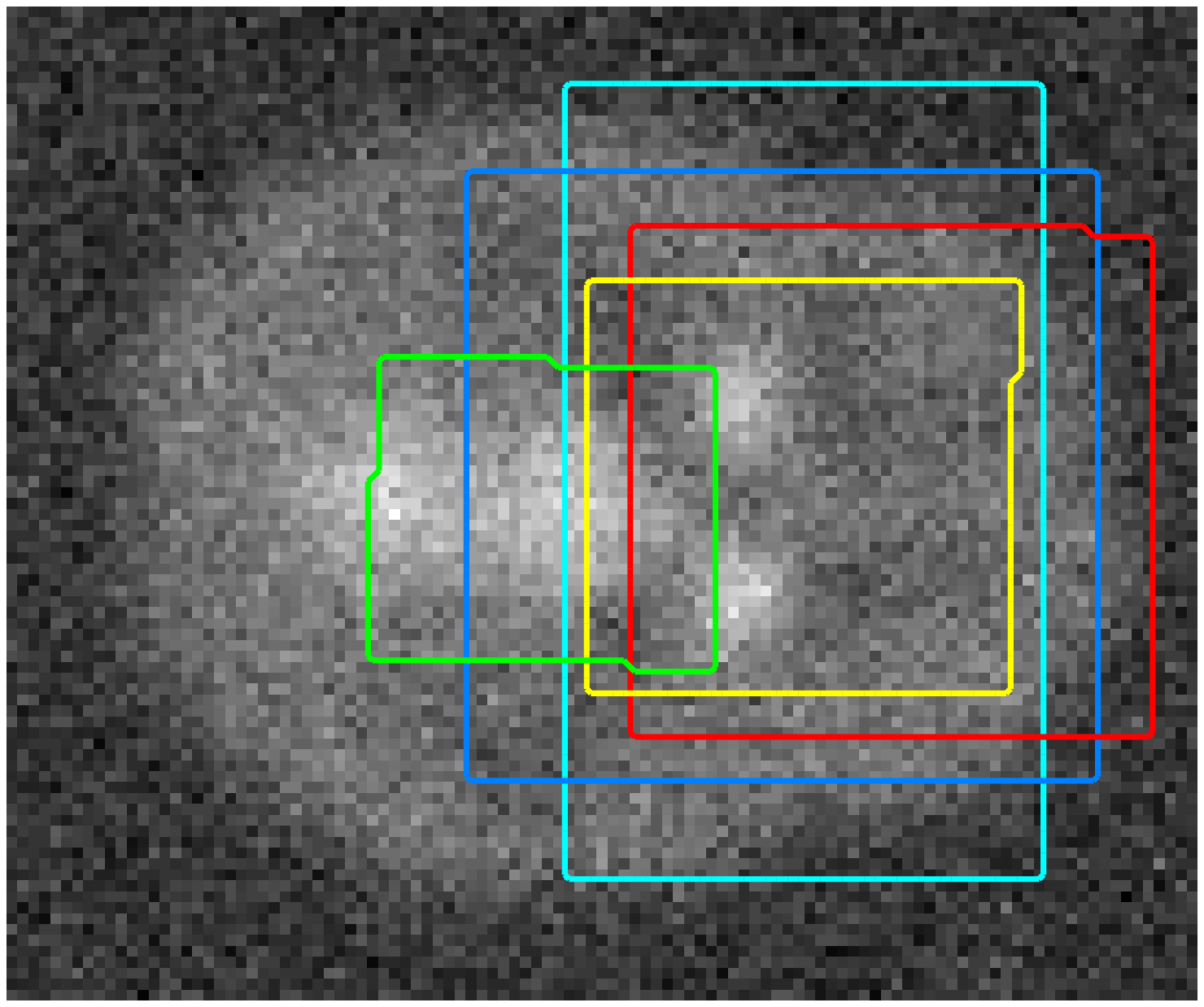}}~
\subfloat[]{\includegraphics[trim=0mm 0mm 0mm 0mm,width=.2\linewidth,clip=true,angle=90]{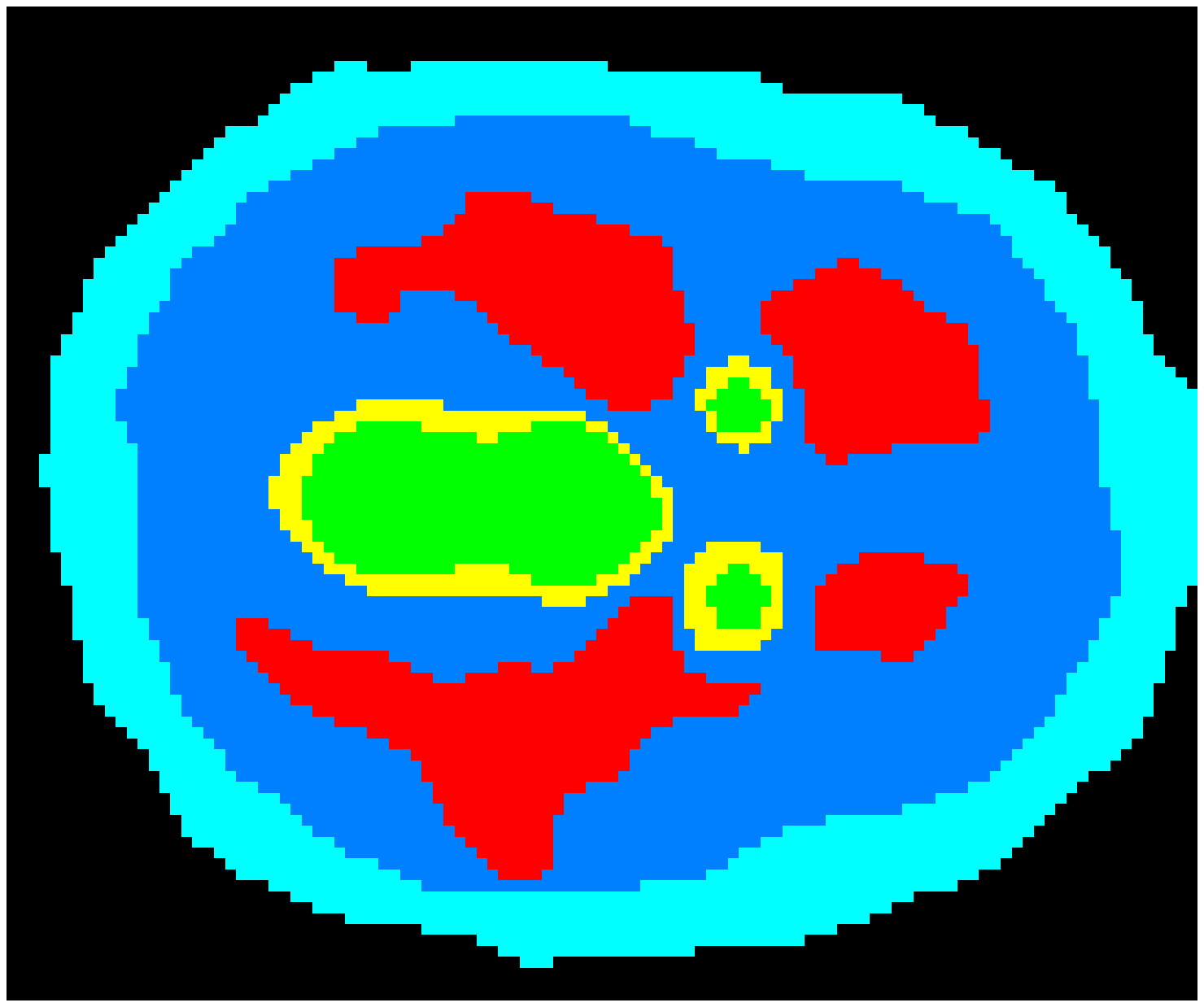}}~%pet_noContainment_n5
%{\includegraphics[trim=0mm 0mm 0mm 0mm,width=.2\linewidth,clip=true,angle=90]{./Figures/NOSRApet_mrf.eps}}
%\subfloat[]{\includegraphics[trim=7mm 0mm 7mm 0mm,width=.222\linewidth,clip=true]{./Figures/NOSRATMI_ahmed6.eps}}~
\subfloat[]{\includegraphics[trim=0mm 0mm 0mm 0mm,width=.2\linewidth,clip=true,angle=90]{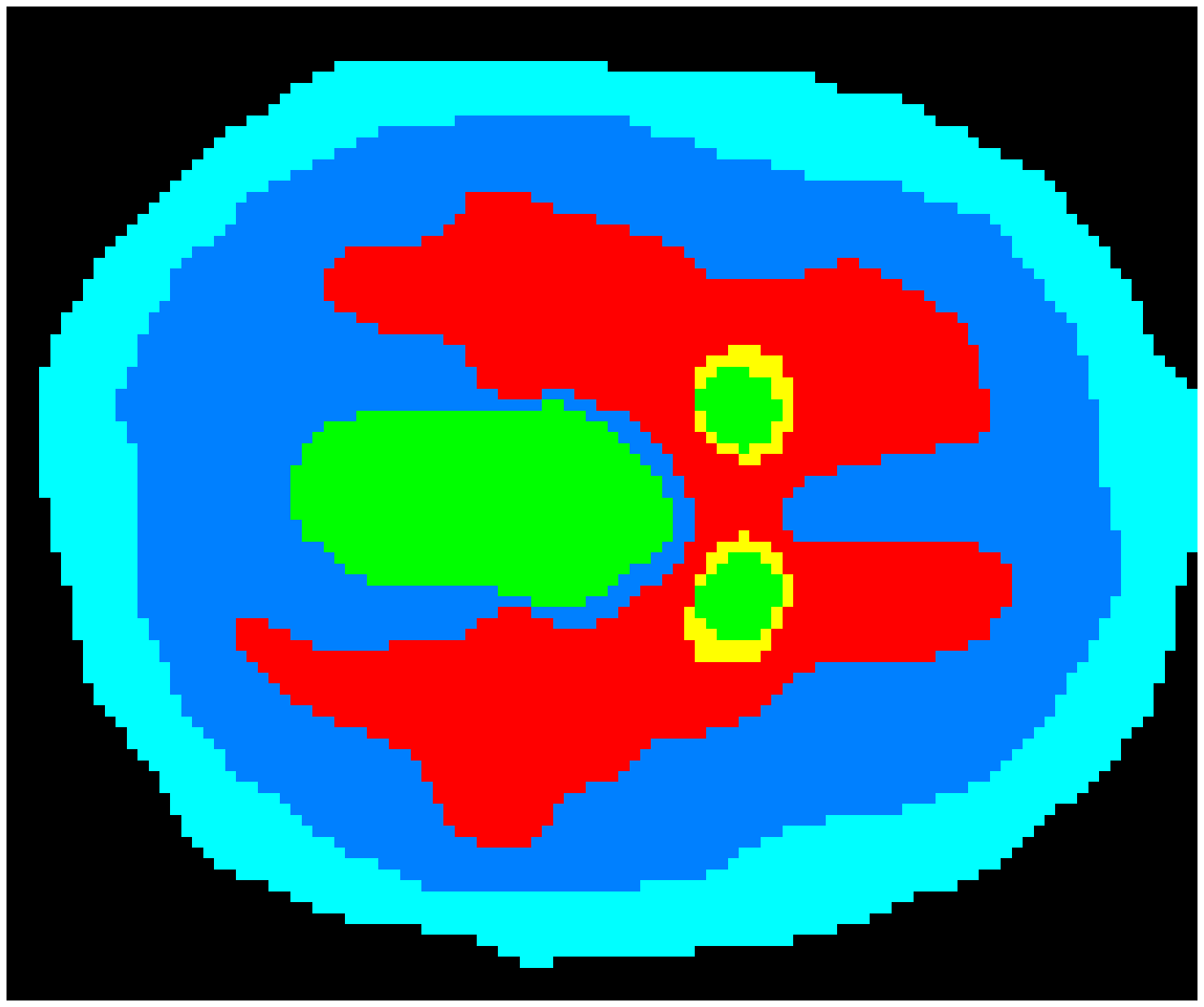}}\\
\subfloat[]{\includegraphics[trim=0mm 0mm 0mm 0mm,width=.2\linewidth,clip=true,angle=90]{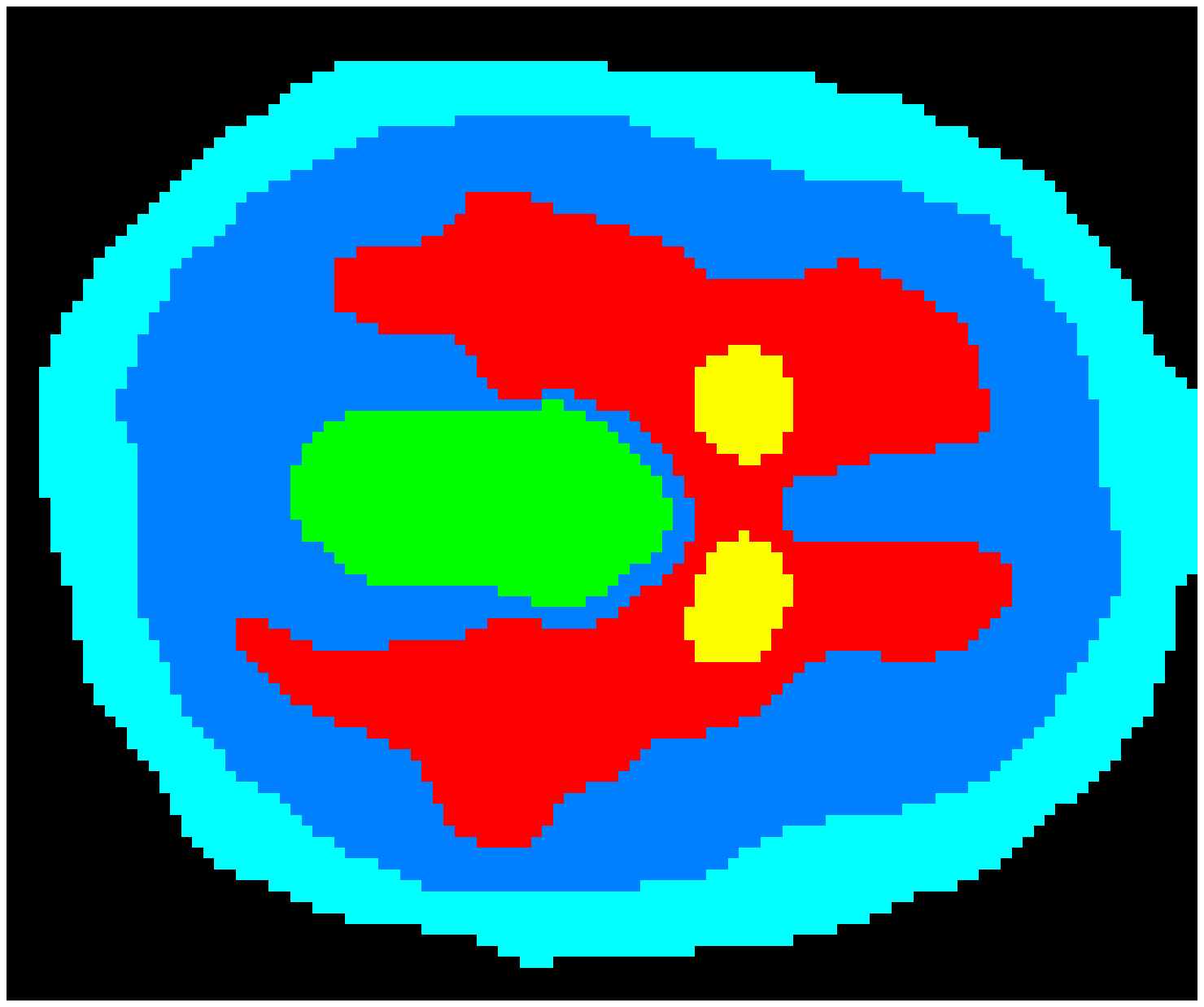}}~~%TMI_pet_result_n5
\subfloat[]{\includegraphics[trim=0mm 0mm 0mm 0mm,width=.75\linewidth,clip=true]{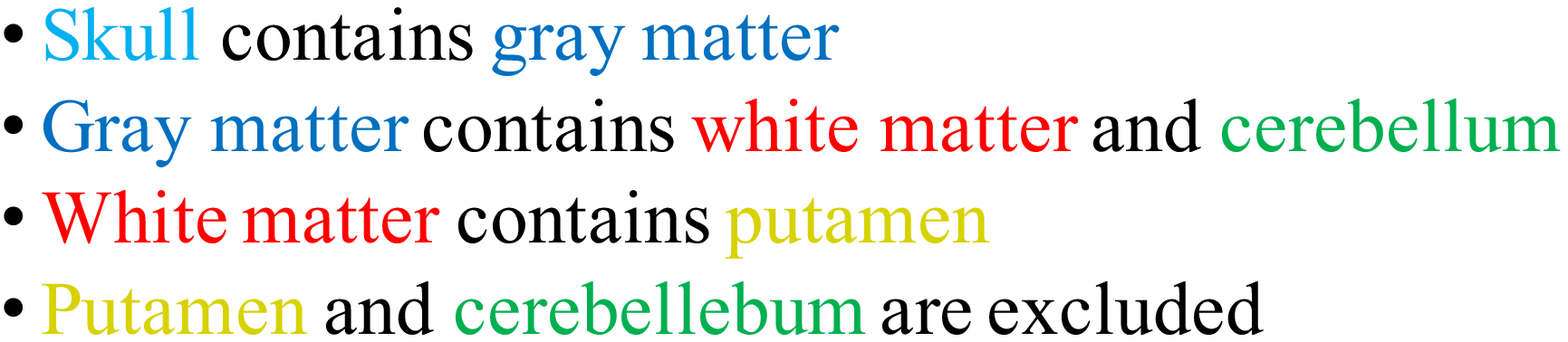}}
\end{center}
\caption{Brain dPET segmentation. (a) Raw image. (b) Ground truth. (c) Initialization. (d) No geometric constraint. (e) Result with containment but without exclusion constraint. (f) Result with containment and exclusion constraints. (g) Color coding of the employed geometric constraints. Note how the putamen is contained by the white matter (red) as it should be, whereas (d) and (e) are anatomically incorrect.
(Images from \citep{nosrati2014local})}
\label{fig:pet}
\end{figure}

\cite{nosrati2014local} augmented the level set
framework with the ability to handle two important and intuitive geometric relationships, \emph{containment} and \emph{exclusion}, along with a distance constraint between boundaries of multi-region objects  (Figures \ref{fig:male} and \ref{fig:pet}). Level set’s
important property of automatically handling topological changes of evolving contours/surfaces enables them to segment spatially recurring objects (e.g. multiple instances of multi-region cells in a large microscopy image) while satisfying the two aforementioned constraints.

\cite{bloch2005fuzzy} briefly reviewed the main fuzzy approaches that define spatial relationships  including \emph{topological relations} (i.e. set relationships and  adjacency) as well as \emph{metrical relations} (i.e. distances and  directional relative position).

None of the aforementioned methods guarantee a globally  optimal solution. In contrast,  \cite{wu2011region} proposed a method that yields globally optimal solution for segmenting a region bounded by two coupled terrain-like surfaces. They do so  by minimizing the intraclass variance.  Despite  the global optimality of their solutions, their method can only segment a single object in an image and is limited to handling objects that can be ``unfolded'' into two coupled
surfaces.

\cite{ukwatta2012efficient} also proposed a method that is based on coupling two surfaces for carotid adventitia (AB) and lumenintima (LIB) segmentation. The advantage of their work over previous works is that they optimized their energy functional by means of convex relaxation. However, their method could only segment objects with coupled surfaces. Using the same framework as \citep{ukwatta2012efficient},  \cite{rajchl2012fast2} presented a graphical model to segment the myocardium, blood cavities and scar tissue. Their method used seed points as hard constraints to distinguish the background from the myocardium.   \cite{nambakhsh2013left} proposed an efficient method for left ventricle (LV) segmentation
that iteratively minimizes a convex upper bound energy
functional for a coupled surface. Their method implicitly
imposes a distance between two surfaces by learning the LV shape. Recently, \cite{nosrati2013bounded} proposed a convex formulation to incorporate containment and detachment constraints between different regions with a specified minimum distance between their boundaries. Their framework used a single labeling function to encode such constraints while maintaining global optimality. They showed that their convex continuous method is superior to other state-of-the art methods, including its discrete counterpart, in terms of
memory usage and metrication errors.

\begin{figure}
\begin{center}
\subfloat[]{\includegraphics[trim=0mm 0mm 0mm 0mm,width=.3\linewidth,clip=true]{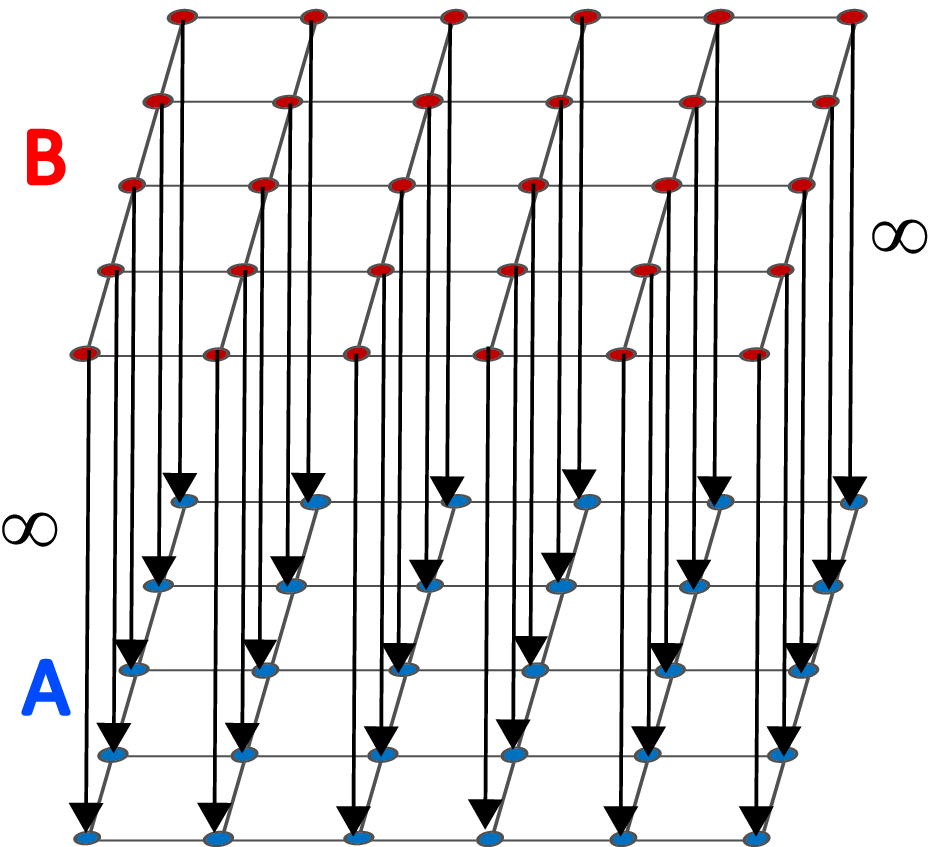}}~~~~~~~~~~
\subfloat[]{\includegraphics[trim=0mm 0mm 0mm 0mm,width=.3\linewidth,clip=true]{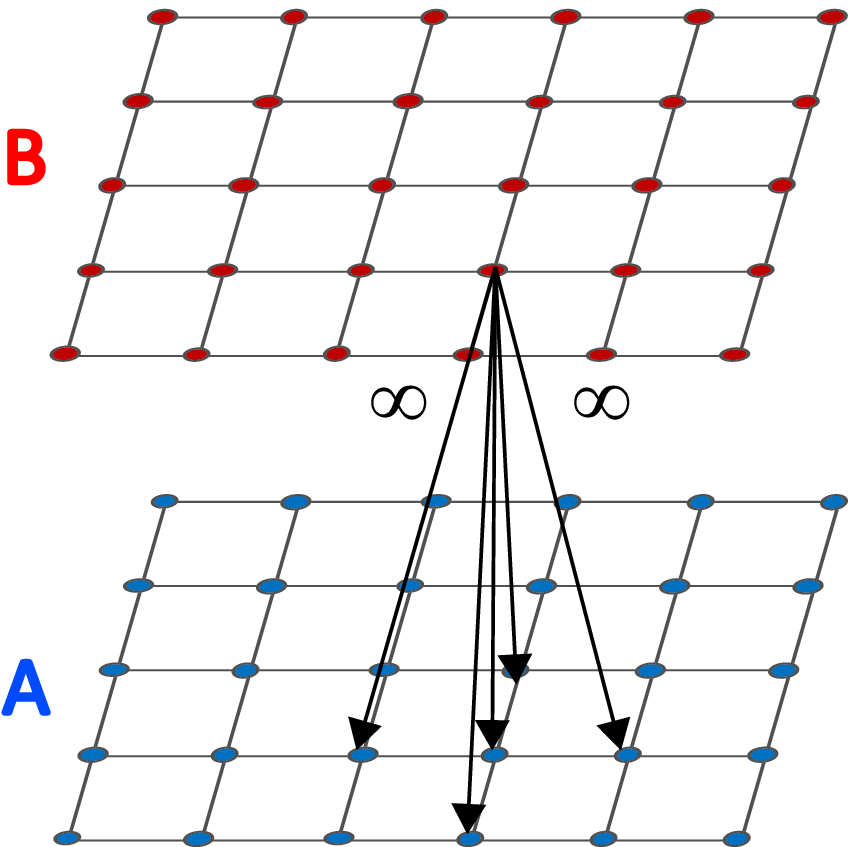}}
\end{center}
\caption{Enforcing containment constraint between objects A and B. Graph constructions used to enforce (a) `A contains B' as proposed by \citep{delong2009Globally}, and  (b) 1-pixel distance between boundaries of regions A and B (shown for a single pixel), respectively.}
\label{fig:delong}
\end{figure}

In the discrete domain, \cite{li2006optimal} proposed a method to segment ``nested objects'' by defining distance constraints between the object's surfaces with respect to a center point. As their formulation employed polar coordinates, their method could
only handle star-shaped objects. Two containment and exclusion constraints between distinct regions have been encoded into a graph-cut framework by  \citep{delong2009Globally} and \citep{ulen2013efficient}. If only containment constraint is enforced, then both approaches  guarantee the global solution. More formally, for a two-region object scenario (region A, B and background), the idea of \citep{delong2009Globally} is to create two graphs for A and B, i.e. $G(V^A,E^A)$ and $G(V^B, E^B)$. The segmentations of A and B are represented by the binary variables $x^A$ and $x^B$, respectively.  \note{The geometrical constraints between regions $A$ and $B$ are enforced by adding an additional penalty term $W^{AB}$ defined in Table \ref{DBtable}. This interaction term, $W$, is implemented in the graph construction by adding inter-layer edges with  infinity values as shown in  Figure \ref{fig:delong}(a).} %In addition to data and regularization terms, 
% The new term interaction term $W$ is added to the data term and regularization term, 
\cite{delong2009Globally} employed what is known as the interaction term $W$ as follows:
\begin{align}
\sum_{pq\in\mathcal{N}^{AB}}W^{AB}_{pq}(f^A_p,f^B_q)
\;,
\label{DB}
\end{align}
where $\mathcal{N}^{AB}$ is the set of all pixel pairs $(p,q)$ at which region $A$ is assigned some geometric interaction with region $B$. Table \ref{DBtable} lists energy terms for the region interaction constraints proposed in \citep{delong2009Globally}. Since the energy terms for containment are submodular, their graph-based method guarantees the globally optimal solution if only containment terms are used. However, the energy for the exclusion constraint is  nonsubmodular and thus harder to optimize. In some cases, because
exclusion is nonsubmodular everywhere, it is possible to make all these nonsubmodular terms submodular by flipping the meaning of layer B's variables so that $f^B_p=0$ designates the region's interior. Nonetheless, there are many useful interaction energy terms that cannot be modelled and optimized efficiently by \citep{delong2009Globally} and other approximation like quadratic pseudo-boolean optimization (QPBO) \citep{kolmogorov2007minimizing,rother2007optimizing} or $\alpha\beta$-swap \citep{boykov2001fast} should be used for the optimization of these terms.

%%%%%%%%%%%%%%%%%%%%%%%%%%%%%%%%%%%%%%%%
%%%%%%%%%%%%%%%%%%%%%%%%%%%%%%%%%%%%%%%%

\begin{table}
\caption{Energy terms for encoding containment and exclusion constraints between regions A and B in \eqref{DB} \citep{delong2009Globally}.}
\label{DBtable}
\centering
\begin{tabular}{|cc|c|}
\hline
\multicolumn{3}{ |c| }{A \textit{contains} B} \\
\hline
$f^A_p$&$f^B_q$&$W^{AB}_{pq}$\\
\hline
0&0&0\\
0&1&$\infty$\\
1&0&0\\
1&1&0\\
\hline
\end{tabular}
\quad\quad
\begin{tabular}{|cc|c|}
\hline
\multicolumn{3}{ |c| }{A \textit{excludes} B} \\
\hline
$f^A_p$&$f^B_q$&$W^{AB}_{pq}$\\
\hline
0&0&0\\
0&1&0\\
1&0&0\\
1&1&$\infty$\\
\hline
\end{tabular}
%\quad
%\quad
%\begin{tabular}{|cc|c|}
%\hline
%\multicolumn{3}{ |c| }{A \textit{attracts} B} \\
%\hline
%$x^A_p$&$x^B_q$&$W^{AB}_{pq}$\\
%\hline
%0&0&0\\
%0&1&0\\
%1&0&$\alpha$\\
%1&1&0\\
%\hline
%\end{tabular}
\end{table}
   
\subsection{Spatial distance prior}
In the literature, works that incorporate  spatial distance priors may be categorized as follows:
\begin{itemize}
\item\textbf{Minimum distance:} In some applications the minimum distance between two structures must be enforced to ensure that sufficient separation between regions exists to obtain plausible results (e.g. distance between carotid AB and LIB). Examples of methods that employ this constraint include  \citep{zeng1998volumetric,goldenberg2002cortex,paragios2002variational, nosrati2014local, nosrati2013bounded} in the continuous settings, and  \citep{wu2011region,li2006optimal,delong2009Globally,ulen2013efficient} in the discrete settings. Looking at \eqref{DB} for example, Delong and Boykov \citep{delong2009Globally} (and similarly, Ul\'en et al. \citep{ulen2013efficient}) enforce the minimum distance between two regions by defining the  $\mathcal{N}^{AB}$ in \eqref{DB}. Figure \ref{fig:delong}(b) shows how 1-pixel margin between region boundaries is enforced by \citep{delong2009Globally,ulen2013efficient}.

\item\textbf{Maximum distance:} In other medical applications, maximum distance between regions is known a priori.  For example, in cardiac LV segmentation, maximum distance between LV and its myocardium can be approximated. Enforcing a maximum distance between LV and its myocardium boundaries prevents the myocardium segmentation from growing too far from the LV. Maximum distance between two boundaries/surfaces is enforced as proposed in \citep{zeng1998volumetric,goldenberg2002cortex,paragios2002variational, nosrati2014local} in the continuous settings. There is not much work on incorporating  maximum distance between region boundaries in discrete settings except for  \cite{wu2011region} and \cite{schmidt2012hausdorff}. In \citep{wu2011region} the maximum distance along with minimum distance prior for segmenting two-region ribbon-like objects can be enforced. To the best of our knowledge, the only work that solely focused on incorporating maximum distance between regions for multi-region object segmentation is the approach proposed by \citep{schmidt2012hausdorff}. They modified the framework of \citep{delong2009Globally} by adding the Hausdorff distance prior to the MRF-based segmentation framework to impose maximum distance constraints. They showed that incorporating this prior into multi-surface segmentation is NP-hard due to the existence of supermodular energy terms.
 
\item\textbf{Attraction/repulsion distance:} In applications like multi-region cell segmentation,  distance between regions should be in a specific range.  A specific distance between different regions can be maintained by enforcing attraction and repulsion forces between region boundaries as proposed in  \citep{zeng1998volumetric,goldenberg2002cortex,paragios2002variational, nosrati2014local}   in the continuous settings.  \cite{vazquez2009multiphase} specifically focused on attraction/repulsion interaction between two boundaries. They defined  elastic couplings between level set functions using dynamic force fields to model ribbon-like partitions. Note that none of the above mentioned methods guarantee the globally optimal solution.

In the discrete domain, Wu et al. \citep{wu2011region} can impose attraction/repulsion force between two surfaces by controlling the minimum and maximum distances between them. \cite{delong2009Globally}, and similarly \citep{ulen2013efficient,schmidt2012hausdorff},  enforced attraction/repulsion between pairs of regions, e.g. A and B,  by penalizing the intersection of $A$ and $B$ (i.e. area/volume of $A-B$). Such constraints are  encoded in \eqref{DB} using the penalty terms shown in  Table \ref{table:containment}.
\begin{table}[!h]
\caption{Energy terms for encoding containment with attraction/repulsion between $A$ and $B$ regions.}
\label{table:containment}
\begin{center}
\begin{tabular}{|cc|c|}
\hline
\multicolumn{3}{ |c| }{A \textit{attracts} B} \\
\hline
$f^A_p$&$f^B_q$&$W^{AB}_{pq}$\\
\hline
0&0&0\\
0&1&0\\
1&0&$\alpha$\\
1&1&0\\
\hline
\end{tabular}
\end{center}
\end{table}
In the containment setting mentioned in Table \ref{DBtable}, replacing infinity value with a positive value for  $W^{AB}_{pq}(0,1)>0$ creates a spring-like repulsion force between inner and outer boundaries. Hence, an attraction/repulsion constraint  is similar to a containment constraint  but with a different orientation.

\end{itemize}
In graph-based methods, e.g. \citep{delong2009Globally,ulen2013efficient}, increasing the
distance (or thickness) between regions requires more edges
to be added to the underlying graph, which increases  memory
usage and computation time. In fact, to impose a distance
constraint of $w$ pixels between two regions, \citep{delong2009Globally} and \citep{ulen2013efficient} need
to add $O(w^2)$ extra edges \emph{per pixel}. Therefore, although these graph-based methods are highly efficient in segmenting images with reasonable size and thickness constraint, they are not that efficient for large distance constraints. On the other hand, the memory usage and time complexity in the continuous methods (e.g. works proposed by \citep{nosrati2014local,nosrati2013bounded}) are independent of thickness constraints.

In addition to the above mentioned approaches, methods based on the artificial life framework (deformable organisms) also employ spatial distance constraints to maintain the organism's structure \citep{hamarneh2009deformable,prasad2011deformable}. In these models, the deformable organism evolves in a restricted way such that the distance between its skeleton and its boundary is restricted to be within a certain range.

\subsection{Adjacency prior}
Recently, several methods focused on ordering constraints and adjacency relationships on labels for semantic segmentation. As an example, ``sheep'' and ``wolf'' are unlikely to be next to each other and label transition from ``sheep'' to ``wolf'' should be penalized \citep{strekalovskiy2012nonmetric}. 

\begin{figure}
\begin{center}
\subfloat[]{\includegraphics[trim=0mm 0mm 0mm 0mm,width=.3\linewidth,clip=true]{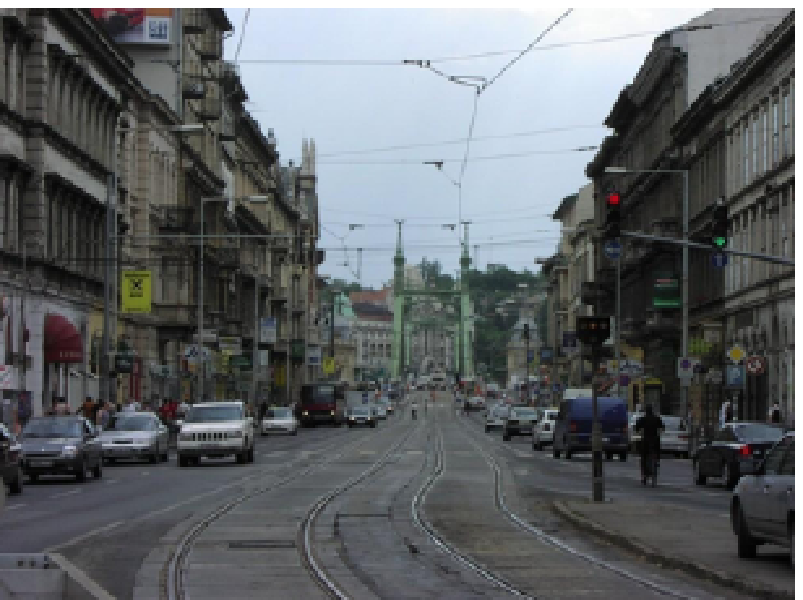}}~
\subfloat[]{\includegraphics[trim=0mm 0mm 0mm 0mm,width=.3\linewidth,clip=true]{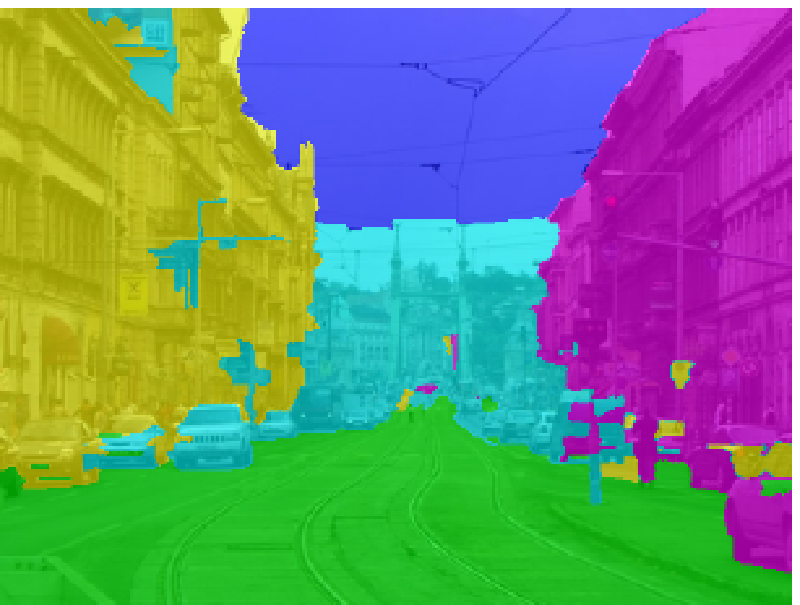}}~
\subfloat[]{\includegraphics[trim=0mm 0mm 0mm 0mm,width=.3\linewidth,clip=true]{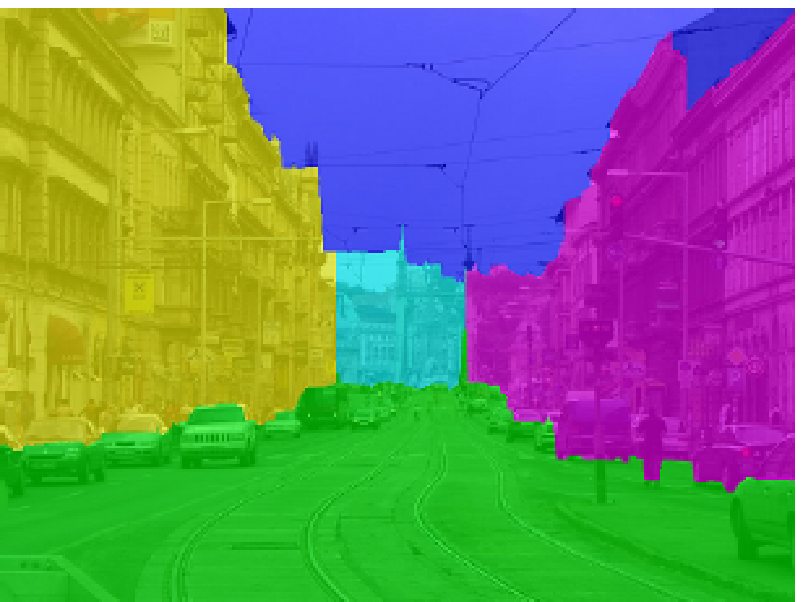}}
\end{center}
\caption{Tiered labelling. (a) Input image. Segmentation result (b) without and (c) with label ordering constraints. (Images from \citep{strekalovskiy2011generalized}) }
\label{fig:proximity}
\end{figure}

In the discrete settings, \cite{liu2008graph} proposed a graph-based method to incorporate label ordering constraints in scene labeling and tiered\footnote{Tiered labeling problem partitions an input image into multiple horizontal and/or vertical tiers.} segmentation. They assumed that an image is to be segmented into five parts (``centre'',``left'', ``right'', ``above'' and ``bottom'') such that a pixel labeled as ``left'' cannot be on the right of any pixel labeled as ``center'', etc. \cite{liu2008graph} encoded such constraints into the pair-wise energy term (regularization), i.e. $\sum_{(p,q)\in\mathcal{N}}V_{pq}(f_p,f_q)$. For example, if pixel $p$ is immediately to the left of $q$, to prohibit $f_p=$``center'' and $f_q=$``left'', then one defines $V_{pq}($``center'',``left''$)=\infty$. Generalizing this rule to other cases gives the following settings for $V_{p,q}$:
\begin{center}
\begin{tabular}{|c|c|c|c|c|c|}
\hline
$f_p \backslash  f_q$&\textbf{L}&\textbf{R}&\textbf{C}&\textbf{T}&\textbf{B}\\
\hline
\textbf{L}&0&$\infty$&$w_{pq}$&$w_{pq}$&$w_{pq}$\\
\textbf{R}&$\infty$&0&$\infty$&$\infty$&$\infty$\\
\textbf{C}&$\infty$&$w_{pq}$&0&$\infty$&$\infty$\\
\textbf{T}&$\infty$&$w_{pq}$&$\infty$&0&$\infty$\\
\textbf{B}&$\infty$&$w_{pq}$&$\infty$&$\infty$&0\\
\hline
\end{tabular}
\\
$p$ is the left neighbour of $q$
\begin{tabular}{|c|c|c|c|c|c|}
\hline
$f_p \backslash  f_q$&\textbf{L}&\textbf{R}&\textbf{C}&\textbf{T}&\textbf{B}\\
\hline
\textbf{L}&0&$\infty$&$\infty$&$\infty$&$w_{pq}$\\
\textbf{R}&$\infty$&0&$\infty$&$\infty$&$w_{pq}$\\
\textbf{C}&$\infty$&$\infty$&0&$\infty$&$w_{pq}$\\
\textbf{T}&$w_{pq}$&$w_{pq}$&$w_{pq}$&0&$\infty$\\
\textbf{B}&$\infty$&$\infty$&$\infty$&$\infty$&0\\
\hline
\end{tabular}
\\ $p$ is the top neighbour of $q$
\end{center}
Figure \ref{fig:proximity} illustrates an example of a tiered labelling. From optimization point of view and according to \citep{liu2008graph}, $\alpha$-expansion technique is more likely to
get stuck in a local minimum when ordering constraints are used, as $\alpha$-expansion acts on a single label ($\alpha$) at each move. In order to improve on $\alpha$-expansion moves, authors introduced two horizontal and vertical moves and allowed a pixel to have a
choice of labels to switch to, as opposed to just a single label $\alpha$. Although their proposed optimization approach leads to better results (compared to $\alpha$-expansion approach), the globally optimal solution is still not guaranteed.  \cite{felzenszwalb2010tiered} proposed an efficient dynamic programming algorithm to impose similar constraints as  \citep{liu2008graph} but with much less complexity. Their method computes the globally optimal solution in the class of tiered labelings.

In the continuous domain,  \cite{strekalovskiy2011generalized} proposed a generalized label ordering constraint which can enforce many complex geometric constraints while maintaining convexity. This method requires that
the constraint term obeys the triangle inequality, a requirement
that was later relaxed by introducing a convex relaxation
method for non-metric priors \citep{strekalovskiy2012nonmetric}. To do so, authors enforce non-metric label  distances in order to model arbitrary probabilities for label adjacency. The distances between different labels\footnote{Note that this is not a spatial distance but is a distance between label classes.} operates only directly on neighbouring pixels. This often leads to artificial one pixel-wide regions between labels to allow the transition between labels with very high or infinite distance. For example, both the ``wolf'' and ``sheep'' labels can be next to ``grass'' but they cannot be next to each other \citep{strekalovskiy2012nonmetric}. The method proposed in \citep{strekalovskiy2012nonmetric} would create an artificial ``grass'' region between ``wolf'' and ``sheep'' to allow for this transition. Obviously this one-pixel wide distance between ``wolf'' and ``sheep'' would not make the sheep any more secured! Generally, a neighbourhood larger than one pixel is needed to avoid these artificial labeling artifacts.  \cite{bergbauer2013morphological} addressed this issue and proposed a \emph{morphological proximity} prior for semantic image segmentation in a variational framework. The idea is to consider pixels as adjacent if they are within a specified neighbourhood of arbitrary size. Consider two regions $i$ and $j$ and their indicator functions $u_i$ and $u_j$, respectively. To see if  $i$ and $j$ are close to each other, the overlap between the dilation of the indicator function $u_i$, denoted by $d_i$, and the indicator function of $u_j$ is computed. The dilation of $u_i$ is formulated as:
\begin{align}
d_i(\bs{x})=\max_{z\in\mathcal{S}} u_i(\bs{x}+z),\;\;\;\forall \bs{x}\in\Omega
\end{align}
with a structuring element $\mathcal{S}$. For each pair of region $i$ and $j$, a \emph{proximity} penalty term is defined as:
\begin{align}
\sum_{1\leq i\leq j\leq n}\int_\Omega A(i,j)d_i(\bs{x})u_j(\bs{x})d\bs{x}
\;,
\end{align}
where $A(i,j)$ indicates the penalty for the co-occurrence of label $j$ in the proximity of  label $i$ such that $A(i,i)=0$. In \cite{bergbauer2013morphological}, this penalty term is relaxed and added to an energy functional (along with  regional and regularization terms), which is then optimized with the help of Lagrange multipliers. 

To the best of our knowledge, the adjacency and proximity priors as described above have not been utilized in medical image segmentation yet.

\subsection{Number of regions/labels}
In most segmentation problems, the number of regions is assumed to  be known beforehand. However, it is not the case in many applications and predefining  a fixed number of labels in these cases often causes over-segmentation.

The intuitive way to handling this problem is to penalize 
the total number of labels. For the given maximum number of regions/labels (at most $n$ labels),  which is available in most applications,  \cite{zhu1996region} proposed to partition images based on the following energy functional in the continuous domain:
\begin{align}
\min_{\Omega_i}\sum_{i=1}^n\left\{\int_{\Omega_i}\rho(\ell_i,\bs{x})d\bs{x}+\int_{\partial\Omega_i}ds\right\}+\gamma M\;,
\end{align}
where $\Omega_i$ is the region corresponding to label $\ell_i$; $\rho(\ell_i,\bs{x})$ is the data term that encodes the model of $\ell_i$ at pixel $\bs{x}$; the second term is the regularization term; and $M$ in the third term is the number of non-empty partitions (known as label cost prior). \cite{zhu1996region} optimized the above energy functional using a local optimization technique which converges to a local minimum. This approach was later adapted in the level set formulation by \citep{kadir2003unsupervised,ben2008region,brox2006level} that allow region-merging. A convex formulation of such constraint was proposed by  \cite{yuan2012continuous}. They enforced the label cost prior into multi label segmentation by solving the following convex optimization problem:
\begin{align}
&\min_{u(\bs{x})}\sum_{i=1}^n\left\{\int_{\Omega}u_i(\bs{x})\rho(\ell_i,\bs{x})d\bs{x}+\int_{\partial\Omega}|\nabla u_i(\bs{x})|d\bs{x}\right\}+\gamma \sum_{i=1}^n \max_{\bs{x}\in\Omega}u_i(\bs{x})\;,\notag \\
&s.t. \sum_{i=1}^n u_i(\bs{x})=1,\;\;u_i(\bs{x})\geq 0;\;\;\forall \bs{x}\in\Omega.
\end{align}

In the discrete domain, \cite{delong2012fast} developed an $\alpha$-expansion method to optimize a general energy functional incorporated with label cost in a graph-based framework. Along with unary (data) and pairwise (regularization) terms, \cite{delong2012fast} penalized each unique label that appears in the image by introducing the  following term:
\begin{align}
&\sum_{l\in\mathcal{L}}h_l\cdot\delta_l(f)\\
&\delta_l(f)=
\begin{cases}
1 & \exists p\in\Omega :f_p=l\\
0 & otherwise
\end{cases}
\;,
\end{align}
where $h_l$ is the non-negative label cost of label $l$.

\begin{figure}
\begin{center}
{\includegraphics[trim=0mm 0mm 0mm 0mm,width=1\linewidth,clip=true]{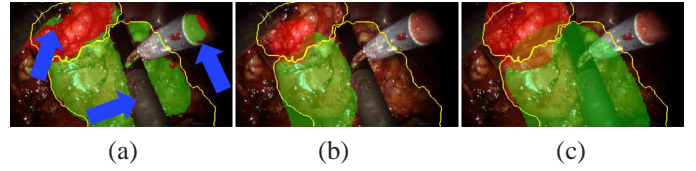}}\\
(a)~~~~~~~~~~~~~~~~~~~~~~~~~~~(b)~~~~~~~~~~~~~~~~~~~~~~~~~~~~~~~(c)
\end{center}
\caption{Endoscopic video segmentation. (a) Result of the active contour without edges (blue arrows indicate errors). (b,c) \cite{nosrati2014efficient}'s method (b) without and (c) with motion prior. Green: kidney; Red: tumor; Yellow: ground truth. (Images adopted from \citep{nosrati2014efficient})}
\label{fig:model}
\end{figure}
\subsection{Motion prior}
The segmentation and tracking of moving objects in videos have a wide variety of applications in medical image analysis, e.g. in echocardiography \citep{dydenko2006level}. 
 \cite{paragios1999geodesic} used motion prior to constrain the evolution of a level set function by integrating motion estimation and tracking into  a level set-based framework. %They assumed the motion is linear.  
 \cite{dydenko2006level} proposed a method to segment and track the cardiac structure in high frame rate echocardiographic images.  The motion field is estimated from the level set evolution. Both \citep{paragios1999geodesic} and \citep{dydenko2006level} perform motion estimations under the constraint of an affine model. More complex motion prior is used in object tracking. For example, to track the LV in echocardiography,  \cite{orderud2007real} employed the Kalman filter, which is an optimal recursive algorithm that uses a series of measurements observed over time to estimate the desired variables (i.e. displacement in motion estimation).

Recently, \cite{nosrati2014efficient} proposed an efficient technique to segment multiple objects in intra-operative multi-view endoscopic videos based on priors captured from pre-operative data. Their method allows for the inclusion of laparoscopic camera motion model to stabilize the segmentation in the presence of a large occlusion (Figure \ref{fig:model}). This feature is especially useful in robotic minimally invasive surgeries as camera motion signals can be easily read using the robot's API and be incorporated into their formulation. 
%As these methods focused more on tracking, which is outside the scope of this report, we refer interested readers to \citep{tang2012tongue}.

\subsection{Model/Atlas}
Atlas-based segmentation has also been particularly useful
in medical image analysis applications. Works that adopt atlas-based approach include \citep{gee1993elastically,collins1995automatic,collins1997animal,iosifescu1997automated}.  An atlas has the
ability to encode (non-pathological) spatial relationships
between multiple tissues, anatomical structures or organs. In atlas-based image segmentation, the image is non-rigidly deformed and registered with a model or atlas that has been labelled previously. Applying the inverse transformation of the labels to the image space gives the segmentation. However, atlas-based segmentation has so far been restricted to single (albeit multi-part or multi-region) object instance, and  does not address spatially-recurring objects in the scene. Also, atlases usually are built from datasets of manually segmented images. These manual segmentations may not always be available, and/or can not be used to define a representative template for a given object in a straightforward manner.

The performance of atlas-based segmentation techniques  relies on an accurate registration. Surveying registration methods is beyond the scope of this paper. Interested readers may refer to \citep{sotiras2013deformable}, \citep{hill2001medical}, and \cite{crc2013c} for more details on image registration.

In the field of computer vision (non-medical), a few techniques used 3D models of objects (more realistic 
but more complex) to segment 2D images. 
\cite{prisacariu2012pwp3d} proposed a variational method to segment an object in a 2D image by optimizing a Chan-Vese energy functional with respect to six pose parameters of the object model in 3D. The idea is to transform the object's model in 3D so that its projection on the 2D image delineates the object of interest. Consider segmenting a single object in an image, \cite{prisacariu2012pwp3d} used the following energy:
\begin{align}
E(\phi(\bs{x}))=\int_\Omega \rho_f(\bs{x})H(\phi(\bs{x}))+\rho_b(\bs{x})(1-H(\phi(\bs{x})))d\Omega
\;,
\end{align}
where $\rho_f$ and $\rho_b$ are two monotonically decreasing functions, measuring matching quality of the image pixels with respect to the foreground and background models, respectively. Instead of optimizing $E(\phi)$ with respect to the level set function $\phi$, authors in \citep{prisacariu2012pwp3d} proposed to minimize $E(\phi)$ with respect to the pose parameters ($\xi_i$) of the object of interest in 3D space:
\begin{align}
\frac{\partial E}{\partial \xi_i}=(\rho_f-\rho_b)\frac{\partial H(\phi)}{\partial \xi_i}=(\rho_f-\rho_b)\delta(\phi)
\begin{bmatrix}
    \frac{\partial \phi}{\partial x}   & \frac{\partial\phi}{\partial y} 
\end{bmatrix}
\begin{bmatrix}
	\frac{\partial x}{\partial\lambda_i}\\
	\frac{\partial y}{\partial\lambda_i}
\end{bmatrix}
\;.
\end{align}
Unlike \citep{prisacariu2012pwp3d},  \cite{sandhu2011nonrigid} derived a gradient flow
for the task of non-rigid pose estimation for a single object and
used kernel PCA to capture the variance in the space of shapes.
Later, \cite{prisacariu2013simultaneous}, introduced non-rigid pose parameters into the same optimization framework. They capture 3D shape variance by learning
non-linear probabilistic low dimensional latent spaces, using
the \emph{Gaussian process latent variable}  dimensionality reduction
technique. All three aforementioned works \citep{prisacariu2012pwp3d,prisacariu2013simultaneous,sandhu2011nonrigid}
assume that the camera parameters (for 3D to 2D projection) are given.

Recently, inspired by \citep{prisacariu2012pwp3d}, \cite{nosrati2014efficient} proposed a closed-form
solution to segment multiple tissues in multi-view endoscopic videos based on pre-operative data. Their method  simultaneously estimates the 3D pose of tissues
in the pre-operative domain as well as their non-rigid deformations from their pre-operative state.   They  validated their approach on \emph{in vivo} surgery data of partial nephrectomy and showed the potential of their method in an augmented reality environment for minimally invasive surgeries. Figure \ref{fig:model} shows an example of segmentation of kidney and tumour in an endoscopic view produced by their method.

\section{Summary, discussion, and conclusions    }
\label{sec:conclusion}
Segmentation techniques are aimed at partitioning (crisply or fuzzily) an image into meaningful parts (two or more). Traditional segmentation approaches (e.g. thresholding, watershed, or region growing) proved incapable of robust and accurate segmentation due to noise, low contrast and complexity of objects in medical images. By incorporating prior knowledge of objects into rigorous optimization-based segmentation formulations, researchers developed more powerful techniques capable of segmenting specific (targeted) objects.

In recent years, several types of prior knowledge have been utilized in a variety of forms (e.g. via user interaction, object shape and appearance, interior and boundary properties, regularization mechanisms, topological and geometrical constraints,  moment priors, distance and adjacency constraints, as well as motion and model/atlas-based priors). In this paper, we attempted to provide a comprehensive survey of such image segmentation priors, with a focus on medical imaging applications, including both high-level information as well as the essential technical and mathematical details. We compared different prior in terms of the domain settings (continuous vs. discrete) and the optimizibility (e.g. convex or not).

It is important to appreciate that, although incorporating richer prior into an objective function may increase the fidelity of the energy functional (by better modelling the underlying problem), this typically comes at the expense of complicating its optimization (lower optimizibility). On the other hand, focusing on optimizibility by simplifying the energies might decrease the fidelity of the energy functions. In other words, be wary of segmentation algorithms that always converge to the globally optimal but inaccurate solution, or ones that rely heavily on intricate initialization or meticulously tweaked parameters. Consequently, recent research surveyed has focused on developing methods that increase the optimizibility of energy functions (e.g. by proposing convex or submodular energy terms) without sacrificing the fidelity.

In addition to the optimizability-fidelity tradeoff that is impacted by the choice of priors, it is important to observe the runtime and memory efficiency of proposed medical image segmentation algorithms. For example, graph-based approaches may not be very efficient in handling very large images and they often produce artifacts like grid-bias errors (also known as metrication error) due to their discrete nature.

Despite the great advances that have been made in terms of increasing the fidelity and optimizibility of various segmentation energy models, there is still more to be done. We believe that through ongoing research, new methods will be proposed that allow for models that are faithful to the underlying problems, while being globally optimizable, memory- and time-efficient regardless of image size, and are free from any artifacts like metrication error.

In extending prior information in medical image segmentation, there are several directions to explore.  %Developing a comprehensive system that can decide on what prior to use for each medical image analysis task seems to be very useful and exciting for the community. In fact, one direction of research may focus on  designing a system that can learn what prior to use for  different medical applications.  
One direction may focus on consolidating  all of these previously mentioned priors such that a user (or an automatic system) can add one or more of these priors as a module to the segmentation task at hand.  Such system is expected to minimize  user inputs like manual initialization.  Although many efforts have been made to convexify energy terms, many priors (especially when combined together) are non-convex (non-submodular) and hard to optimize. As convex relaxation and convex optimization techniques are becoming popular recently, research emphasis that focuses on convexification of energy functions with as many priors as needed would be an important step toward automatic image segmentation.

In optimization-based segmentation that encodes a set of desired priors, it is important to consider how to combine their respective energy (or objective) terms. The most common approach for dealing with such a multi-objective optimization is to scalarize the energies (via a linear sum of terms).  Aside from choosing which priors are relevant and which mathematical formulae encode them, how to learn and set the contribution weight of each term needs to be explored carefully especially when there is not enough training data. When large sets of training data are available, machine learning techniques have been used to discover a good set of weights that adapt to image class, weights that change per image, and spatially adaptive weights.

The priors we reviewed and introduced in this thesis have been specifically and carefully designed to address particular segmentation problem. Another potential complementary approach that is worthy of future exploration is to attempt to learn the priors (not only their weight in the objective function) from available training data.

Future research directions could also focus on combining the hand-crafted features with machine learning techniques in case of availability of training data. For example it makes sense to use expert knowledge when the training data is not available and increase the contribution of machine learning techniques as more data becomes available and/or expert knowledge is harder to collect.

\section*{Conflict of Interest}
The authors declare that there is no conflict of interest regarding the publication of this article.

\section*{References}
\bibliographystyle{model2-names}
\bibliography{Depth2}

%% not want to use model2-names.bst.



%% \bibitem must have one of the following forms:
%%   \bibitem[Jones et al.(1990)]{key}...
%%   \bibitem[Jones et al.(1990)Jones, Baker, and Williams]{key}...
%%   \bibitem[Jones et al., 1990]{key}...
%%   \bibitem[\protect\citepauthoryear{Jones, Baker, and Williams}{Jones
%%       et al.}{1990}]{key}...
%%   \bibitem[\protect\citepauthoryear{Jones et al.}{1990}]{key}...
%%   \bibitem[\protect\astroncite{Jones et al.}{1990}]{key}...
%%   \bibitem[\protect\citepname{Jones et al., }1990]{key}...
%%   \harvarditem[Jones et al.]{Jones, Baker, and Williams}{1990}{key}...
%%


% \end{thebibliography}

\end{document}